\documentclass[12pt]{article}
\usepackage{geometry}
\usepackage{url,amsmath,amsfonts,amssymb,amsthm,algorithm,longtable,bm}
\usepackage{authblk,amsmath,amssymb,amsthm,amsfonts,natbib,bm,setspace,graphics,graphicx,url,caption,multicol,multirow,longtable,lscape,verbatim,enumerate,color,float,graphicx,booktabs,setspace}
\usepackage[normalem]{ulem}
\usepackage{xr} 

\newcommand{\tR}{\mathbb{R}}

\newcommand{\bJ}{\mathbf{J}}
\newcommand{\bQ}{\mathbf{Q}}

\newcommand{\bb}{\mathbf{b}}
\newcommand{\bD}{\mathbf{D}}
\newcommand{\bM}{\mathbf{M}}

\newcommand{\bL}{\mathbf{L}}

\newcommand{\bG}{\mathbf G}

\newcommand{\bS}{\mathbf{S}}

\newcommand{\bv}{\mathbf{v}}
\newcommand{\bSigma}{\mathbf{\Sigma}}

\newcommand{\bepsilon}{\bm{\epsilon}}
\newcommand{\bE}{\mathbf{E}}
\newcommand{\bI}{\mathbf{I}}
\newcommand{\bR}{\mathbf{R}}

\newcommand{\bA}{\mathbf{A}}
\newcommand{\bC}{\mathbf{C}}

\newcommand{\bO}{\mathbf{O}}

\newcommand{\cO}{\mathcal{O}}
\newcommand{\bt}{\mathbf{t}}

\newcommand{\diag}{\mathrm{diag}}
\newcommand{\Diag}{\mathrm{Diag}}
\newcommand{\tr}{\mathrm{tr}}
\newcommand{\bX}{\mathbf{X}}

\newcommand{\bW}{\mathbf{W}}

\newcommand{\bT}{\mathbf{T}}

\newcommand{\bx}{\mathbf{x}}

\newcommand{\bz}{\mathbf{z}}

\newcommand{\btheta}{\mathbf{\theta}}

\newcommand{\bzero}{\mathbf{0}}
\def\Cov{{\rm Cov}\,}
\def\E{{\mathbb E}\,}

\newcommand{\Col}{\textrm{Col}}
\newcommand{\Row}{\textrm{Row}}
\newtheorem{thm}{Theorem}
\newtheorem{lem}{Lemma}

\usepackage{xr-hyper}

\usepackage{hyperref}
\hypersetup{
    colorlinks=true,
    linkcolor=blue,
    filecolor=black,      
    urlcolor=black,
    citecolor=blue
}

\makeatletter
\newcommand*{\addFileDependency}[1]{
  \typeout{(#1)}
  \@addtofilelist{#1}
  \IfFileExists{#1}{}{\typeout{No file #1.}}
}
\makeatother

\begin{document}

\title{Probabilistic Joint and Individual Variation Explained (ProJIVE) for Data Integration}

\author[1]{Raphiel J. Murden}
\author[1]{Ganzhong Tian}
\author[2]{Deqiang Qiu}
\author[1]{Benjamin B. Risk\footnote{brisk@emory.edu, 1518 Clifton Rd NE, Atlanta, GA 30322}}
\author[3]{FOR THE ALZHEIMER'S DISEASE NEUROIMAGING INITIATIVE\footnote{Data used in preparation of this article were obtained from the Alzheimer’s Disease Neuroimaging Initiative (ADNI) database (adni.loni.usc.edu).}}

\affil[1]{Department of Biostatistics and Bioinformatics, Rollins School of Public Health, Emory University, 1518 Clifton Rd NE, Atlanta, GA 30322, USA 
}
\affil[2]{Department of Radiology and Imaging Sciences, Emory University School of Medicine, 1364 Clifton Rd NE, Atlanta, GA  30322, USA}


\date{}
\maketitle

x\begin{abstract}
{Collecting multiple types of data on the same set of subjects is common in modern scientific applications including genomics, metabolomics, and neuroimaging. Joint and Individual Variation Explained (JIVE) seeks a low-rank approximation of the joint variation between two or more sets of features captured on common subjects and isolates this variation from that unique to each set of features. We develop an expectation-maximization (EM) algorithm to estimate a probabilistic model for the JIVE framework. The model extends probabilistic PCA to multiple datasets. Our maximum likelihood approach simultaneously estimates joint and individual components, which can lead to greater accuracy compared to other methods. We apply ProJIVE to measures of brain morphometry and cognition in Alzheimer's disease. ProJIVE learns biologically meaningful sources of variation, and the joint morphometry and cognition subject scores are strongly related to more expensive existing biomarkers. Data used in preparation of this article were obtained from the Alzheimer’s Disease Neuroimaging Initiative (ADNI) database. Code to reproduce the analysis is available at \href{https://github.com/thebrisklab/ProJIVE/}{https://github.com/thebrisklab/ProJIVE}. Supplementary materials for this article are available online.}
\end{abstract}

\pagebreak
\doublespacing

\section{Introduction}

Data integration methods analyze multiple datasets collected from the same participants. Collections of datasets have become common in neuroimaging, genomics, proteomics, and other fields \citep{MeullerADNI2007, allen2014uk,tomczak2015cancer}. Statistical approaches to data integration date as far back as the 1930s, with the introduction of Canonical Correlation Analysis (CCA) \citep{Hotelling1936}. While CCA focuses on exploring structure that is shared (i.e., joint) between two datasets, more recently, methods such as JIVE (Joint and Individual Variance Explained) add to the data integration framework by also extracting structure that is unique (i.e., individual) to each of the datasets \citep{Lock2013}.

The current manuscript develops a probabilistic, model-based method to implement JIVE. 
Several data integration methods related to JIVE have been developed in recent years. R.JIVE iterates between estimating the joint matrices and individual matrices until convergence, and it was used to explore the relationship between microRNA and gene expression in patients with malignant brain tumors \citep{Lock2013,o2016r}.  AJIVE \citep{Feng2018Angle-basedExplained} uses matrix perturbation theory to estimate joint ranks and a non-iterative method for extracting joint structure, followed by a separate step to estimate individual structure. This approach was applied to behavioral and imaging data from the Human Connectome Project \citep{Yu2017}. Bayesian Simultaneous Factorization and Prediction uses a Bayesian probabilistic model for a JIVE decomposition of multi-omic predictors to predict an outcome \citep{samorodnitsky2024bayesian}. Other data integration methods include the following: robust-JIVE, which utilizes an L-1 norm minimization technique for computer vision applications \citep{Sagonas2017}; Common and Orthogonal Basis Extraction, which is similar to JIVE \citep{Zhou2016}; Structural Learning and Integrative Decomposition, which allows for partially shared structure \citep{Gaynanova2019StructuralData}; Data Integration Via Analysis of Subspaces (DIVAS), which estimates partially shared structure using principal angles between subspaces and uses a rotational bootstrap to estimate the signal ranks \citep{prothero2024data}; Decomposition-based Canonical Correlation Analysis (D-CCA), a CCA method that includes individual random variables with covariance equal to zero \citep{Shu2019D-CCA:Datasets}; Simultaneous Non-Gaussian Component Analysis, which uses non-Gaussianity for dimension reduction rather than variance \citep{risk2021simultaneous}; and Generalized Integrative Principal Component Analysis (GIPCA), which extends principal component analysis for exponential family densities to data integration \citep{zhu2020generalized}.

While much progress has been made in data integration studies, many of the previous methods do not propose statistical models. For instance, JIVE was formulated as a framework for decomposing data into matrices defined by score subspaces, and a generative model for a new subject was not defined. We will argue that a data integration method using maximum likelihood to model subject scores and loadings may improve both accuracy and interpretability. Unlike JIVE, Bayesian Simultaneous Factorization and Prediction, D-CCA, and GIPCA propose statistical models with latent variables. Bayesian simultaneous factorization and prediction involves priors and hyperparameters in a Bayesian framework, which may be computationally challenging. D-CCA is only formulated for two datasets and involves multiple estimation steps. GIPCA uses the single parameter exponential family without dispersion (variance) parameters, which we will examine in simulations. 

Probabilistic principal component analysis, or pPCA, formulates a statistical model for PCA in which subject scores are normally distributed random effects and variable loadings are fixed parameters \citep{Bishop1999}. The pPCA model is a special case of the factor analysis model \citep{lawley1962factor}, where the former assumes the errors have a diagonal covariance matrix with equal variances, and the latter assumes a diagonal covariance matrix in which variances can differ. We generalize the pPCA framework to two or more datasets and develop a probabilistic approach to JIVE, which we call Probabilistic JIVE or ProJIVE. ProJIVE models sources of joint variation as subject random effects shared between datasets and sources of individual variation as subject random effects unique to each dataset. Our approach increases interpretability by directly modeling quantities of interest, may be more accurate than a multi-step decomposition process, and can be useful when the covariance matrix is a useful summary measure, including some applications in which data are not Gaussian.

Our model is closely related to inter-battery factor analysis (IBFA) \citep{Tucker1958AnAnalysis}. Our contributions include a latent variable model and novel algorithm. IBFA posits latent variables shared by two datasets and latent variables unique to each dataset and, therefore, is closely related to JIVE. Early work derived a maximum likelihood solution for Gaussian-distributed joint scores but did not assume a low-rank structure for the unique factors \citep{Browne1979TheAnalysis}. More recently, \cite{Klami2013BayesianAnalysis} proposed a Bayesian inter-battery factory analysis for two datasets and \cite{klami2014group} proposed group factor analysis for multiple datasets, which use  variational inference to approximate the likelihoods and sparsity priors to approximate block-wise sparsity.

In Section \ref{methods}, we describe the ProJIVE model, identifiability, and an EM algorithm to estimate its parameters. Section \ref{sec:sims.pjive} conducts a simulation study comparing ProJIVE to existing JIVE methods. ProJIVE achieved greater accuracy in estimating  scores and variable loadings when compared to R.JIVE, AJIVE, and GIPCA in a variety of simulation settings, including non-Gaussian simulations. In Section \ref{sec:data}, we evaluate the effectiveness of ProJIVE via an analysis of data obtained from the Alzheimer's Disease Neuroimaging Initiative (ADNI). Our application to the ADNI data integrates
measures of brain morphometry (i.e., thickness, surface area, and volume) with measures of cognition and behavior to estimate their joint sources of variation. Results demonstrate ProJIVE's utility: joint subject scores strongly associate with variables such as AD diagnosis, the presence of the genetic marker apolipoprotein E4 (ApoE4),  amyloid-beta levels from Positron Emission Tomography (PET) imaging using florbetapir (18F-AV45), and neuronal cell metabolism derived from fluorodeoxyglucose (FDG) PET. Section \ref{sec:discussion} discusses our model and limitations. Proofs are in the Supporting Information.

\section{Methods}\label{methods}

\subsection{The original JIVE decomposition}

Suppose features arising from the same set of $n$ observational units (e.g., subjects) are collected in different datasets, $\bX_k$ for $k=1, \dots , K$, such that each dataset has dimension $p_k \times n$. JIVE assumes $\bX_k=\bJ_k+\bA_k+\bE_k$, where $\bJ_k$ is the joint signal, $\bA_k$ the individual signal, and $\bE_k$ the noise, where $\E(\bE_k)=\bzero$ with mutually independent entries. Let $\Row(\bG)$ define the row space of a matrix $\bG$: $\Row(\bG)=\{\bv \in \tR^n:  \bv^\top = \bt^\top \bG \textrm{ for some } \bt \in \tR^{p_k}\}$.
The JIVE framework assumes (i) $\Row(\bJ_k) = \Row(\bJ_{k'})$ for $k \ne k' \in \{1,\dots,K\}$; (ii) $\Row(\bJ_k) \perp \Row(\bA_k)$ (i.e., $\bJ_k\bA_k^\top  = \bzero$); and (iii) $\cap_{k=1,\dots,K} \Row(\bA_k) = \bzero$ \citep{Lock2013,Feng2018Angle-basedExplained}. R.JIVE includes a variant in which $\Row(\bA_k) \perp \Row(\bA_{k'})$, $k=1,\dots,K$ \citep{o2016r}. The decomposition $\bX_k = \bJ_k + \bA_k + \bE_k$ is unique when one assumes rank($\bJ_k$)+rank($\bA_k$) = rank($\bJ_k + \bA_k$), $k=1,\dots,K$ \citep{Lock2013}. \cite{Feng2018Angle-basedExplained} further assume that the noise is isotropic. 

\subsection{Probabilistic JIVE} 
We propose a probabilistic model for $\bx_{ik} \in \tR^{p_k}$, which represents the random vector of $p_k$ features from the $i$th subject, $i=1,\dots,n$, in the $k$th data block, $k=1,\dots,K$. Assume $\mathbb{E}(\bx_{ik}) = \mathbf{0}$. Let $r_J$ define the number of joint components and $r_{Ik}$ the number of individual components in the $k$th dataset. The loading matrices $\bW_{Jk} \in \tR^{p_k \times r_J}$ and $\bW_{Ik} \in \tR^{p_k \times r_{Ik}}$ represent variable-specific summaries to joint and individual variance, respectively, for the $k^{th}$ data-block. Our model is given by 
\begin{align} 
	&\bx_{ik}=\bW_{Jk} \bz_i + \bW_{Ik} \bb_{ik}+\bepsilon_{ik},\label{eqn:jive.ml.mod} \\
&(\bz_i^\top, \bb_{i1}^\top,\dots, \bb_{iK}^\top)^\top  \overset{iid}{\sim} N(\mathbf{0},\bI),	\; \bepsilon_{ik} \overset{iid}{\sim}N(\mathbf{0},{\bD_k}),\;\Cov (\bepsilon_{ik},\bepsilon_{ik'})=\bzero, \nonumber \\
	&\Cov\left[(\bz_i^\top, \bb_{i1}^\top,\dots,\bb_{iK}^\top)^\top,\bepsilon_{ik}\right]=\bzero, \nonumber \\
	&\textrm{rank}(\bW_{Jk})=r_J,\; 	\textrm{rank}([\bW_{Jk},\bW_{Ik}]) < p_k, \mbox{ and } \{k,k'\} \subseteq 1,\dots,K, \mbox{ where } \bD_k \mbox{ is diagonal}. \nonumber
\end{align}
In our identifiability results in Section \ref{sec:identifiability}, we further assume that $\bD_k = \sigma_k^2 \bI$. Then under this block-specific isotropic error assumption, the model can be seen as an extension of probabilistic PCA \citep{tipping1999probabilistic} to multiple datasets, where the joint scores $\bz_i$ capture shared information.

In our application (Section 4), we center the data such that $\sum_i \bx_{ik}=\bzero$ and we normalize each variable to unit variance. We recommend normalization when the variables within each data block are measured in different units or scales, as in the cognitive dataset, or when variables should be weighed equally, as desired for the brain regions in our neuroimaging dataset. However, variance normalization may be inappropriate in bulk sample gene expression, where normalization could result in a large biological signal being downweighted relative to genes with no signal but comparable measurement error. 

In many applications, joint scores are of particular interest and can be investigated as potential prodromes or discriminating factors among subgroups \citep{Lock2013}. On the other hand, examination of the joint and individual variable loadings $\{\bW_{Jk}, \bW_{Ik}: k=1,\dots,K\}$ can lead to discovery of the features that drive shared variation or individual variation \citep{Yu2017,Kashyap2019Individual-specificBehavior}. 
	
For the case of $K=2$, the covariance of the ProJIVE model is
\begin{equation} 
    \Cov \begin{pmatrix} \bx_{i1} \\ \bx_{i2}\end{pmatrix} = \begin{pmatrix} \bW_{J1}\bW_{J1}^\top + \bW_{I1}\bW_{I1}^\top + \bD_1 & \bW_{J1}\bW_{J2}^\top \\ \bW_{J2}\bW_{J1}^\top & \bW_{J2}\bW_{J2}^\top + \bW_{I2}\bW_{I2}^\top + \bD_2 \end{pmatrix}.\label{eqn:cov}
\end{equation}
Let $\bx_i = [\bx_{i1}^\top,\dots,\bx_{iK}^\top]^\top$, $i=1,\dots,n$, and recall that we assume $\E \bx_i = 0$. Define the population covariance matrix: $\bC = \Cov (\bx_i)$. The log-likelihood is 
\begin{equation}
\ell(\bW_{Jk},\bW_{Ik},{\bD_k},k=1,\dots,K;\;\bx_1,\dots,\bx_n)= - \frac{n}{2} \{\log(|2 \pi \bC|) + tr(\bC^{-1}\bS)\},\label{eqn:dat.ll}
\end{equation}
where $\bS=\frac{1}{n}\Sigma_{i=1}^n \bx_i \bx_i^\top$, $|\cdot|$ denotes the determinant, and $tr(\cdot)$, the trace.

\subsection{Connections between ProJIVE and previous models}\label{sec:connections} 
We discuss three orthogonality assumptions and their relationship to other data integration methods.
(i) \emph{Orthogonality between joint and individual scores.} We assume $\Cov (\bz_i,\bb_{ik})=0$, which is the random variable analogue of the orthogonality between joint and individual subspaces that is assumed in JIVE \citep{Lock2013,Feng2018Angle-basedExplained}. Rather than enforce orthogonality of the matrices with respect to the Euclidean inner product (which is done in JIVE), we will use a Gaussian likelihood with the EM algorithm to achieve our decomposition. (ii) \emph{Orthogonality between individual scores.} We further assume $\Cov (\bb_{ik},\bb_{ik'}) = 0$, and in this respect is similar to the assumption of orthogonality on the $\mathcal{L}$-2 space of random variables in D-CCA (D-CCA is defined for the special case of $K=2${, does not assume Gaussianity,} and does not assume orthogonality between joint and individual) \citep{Shu2019D-CCA:Datasets}. (iii) \emph{Mutual orthogonality: $\Cov (\bz_i,\bb_{ik})=0$ and $\Cov(\bb_{ik},\bb_{ik'})=0$}. This contrasts with AJIVE and D-CCA, but mutual orthogonality conditions were considered in R.JIVE, where it increased robustness to mis-specified ranks \citep{o2016r}. Mutual orthogonality was also enforced in \cite{Gaynanova2019StructuralData}, who developed the model for partially shared structure. 

Although beyond the scope of this paper, our approach naturally extends to the partially shared structure considered in  \cite{Gaynanova2019StructuralData}. We can define random variables that are only shared between some of the datasets. For $K=3$, we could augment the model to include partially shared latent factors, e.g., $\bz_{i,12}$ shared between datasets 1 and 2 and $\bz_{i,13}$ shared between datasets 1 and 3, 
\begin{align}
    \bx_{i1}=\bW_{J1} \bz_i + \bW_{P(1,2),1}\bz_{i,12} + \bW_{P(1,3),1} \bz_{i,13}+ \bW_{I1} \bb_{i1}+\bepsilon_{i1},\label{eq:partialshared}
\end{align}
where $\bW_{P(1,2),1}$ and $\bW_{P(1,3),1}$ capture the loadings on dataset 1 for the partially shared structure between datasets 1 and 2 and datasets 1 and 3, respectively.

\subsection{Model identifiability}\label{sec:identifiability}

It is easy to see that the density defined by \eqref{eqn:jive.ml.mod} is invariant to orthogonal transformations of the components. For the general case in which we allow for diagonal $\bD_k$, identifiability results are  challenging. The situation is the same as occurs in factor analysis, in which there exist cases where the signal and noise are not uniquely identified \citep{SHAPIRO19851}.
In this section, we examine the case with $\bD_k = \sigma_k^2 \bI$ for $k=1,2$. Isotropic errors were also assumed in AJIVE \citep{Feng2018Angle-basedExplained} Interestingly, the joint scores and loadings have an indeterminacy up to certain linear transformations that form a set larger than the set of orthogonal transformations. Let $\bA \succeq 0$ denote that a matrix $\bA$ is positive semidefinite. Let $\cO$ denote the class of orthogonal matrices. 
\begin{thm}
Suppose $K=2$ and let $f_\Phi$ define the multivariate normal density according to the parameters $\{ \mathbf{W}_{J1}, \mathbf{W}_{J2}, \mathbf{W}_{I1}, \mathbf{W}_{I2}, \; \sigma_1^2, \; \sigma_2^2 \}$ and the assumptions in \eqref{eqn:jive.ml.mod}. Let $f_{\Phi^*}$ denote the multivariate normal density with parameters $\{\mathbf{W}_{J1}^*,\mathbf{W}_{J2}^*,\mathbf{W}_{I1}^*,\mathbf{W}_{I2}^*,\sigma_1^{*2},\sigma_2^{*2}\}$ and the assumptions in \eqref{eqn:jive.ml.mod}. Then $f_\Phi = f_{\Phi^*}$ if and only if 
\begin{enumerate}
    \item $\sigma_1^{*2}=\sigma_1^2,\;\sigma_2^{*2}=\sigma_2^{2}$, 
    \item There exists an $r_J \times r_J$ non-singular matrix $\mathbf{T}_1$ such that $\mathbf{W}^*_{J1} = \mathbf{W}_{J1} \mathbf{T}_1$ and $\mathbf{W}^*_{J2} = \mathbf{W}_{J2} (\mathbf{T}_1^{-1})^\top$   and
\begin{align}
\mathbf{W}_{I1} \mathbf{W}_{I1}^\top + \mathbf{W}_{J1}(\mathbf{I} - \mathbf{T}_1 \mathbf{T}_1^\top )\mathbf{W}_{J1}^\top \succeq 0,\label{eq:part1} \\ 
\mathbf{W}_{I2} \mathbf{W}_{I2}^\top + \mathbf{W}_{J2}(\mathbf{I} - (\mathbf{T}_1^{-1})^\top \mathbf{T}_1^{-1})\mathbf{W}_{J2}^\top \succeq 0 \label{eq:part2},
\end{align}

\item $\mathbf{W}_{I1}^*$ is defined such that
$\mathbf{W}_{I1}^* (\mathbf{W}_{I1}^*)^\top = \mathbf{W}_{I1} \mathbf{W}_{I1}^\top + \mathbf{W}_{J1}(\mathbf{I} - \mathbf{T}_1 \mathbf{T}_1^\top )\mathbf{W}_{J1}^\top$, and $\mathbf{W}_{I2}^*$ is defined such that
$\mathbf{W}_{I2}^* (\mathbf{W}_{I2}^*)^\top = \mathbf{W}_{I2}\mathbf{W}_{I2}^\top + \mathbf{W}_{J2}(\mathbf{I} - (\mathbf{T}_1^{-1})^\top \mathbf{T}_1^{-1})\mathbf{W}_{J2}^\top$.
\end{enumerate}\label{thm:kequal2}
\end{thm}
The proof is in Web Appendix A.  
In short, this theorem provides the necessary and sufficient conditions for different parameters to obtain the same likelihood. Item 1 indicates the error variances are identifiable. Item 2 indicates the joint loadings are identifiable up to a class of linear transformations that includes all orthogonal matrices and possibly some nonsingular matrices. The number of joint components is identifiable, since $\Cov (\bx_{i1},\bx_{i2})$ has to have rank $r_J$. Item 3 indicates that when a non-orthogonal $\bT_1$ exists satisfying \eqref{eq:part1} and \eqref{eq:part2}, the individual loading matrices are susceptible to leakage from the joint signal. In this case, the number of individual components may also not be identifiable. We provide an example of this with our supplementary code available at \url{https://github.com/thebrisklab/ProJIVE}. Define the joint variation as $R_{Jk}^2=||\bW_{Jk}||_F^2/\sum_{j=1}^{p_k} \textrm{var}(x_{ijk})$, where $\textrm{var}(x_{ijk})$ is the variance of the $j$th variable in the $k$th dataset. When a non-orthogonal $\bT_1$ exists satisfying \eqref{eq:part1} and \eqref{eq:part2}, the breakdown of the joint versus individual variation is not identifiable.

It is interesting to consider the case in which the columns of $\bW_{Jk}$ and $\bW_{Ik}$ are not linearly independent, i.e., $\Col (\bW_{Jk}) \cap \Col (\bW_{Ik}) \ne \{\bzero\}$. Then there can exist $\bT_k$ not in $\cO$ that satisfy \eqref{eq:part1} and \eqref{eq:part2}. Linearly dependent columns between $\bW_{Jk}$ and $\bW_{Ik}$ occur by construction in the D-CCA decomposition, see equation (16) in  \cite{Shu2019D-CCA:Datasets}. Insight can be gained by examining the case of $\bW_{J1}=\bW_{I1}$, $\bW_{J2}=\bW_{I2}$, and $\bT_1 = \lambda \bI$. Then from \eqref{eq:part1} and \eqref{eq:part2}, we have
$(2 - \lambda^2) \bW_{I1} \bW_{I1}^\top \succeq 0$ and $(2 - 1/\lambda^2) \bW_{I2} \bW_{I2}^\top \succeq 0$, which implies $\frac{1}2 \le \lambda^2 \le 2$. Then defining $\bW_{I1}^*=\sqrt{2-\lambda^2}\bW_{I1}$ and $\bW_{I2}^* = \sqrt{2-1/\lambda^2} \bW_{I2}$, we see how the decomposition of joint and individual variation is not unique.

In practice, conventional approaches such as Varimax \citep{kaiser1959computer} may be applied to the stacked $[\bW_{J1}^\top,\dots,\bW_{JK}^\top]^\top$ matrices to improve the interpretability. Another approach is to apply ICA to the stacked loadings to obtain an orthogonal matrix that maximizes their non-Gaussianity similar to multiset CCA + joint ICA \citep{Sui2011DiscriminatingModel}. 

\subsection{Expectation-Maximization algorithm for ProJIVE}\label{subsec:EM.algorithm}
We now develop an Expectation-Maximization (EM) algorithm to estimate the variable loadings, noise variances, and subject scores.

Consider the latent subject scores, $\{\btheta_i = ( \bz_i^\top,\bb_{i1}^\top,\dots,\bb_{iK}^\top)^\top: i = 1, \dots , n\}$, as ``missing" data, so that the ``complete" data include the latent scores and the observed $\bx_i$. Let $\bepsilon_i=(\bepsilon_{i1}^\top,\dots,\bepsilon_{iK}^\top)^\top$ and  
\begin{align*}
\bW = 
        \begin{pmatrix} 
            \bW_{J1} & \bW_{I1} & \bzero   & \cdots & \bzero \\
            \vdots   & \bzero   & \ddots &  \bzero & \vdots \\
            \vdots & \vdots & \bzero & \ddots & \bzero \\
            \bW_{JK} & \bzero  & \cdots & \bzero & \bW_{IK} \\
        \end{pmatrix}.
\end{align*}
Then, equation \eqref{eqn:jive.ml.mod} is equivalent to $\bx_i=\bW \btheta_i + \bepsilon_i$.

Let $p=\sum_{k = 1}^K p_k$ represent the total number of variables/features in both datasets and $r = r_J + \sum_{k = 1}^K r_{Ik}$ the total number of latent components. Let $\bD$ be the diagonal matrix formed with $\bD_1,\dots,\bD_K$ on the block diagonals. Then, we can write the complete-data log likelihood as
\begin{equation}
\begin{split}
    \ell_C(\bW, \bD) = 
    &  -\frac{n}{2}\left\{(p+r)\log (2\pi) +\log|\bD|\right\}-\frac{1}{2} \sum_{i=1}^n \left\{(\bx_i - \bW \btheta_i)^\top\bD^{-1}(\bx_i -\bW \btheta_i) + \btheta_i^\top\btheta_i \right\}.\label{eqn:complete.dat.ll1}
\end{split}
\end{equation}
The definitions of the latent and observed variables lead to a multivariate normal distribution for the complete data vector $(\btheta_i^\top, \bx_i^\top)^\top \sim N(\mathbf{0},\bSigma)$ where, 
$\bSigma =  \begin{pmatrix} \bI_r & \bW^\top \\ \bW & \bC \end{pmatrix}$.

Define loadings and scores for block $k$ as $\bW_k = \left(\bW_{Jk}, \bW_{Ik}\right)$ and $\btheta_{ik} = (\bz_i, \bb_{ik})$ and  selection matrices $\bL_k = 
\left ( \mathbf{0}_{p_k \times p_1} \dots \bI_{p_k \times p_k} \dots \mathbf{0}_{p_k \times p_K} \right ) $ and 
\begin{align*}
\bM_k = \begin{pmatrix}
\bI_{r_J \times r_J} & \dots & \mathbf{0} & \dots & \mathbf{0}\\
\mathbf{0} & \dots & \bI_{r_{Ik} \times r_{Ik}} & \dots & \mathbf{0}
\end{pmatrix}. 
\end{align*}
With this parameterization, $\bx_{ik} = \bL_k \bx_i$, $\btheta_{ik} = \bM_k \btheta_i$, and $\bL_k \bW = \bW_k \bM_k$. The expectation of the complete log-likelihood, conditioned on the observed data, is 
\begin{equation}\label{eqn:e.step.2}
\begin{split}
 \mathbb{E}\{ \ell_C |\bx_i\} &= -\frac{n}{2}\left((p+r)\log(2\pi) + \sum_{k=1}^K \log [\tr(\bD_k) ]\right) \\ & - \frac{1}{2} \sum_{i=1}^n \sum_{k=1}^K \bD_k^{-1} \left [ \bx_{ik}^\top \bx_{ik} + \tr\{\bW^\top_k \bW_k \mathbb{E}(\theta_{ik}\theta_{ik}^\top|\bx_i)\} - 2 \bx_{ik}^\top \bW_k \mathbb{E} (\btheta_{ik}|\bx_i) \right ] + \mathbb{E} (\btheta_{ik}^\top \btheta_{ik}|\bx_i).
\end{split}
\end{equation}
The first and second conditional moments of the scores for block $k$ take the form 
\begin{align*} 
    \mathbb{E}(\btheta_{ik}|\bx_i) &= 
    \bM_k \bW^\top \bC^{-1} \bx_i,\\ 
    \Cov(\btheta_{ik}|\bx_i) &= 
    \bM_k (\bI_r - \bW^\top\bC^{-1} \bW)\bM_k^\top, \\
    \mathbb{E}(\btheta_{ik} \btheta_{ik}^\top|\bx_i) &= \bM_k (\bI_r - \bW^\top\bC^{-1} \bW + \bW^\top \bC^{-1} \bx_i \bx_i^\top \bC^{-1} \bW) \bM_k^\top .
\end{align*}
Differentiating the conditional expected log-likelihood with respect to the parameters $\bW_k$ and $\bD_k$, then setting each equal to 0 yields closed form solutions given by
\begin{equation}\label{eqn:EM.roots.2}
\begin{split}
    \widetilde \bW_k = & \left( \sum_i \bx_{ik} \mathbb{E}(\btheta^\top_{ik}|\bx_i)\right) \left( \sum_i \mathbb{E}(\btheta_{ik} \btheta^\top_{ik}|\bx_i) \right)^{-1}, \\
    \widetilde {\bD}_k = & \Diag \{\frac{1}{n} \sum_i \diag \{\bL_k  \bx_i\bx_i^\top \bL_k^\top + \bW_k\mathbb{E}(\btheta_{ik} \btheta_{ik}^\top|\bx_i)\bW_k^\top - 2\bW_k\mathbb{E}(\btheta_{ik}|\bx_i)\bx_{ik}^\top \}\},
\end{split}
\end{equation}
where $\diag\{\cdot\}$ extracts the diagonal from a square matrix and $\Diag\{\cdot\}$ constructs a diagonal matrix from the elements of its argument. Assuming $\bD_k = \sigma_k^2 \bI$, the second equation in \eqref{eqn:EM.roots.2} becomes $\widetilde \sigma_k^2 = \frac{1}{np_k} \sum_i \tr \{\bL_k  \bx_i\bx_i^\top \bL_k^\top + \bW_k\mathbb{E}(\btheta_{ik} \btheta_{ik}^\top|\bx_i)\bW_k^\top - 2\bW_k\mathbb{E}(\btheta_{ik}|\bx_i)\bx_{ik}^\top \}.$ We initialize $\bD_k=\hat \sigma_k^2 \bI$, where $\hat \sigma_k^2$ is the MLE from the rank $r_J+r_{Ik}$ pPCA of each dataset (average of the smallest $p_k - r_J - r_{Ik}$ eigenvalues for $n>p_k$). We initialized the loading matrices $\bW_{Jk}$ and $\bW_{Ik}$ using three strategies: (1) the AJIVE solution \citep{Feng2018Angle-basedExplained}; (2) the Cholesky decomposition on the covariance matrix of each dataset, where $W_{Jk}$ is initialized using the first $r_{J}$ columns and $W_{Ik}$ with the next $r_{Ik}$ columns; (3) standard normal entries. Our code on github provides additional options.

\section{Simulation study}
\label{sec:sims.pjive}

\subsection{Simulation design}
\label{sec:sims.pjive.methods}	
We conducted simulation studies to examine 1) the accuracy of estimates when the joint signal strength (i.e., proportion of variation) is low versus high and 2) robustness against model misspecification. For the centered data matrices such that $\bX_k \bm{1} = \bzero$, the proportion of total variation in the $k^{th}$ dataset attributable to the joint signal is $R^2_{Jk}=\frac{||\bJ_k||_F^2}{||\bX_k||_F^2}$. The proportion attributable to the individual signal is $R^2_{Ik}=\frac{||\bI_k||_F^2}{||\bX_k||_F^2}$. The individual ranks, sample size, number of features in $\bX_1$, and proportions of individual variation explained for both data blocks were held constant at $r_{I1}=r_{I2}=2$, $n=1000$, $p_1 = 20$, and $R_{I1}^2 = R_{I2}^2 = 0.25$, respectively.
We use a full factorial design with the following factors:
        \begin{enumerate}
            \item The joint rank: with levels (a) $r_J=1$ and (b)  $r_J=3$
            \item The number of features in $\bX_2$: with levels (a) $p_2 = 20$ and (b) $p_2 = 200$,
            \item Joint Variation Explained in $\bX_1$: with levels (a)  $R_{J1}^2 = 0.1$ and (b) $R_{J1}^2 = 0.5$,
            \item Joint Variation Explained in $\bX_2$: with levels (a) $R_{J2}^2 = 0.1$ and (b) $R_{J2}^2 = 0.5$,
            \item Data generating distributions: with levels (a) Gaussian scores and loadings and (b) mixture of Gaussian joint scores (mixing proportions equal to 0.2, 0.5, and 0.3 with means equal to -4, 0, and 4, respectively, and sd=1 for all three) and Rademacher loadings for both joint and individual components.
        \end{enumerate}
We define $\bW_{Jk} = \bR_{Jk}\bQ_{J}$, $\bW_{Ik} = \bR_{Ik}\bQ_I$ with diagonal matrices $\bQ_{J}=\diag(3,2,1)$ and $\bQ_I=\diag(2,1)$ and $\bR_{Jk}$ and $\bR_{Ik}$ with entries defined by factor five (Gaussian or Rademacher). Entries of the noise matrices $\bE_1$ and $\bE_2$ were randomly drawn from a standard Gaussian distribution.  In order to achieve the desired values of $R_{Jk}^2$ and $R_{Ik}^2$, we rescaled the joint and individual matrices such that $\bX_{k} = d_k \bJ_K + c_k \bA_k + \bE_k$ for appropriate constants $c_k$ and $d_k$, as described in the Web Appendix D.1. 

For each combination of settings, we performed 100 simulations using five methods of JIVE analysis: ProJIVE, AJIVE, GIPCA, R.JIVE, and D-CCA, using true signal ranks as input for each. In ProJIVE, the EM algorithm is implemented using the isotropy assumption and initiated using all three strategies discussed in Section \ref{subsec:EM.algorithm}. We evaluated accuracy using the chordal norm, a measure of distance between subspaces \citep{YeLim2014}, scaled by the rank of the subspace to result in a value contained in $[0,1].$ The chordal norm is defined in the Web Appendix D.2. 

We also conducted a simulation study based on the design from the \citet{Feng2018Angle-basedExplained} example. In this design, observational units in each half of the data block were given joint scores of -1 and 1, respectively. Joint loadings in $\bX_1$ have values 0 for half the variables and 1 for the other half. Joint loadings in $\bX_2$ have values 0 for 80\% of the variables and 1 for the remaining 20\%. The individual signal for $\bX_1$ comprised one component driven by all variables, which partitioned subjects into two groups. The individual signal for $\bX_2$ comprised two components. The loadings for each individual signal were orthogonal to the joint loadings, and the noise was standard Gaussian. Data were generated with sample size $n=100$, $p_1 = 100$, and $p_2 = 1,000$ features. Note that in \cite{Feng2018Angle-basedExplained}, $p_2 =10,000$. We also attempted to fit ProJIVE with $p_2=10,000$. However, we experienced issues with computational feasibility, and the algorithm did not converge.

\subsection{Simulation results}\label{sec:sims.pjive.results}

Figures \ref{fig:sim_gg_norms_rJ1} and \ref{fig:sim_gg_norms_rJ3} summarize chordal distances between true score and loading subspaces and their estimates when simulated data conform to Gaussian assumptions in ProJIVE (level (a) of factor 4) with $r_J=1$ and $r_J=3$ joint component(s), respectively. For the simulations with $p_2=20$ (sub-figures \ref{fig:sim_gg_norms_rJ1}A and \ref{fig:sim_gg_norms_rJ3}A), the joint scores were most accurate in ProJIVE across the $R_{Jk}^2$ settings, except in the small signal setting, $R_{J1}^2 = R_{J2}^2 = 0.1$ with $r_J=3$ joint components. 
ProJIVE estimates of individual scores were more accurate than those from other methods when signal proportions were mixed, i.e., $R_{J1}^2 \neq R_{J2}^2$ and at least as accurate as others when signal proportions were equal across data blocks. 

In the mixed-dimension setting ($p_2 = 200$; sub-figures \ref{fig:sim_gg_norms_rJ1}B and \ref{fig:sim_gg_norms_rJ3}B), ProJIVE achieves substantial improvement in estimating joint and individual scores and individual loadings, and moderate improvement or similar performance in joint loadings compared to other methods.

Figures \ref{fig:sim_mixrad_norms_rJ1} and \ref{fig:sim_mixrad_norms_rJ3} summarize chordal distances for simulations when joint scores were from a mixture of Gaussians and loadings from a Rademacher distribution (setting (b) of factor 4), with $r_J=1$ and $r_J=3$ joint component(s), respectively. For $r_J=1$ (figure \ref{fig:sim_mixrad_norms_rJ1}), ProJIVE estimates of scores and individual loadings were markedly more accurate than those from other methods when signal proportions differ and as accurate as other methods when signal proportions are equal across data sets. Chordal distances between estimated and true subspaces tended to be larger when $r_J=3$ in these settings. However, ProJIVE estimates of joint and individual scores were generally the most accurate (figure \ref{fig:sim_mixrad_norms_rJ3}). We note that joint scores and joint loadings from all methods were inaccurate when $R_{J1}^2=R_{J2}^2=0.1$ and $p_2 = 20$. Yet, ProJIVE improves when $p_2 = 200$. In particular, when $p_2=200$ and $R_{J2}^2=0.5$, ProJIVE leverages the information in the second dataset, and the joint scores have substantially lower chordal norms than the other methods (figure \ref{fig:sim_mixrad_norms_rJ3}). Overall, the loadings are estimated with greater or similar accuracy in ProJIVE than in other methods. However, the discrepancies between ProJIVE-Oracle and ProJIVE indicate that our method is sensitive to initial values. These results suggest that ProJIVE may still be preferred even when the data are not generated from the ProJIVE model \eqref{eqn:jive.ml.mod}. 

{In the simulations based on the \citet{Feng2018Angle-basedExplained} design, ProJIVE estimates of scores had lower chordal norms than those from other methods, and ProJIVE-estimated loadings had smaller or equal chordal norms compared to other methods (Figure \ref{fig:sim_feng}).}

\section{Joint Analysis of Brain Morphometry and Cognition in ADNI Data}\label{sec:data}
Our data application examines shared variability in measures of cognition/behavior and brain morphometry. Data were obtained from The Alzheimer's Disease Prediction Of Longitudinal Evolution (TADPOLE) Challenge (\url{https://tadpole.grand-challenge.org/}). The TADPOLE Challenge was an open competition to predict the onset of symptomatic AD in the short to medium term in an age group at risk for AD. We apply ProJIVE to data from $n=587$ participants for whom the full battery of cognitive and behavioral assessments were available. This battery includes 22 scales and sub-scales from cognitive and behavioral assessments given to participants in the ADNI-GO and ADNI-2 phases of the study. Additional information is in Web Appendix E.1, including Table S.1 summarizing age, gender, and ApoE4 (0=negative, 1=heterozygous, and 2=homozygous) stratified by diagnosis.

\subsection{Cognition}
Cognitive and behavioral measures included the following assessments, each of which is represented by a single variable except where noted: 1) Clinical Dementia Rating - Sum of Boxes (CDR-SB) (higher=worse cognition); 2)  Alzheimer's Disease Assessment Scale - Cognition (ADAS), in which the 11-item and 13-item scores were used as separate variables (higher=worse cognition); 3) the Mini-Mental State Exam (MMSE) (higher=better cognition); 4) Rey's Auditory Verbal Learning Test (RAVLT), in which Forgetting, Immediate, and Learning sub-scales were used as separate variables (higher=better cognition); 5) Montreal Cognitive Assessment (MOCA) (higher=better cognition); and 6) Everyday Cognition (ECOG) \citep{farias2008measurement}, in which seven pairs of sub-scales were used (higher=worse cognition). Summary statistics for eachof the $p_1=22$ cognition measures used in JIVE analyses are shown in Web Appendix Table S.2. Some cognitive measures were used to inform diagnoses (e.g., ADAS13, MOCA, and MMSE). 

\subsection{Brain morphometry}
We used brain morphometry measures in the TADPOLE data, which were preprocessed using the cross-sectional Freesurfer pipeline \citep{reuter2010highly}. The brain morphometry dataset contains measures of cortical thickness (CT), cortical surface area (SA), and cortical volume (CVol) for each of the thirty-four Desikan cortical surface ROIs  \citep{desikan2006automated}, as well as the volumes for the following: eight subcortical ROIs in each hemisphere, the brainstem, ventricles, cerebrospinal fluid, optic chiasm, corpus callosum, and total intracranial volume. We refer to all non-cortical regions as ``sub-cortical'' (SVol). In total, 245 measures of brain morphometry are included in ProJIVE and as a single dataset.  

We also repeated the analysis, treating each set of measures as a separate dataset (CT, SA, CVol, and SVol), as described in Section \ref{sec:fivedatasets}. 

\subsection{Dimension reduction, preprocessing, and summary} \label{subsec:data.pjive.preproc}

For each feature (e.g., brain morphometry or cognition measure), we fit a linear regression with age and sex as predictors, extracted the residuals, and then standardized to have unit variance. We then inspected scree plots and selected $r_{Cog}=5$ total components in the cognition dataset and $r_B=12$ in the brain morphometry datasets (Web Appendix Figure S.1). The relatively small ranks are based on our previous experience that larger ranks can create issues when selecting the joint rank.

\subsection{Joint subspace}\label{subsec:pjive.jnt.scores}
A permutation test based on permuting the participants in the second dataset to generate a null distribution of the canonical correlations between principal component scores \citep{murden21CJIVE} resulted in one joint component, and the test of principal angles from AJIVE also resulted in one component \citep{Feng2018Angle-basedExplained}. 

\subsubsection{Variable loadings}
Joint loadings for a dataset indicate the extent to which each measure in that dataset contributes to the shared subspace. To aid interpretation, we scaled joint loadings by the reciprocal of the maximum absolute value across data sets. We focus on the top 10 largest (in absolute value) cognition loadings and the upper $90^{th}$ percentile of brain loadings. 

Figure \ref{fig:joint.loads.all} shows the most prominent cognition loadings include ADAS, MMSE, and MOCA, which were used for diagnosis. Lower MOCA and MMSE scores and higher ADAS, CDRSB, and ECOG scores represent worse cognitive performance. Their loadings point in opposite directions. 

Measures of cortical thickness (CT) and subcortical volume (SVol) were prominent in morphometry. The loadings for subcortical volumes in the upper $90^{th}$ percentile of all loadings occur within left-right pairs, including the hippocampus (Figure \ref{fig:joint.loads.all}). Results were similar for left-right cortical thickness pairs, including the entorhinal cortex (Figure \ref{fig:joint.loads.brain}). No surface area measurements were in the upper $90^{th}$ percentile. Brain loadings were consistent with associations between specific ROIs and AD found in the literature. The atrophy of gray matter structures, including the hippocampus, amygdala, and entorhinal cortex increases the risk of AD \citep{NestorVentriclesAndAD2008, frisoni2009vivo, poulin2011amygdala}. The ventricles are not gray matter structures, and larger ventricles indicate greater atrophy. Consistent with this biology, they load in the opposite direction of the gray matter regions. 

\subsubsection{Subject scores}
We examined the relationship between subject scores and genetic risk factors (ApoE4), diagnosis, and PET biomarkers (AV45 and FDG). The location of the distribution of joint scores increased in ApoE4=1 relative to ApoE4=0, and then there was a small increase in ApoE4=2 relative to ApoE4=1, although in all cases we see overlapping distributions (Figure \ref{fig:joint.scores}). Using multinomial logistic regressions, joint subject scores were significantly related to the odds of having the ApoE4 allele (z=5.5{2}, $p<10^{-7}$ for 1 versus 0 and z=4.{7}8, $p<10^{-5}$ for 2 versus 0 allele counts, McFadden's pseudo-R$^2$ = 0.03). 

For diagnosis, we also see an increase in the location of the distribution of joint scores for CN versus MCI, and then a more substantial increase from MCI to AD with non-overlapping interquartile ranges. Joint scores were significant in the multinomial regression (z=6.43, $p<10^{-9}$, and  z=11.40, $p<10^{-10}$ for MCI versus CN and AD versus CN, respectively, McFadden's pseudo-R$^2$=0.20). 
These results indicate that the joint subject scores capture variation shared between cognition and brain morphometry, which is significantly associated with diagnoses.  

Lastly, joint subject scores were significantly associated with biomarkers derived from AV-45 PET (t=10.03, $p<10^{-15}$, $R^2=0.15$) and FDG PET (t=-15.50, $p<10^{-15}$, $R^2=0.29$) using simple linear regression (Figure \ref{fig:joint.scores}). Previous studies have shown that the accumulation of beta-amyloid as measured by AV-45 PET increases the risk for AD \citep{camus2012AV45PET}, while lower FDG levels are associated with dementia \citep{langbaum2009FDGPET}. PET imaging utilizes a radioactive tracer, which is more expensive and invasive than MRI. These results indicate that a significant portion of the information captured in PET biomarkers is contained in the joint subject scores of MRI and cognition data. 


\subsection{Individual subspaces}

\subsubsection{Variable loadings}

Loadings for the first two components of the individual subspace of cognition were dominated by ECOG measures, with primarily self-reported measures for the first component and study partner-reported measures for the second (Web Appendix Figure S.2). RAVLT-immediate, RAVLT-learning, ADAS13, and ADAS11 loaded most strongly onto the third component, while ECOG self-reported memory dominated the fourth component. The 90$^{th}$ percentile of loadings for the first individual component of brain morphometry mainly consisted of surface area measures, which were not prominent in the joint subspace. Additional components are discussed in Web Appendix E.3 and displayed in Web Appendix Figure S.3.

\subsubsection{Subject scores}
Scores from the first three individual cognition components were associated with diagnosis, with the third component exhibited the strongest association (z=8.79, $p<10^{-8}$, and  z=-5.02, $p<10^{-10}$, respectively, for MCI versus CN and AD versus CN, adjusted pseudo-R$^2$=0.13). The third individual cognition scores were also associated with ApoE4 (z=3.57, $p<10^{-3}$ for 1 versus 0 allele counts and z=-6.32, $p<10^{-9}$ for 2 versus 0 allele counts, adjusted pseudo-R$^2$=0.03), and most strongly associated with AV45 ($t=8.82, p<10^{-10}, R^2=0.12$) and FDG ($t=-6.23, p<10^{-9}, R^2=0.06$). On the other hand, no individual brain morphometry scores were associated with diagnosis, ApoE4 allele counts, or AV45, while the seventh individual morphometry scores were associated with FDG ($t = -4.1, p<10^{-4}, R^2=0.03$ ). 

\subsection{Alternative specification with five datasets}\label{sec:fivedatasets}

We conducted a similar analysis treating each collection of brain morphometry measures (i.e., CT, CVol, SA, and SVol) as a separate dataset, resulting in $K=5$ datasets. Scree plot analyses were again employed to choose the total signal rank for each data set, which resulted in ranks of 5 for cognition, 10 for cortical thickness, 9 for cortical volume, 12 for surface area, and 10 for subcortical volume. The joint rank was chosen as $r_J=1$ to facilitate comparison to the analysis with $K=2$ datasets. We focus on the joint subspaces here and provide additional information about individual subspaces in the Web Appendix.

Joint subject scores from the $K=5$ setting had a similar association to ApoE4 and diagnosis as in the $K=2$ case, with McFadden's pseudo-$R^2$ values of 0.04 and 0.21, respectively. Similarly, associations between joint scores and PET biomarkers were similar to those in the $K=2$ case, with pseudo-$R^2$ values of 0.15 for AV45 and 0.26 for FDG (Web Appendix Figure S.7).

\section{Discussion}\label{sec:discussion}
We propose ProJIVE, a statistical model and novel EM algorithm for data integration that extends probabilistic PCA \citep{tipping1999probabilistic} to multiple datasets. In probabilistic PCA, there is an equivalence between the MLE and the principal subspace of classic PCA. In contrast, ProJIVE leads to novel estimators distinct from AJIVE, R.JIVE, and other data integration methods. Our simulations demonstrate that ProJIVE improves estimation of joint subject scores. ProJIVE assumes mutual orthogonality between joint and individual subject scores. Arguably, defining orthogonality on these random variables may be easier to understand for some people than assumptions on row spaces. We analyze brain morphometry and cognition in ADNI datasets. The joint subject scores were associated with ApoE4, diagnosis, and PET biomarkers. The largest joint variable loadings show patterns that are consistent with associations found in the literature between measures of cognition, brain morphometry, and Alzheimer's disease \citep{kim2015apolipoprotein, piers2018structural, SULTANA2024102414}. On the other hand, we find that no individual brain morphometry components were related to ApoE4 or diagnosis. This suggests the relationship between ApoE4 and brain morphometry, and the relationship between diagnosis and brain morphometry, is due to shared variation in the brain morphometry and cognition data. We found that each of the individual cognition components were related to diagnosis, which suggests information pertinent to diagnosis is partly, but not entirely, captured by brain morphometry. 

ProJIVE has a number of limitations. In simulations, the computational costs were greater than AJIVE and D-CCA, but less than R.JIVE and GIPCA in the mixed dimension settings (i.e., $p_2=20$). AJIVE and D-CCA run times were typically less than one second when ranks were specified. Moreover, our current implementation of ProJIVE in R was not computationally feasible for the simulation setting with $p_2=10,000$ (see Section 3.1). While ProJIVE run times were often slower than R.JIVE and GIPCA in the high-dimensional settings, ProJIVE estimates tended to be more accurate than other methods. Additionally, we used scree plots to determine the total ranks of each dataset and a permutation test to determine the joint rank. Since we define a likelihood, one can use information criteria to determine ranks. The ProJIVE R implementation also calculates the Akaike Information Criterion (AIC) and Bayesian Information Criterion (BIC), which can be used to guide total and joint rank choices. However, this involves all combinations of $r_J$, $r_{Ik}$ for $k=1,\dots,K$, which is challenging at best. 

Care must also be given to the decision of what constitutes a ``dataset." One view is outlined in  \citet{zhu2020generalized}, who suggest defining a dataset as a collection of variables that all follow a similar probability distribution. In our framework, all scores are assumed to be Gaussian, although our simulations include settings where ProJIVE is robust to this assumption. We recommend careful consideration of the scientific impact of the datasets' definitions, for example, through evaluating multiple approaches. Treating the brain morphometry as a single dataset, as presented in the main analysis (Figures \ref{fig:joint.loads.all}, \ref{fig:joint.loads.brain}, \ref{fig:joint.scores}), or treating each grouping as a separate dataset, as presented in the Web Appendix (Figures S.6 and S.7), had relatively minor impacts on the joint scores.

Future work could extend ProJIVE to non-Gaussian settings. In independent factor analysis, the source distributions are modeled using a mixture of Gaussians \cite{attias1999independent}. The independent factor analysis likelihood could be extended to joint and individual components. However, the likelihood is difficult to maximize. Another direction for future research is to model observations from multiple datasets as a mixture of ProJIVE models. An EM algorithm could be developed for this setting, which would include latent variables for class membership. This would enable high dimensional clustering.

\section*{Acknowledgments}
This work uses the TADPOLE data sets \url{https://tadpole.grand-challenge.org} constructed by the EuroPOND consortium http://europond.eu funded by the European Union’s Horizon 2020 research and innovation programme under grant agreement No 666992. Data used in preparation of this article were obtained from the Alzheimer's Disease Neuroimaging Initiative (ADNI) database (\url{http://adni.loni.usc.edu}). As such, the investigators within the ADNI contributed to the design and implementation of ADNI and/or provided data but did not participate in the analyses or writing of this report. A complete listing of ADNI investigators can be found at: \url{http://adni.loni.usc.edu/wp-content/uploads/how_to_apply/ADNI_Acknowledgement_List.pdf}. The authors thank members of the Center for Biomedical Imaging Statistics (CBIS) at Emory University. This study was supported by R21AG066970 (B.R. and D.Q.), R21AG064405 (D.Q.), R01AG072603 (D.Q.), R01MH129855 (B.R.), and an ADRC pilot grant (D.Q. and B.R.) from parent grant P30AG066511. 

\section*{Declaration of Interest}

No authors have conflicting interests.

\section*{Web Appendix}
\label{sec:supp.material}

Supporting information referenced in Sections 2, 4, and 5 is available with this paper on the Journal of Computational and Graphical Statistics website.
R code reproducing simulations is available at \url{https://github.com/thebrisklab/ProJIVE}.

\bibliographystyle{apalike}
\bibliography{refs}

 

\newpage
\section*{Figures}

\begin{figure}[htbp]
    \centering
    \includegraphics[width = \textwidth]{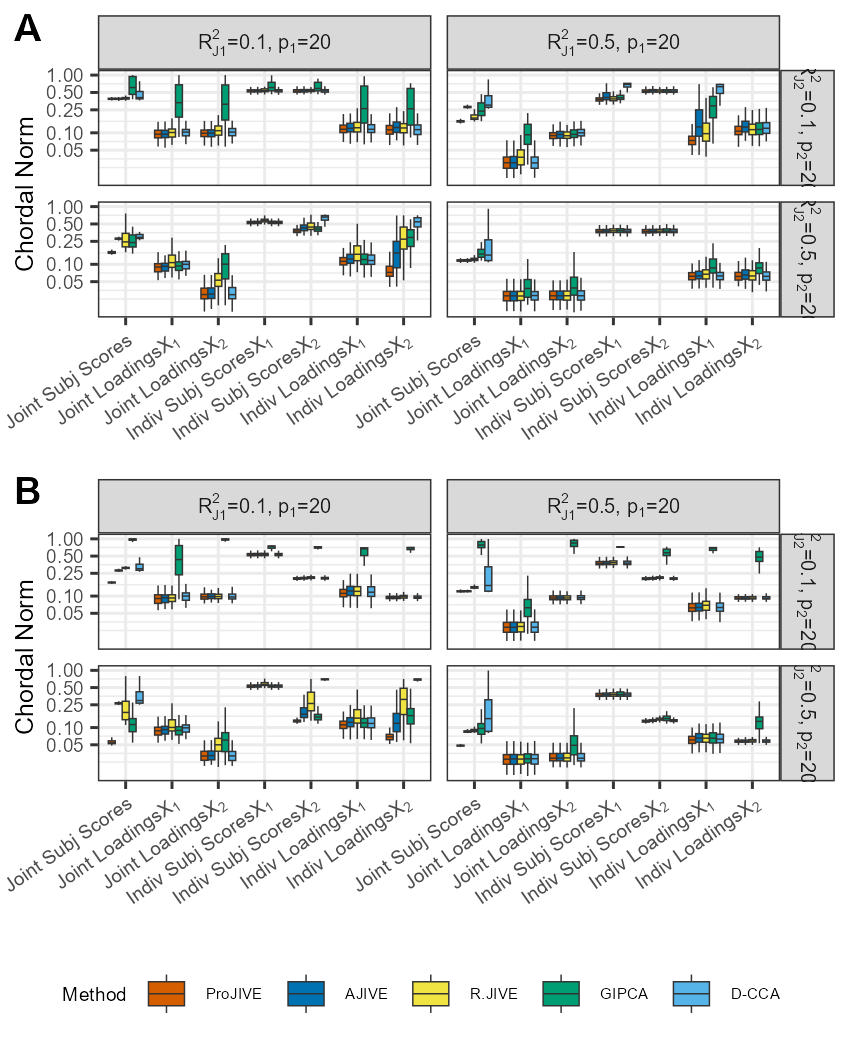} 
    \caption{}
    \label{fig:sim_gg_norms_rJ1}
\end{figure}    
\pagebreak
\clearpage

\begin{figure}[htbp]
    \centering
    \includegraphics[width = \textwidth]{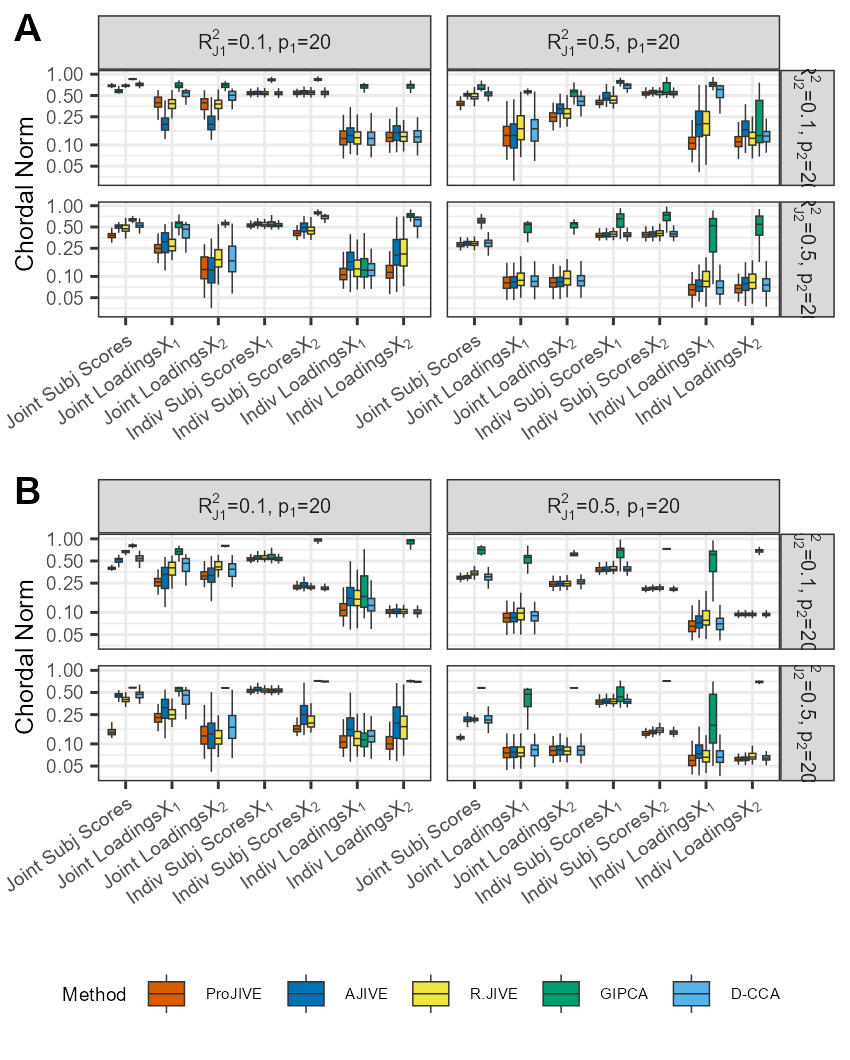}
    \caption{}
    \label{fig:sim_gg_norms_rJ3}
\end{figure}    
\pagebreak
\clearpage

\begin{figure}[htbp]
    \centering
    \includegraphics[width = \textwidth]{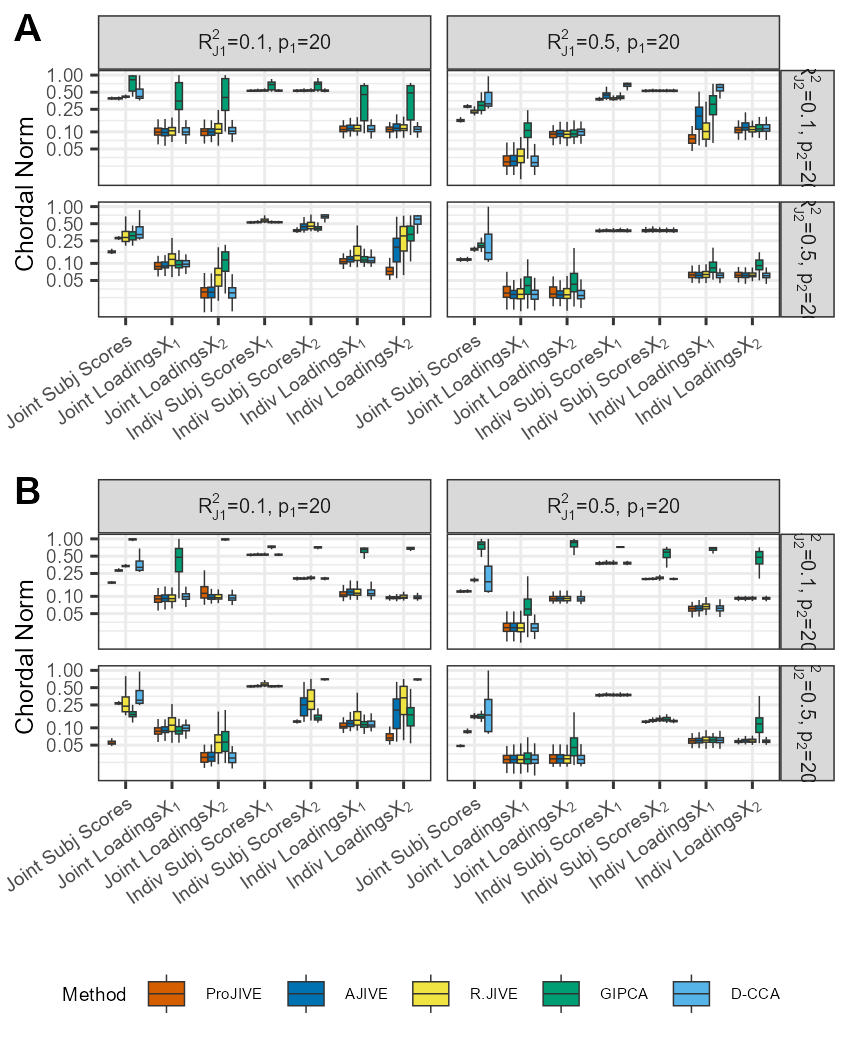}
    \caption{}
    \label{fig:sim_mixrad_norms_rJ1}
\end{figure}    
\pagebreak
\clearpage

\begin{figure}[htbp]
    \centering
    \includegraphics[width = \textwidth]{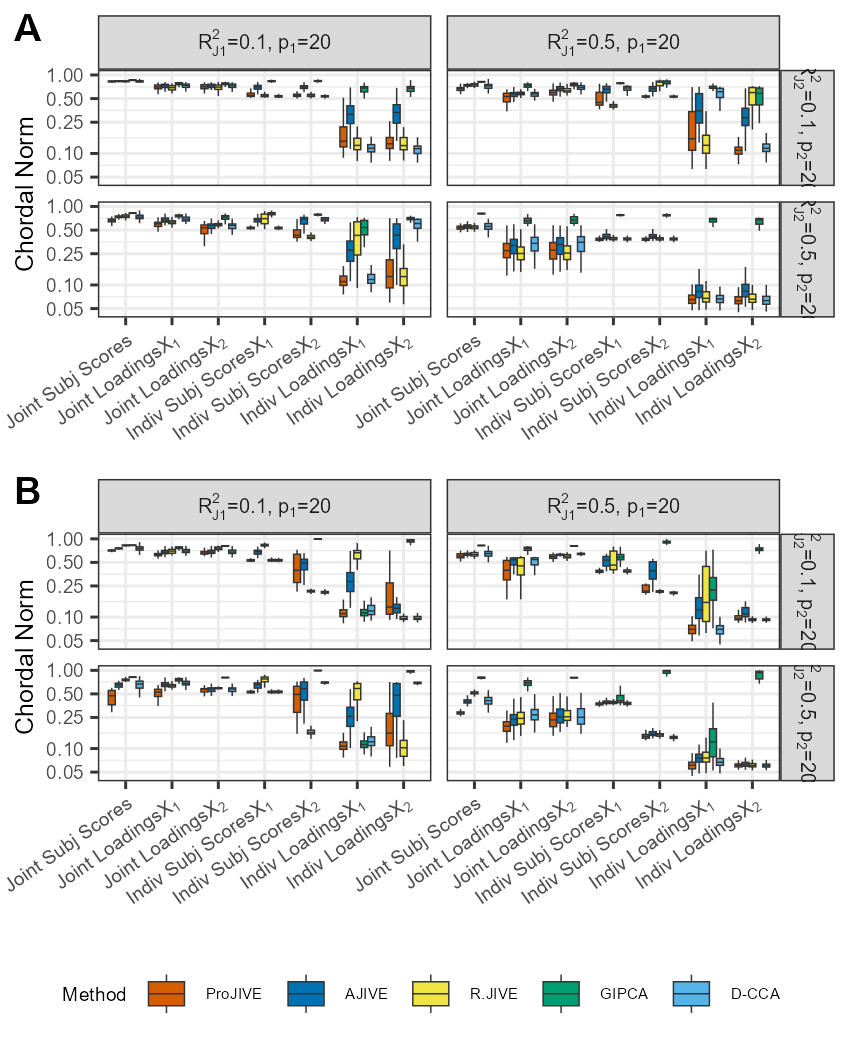}
    \caption{}
    \label{fig:sim_mixrad_norms_rJ3}
\end{figure}    
\pagebreak
\clearpage

\begin{figure}[htbp]
    \centering
    \includegraphics[width = \textwidth]{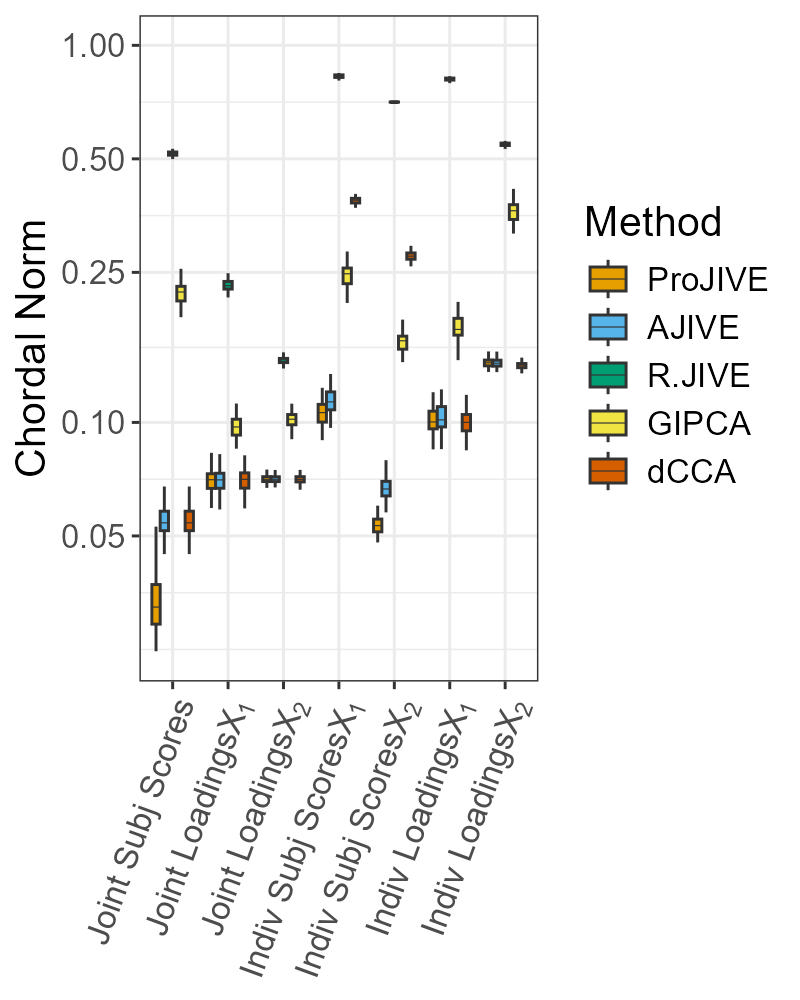}
    \caption{}
    \label{fig:sim_feng}
\end{figure}    
\pagebreak
\clearpage

\begin{figure}[htbp]
    \centering
    \includegraphics[width = \textwidth]{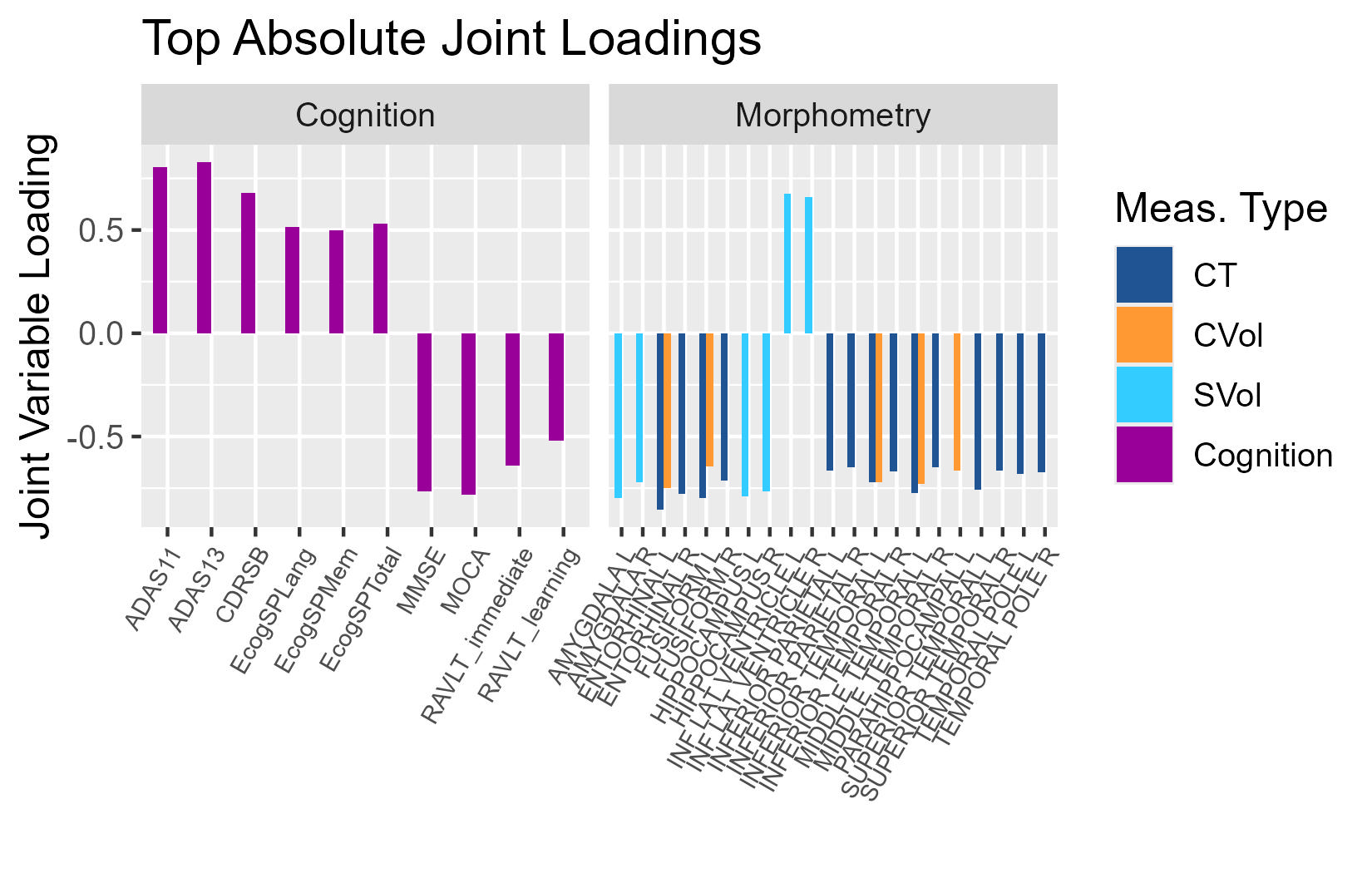}
    \caption{}
    \label{fig:joint.loads.all}
\end{figure}    
\pagebreak
\clearpage

\begin{figure}[h]
    \centering
    (a) \includegraphics[width = 0.75\textwidth]{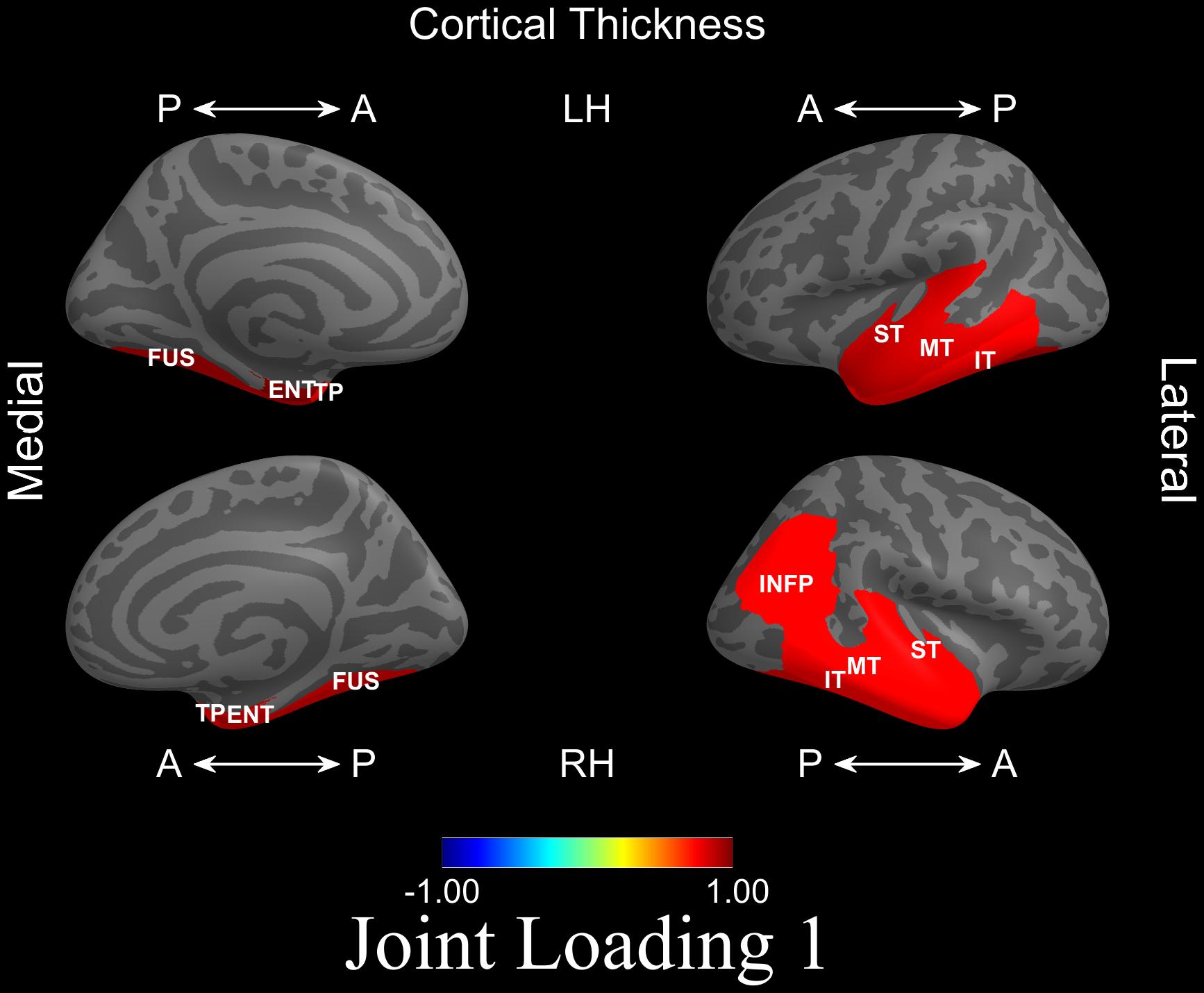}  \\
    (b) \includegraphics[width = 0.75\textwidth]{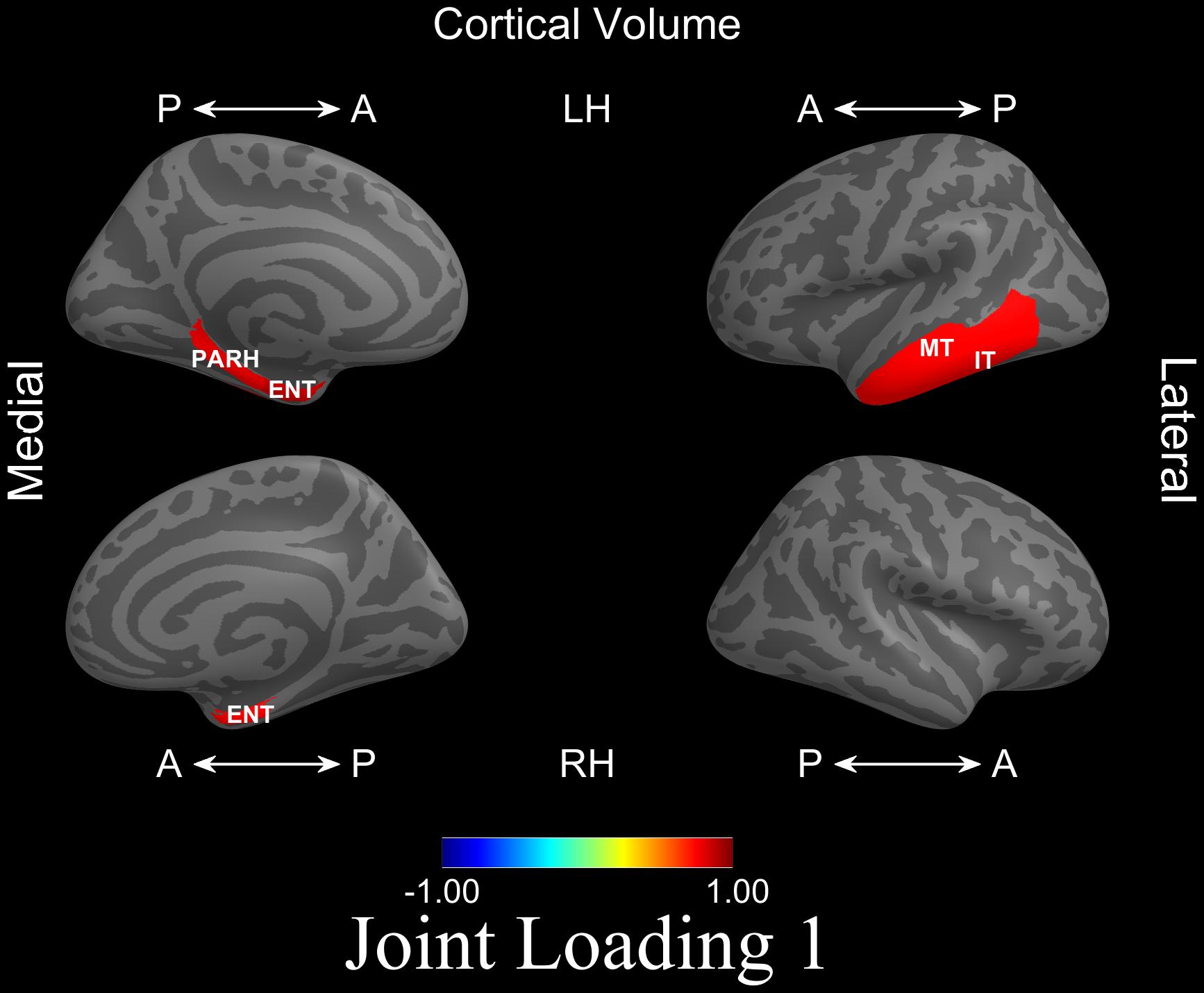} 
    \caption{}
    \label{fig:joint.loads.brain}
\end{figure}    
\pagebreak
\clearpage

\begin{figure}[h]
    \centering
    (a) \includegraphics[width=0.35\textwidth]{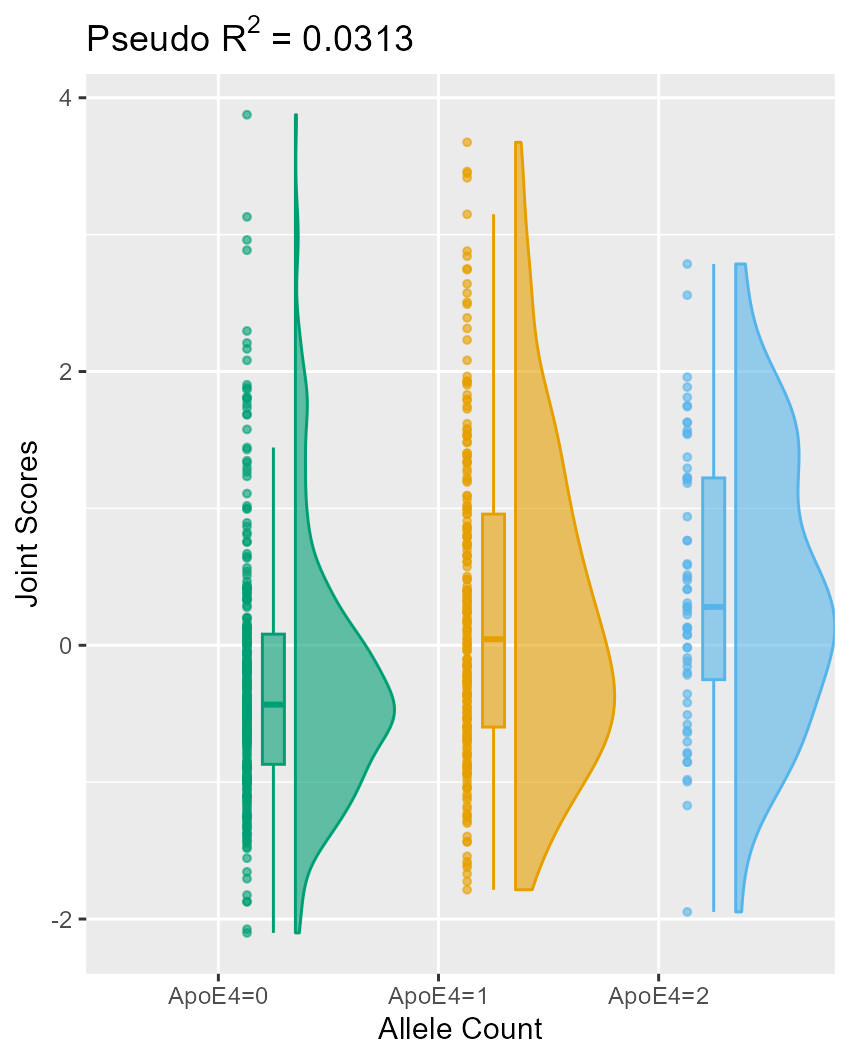} 
    (b) \includegraphics[width=0.35\textwidth]{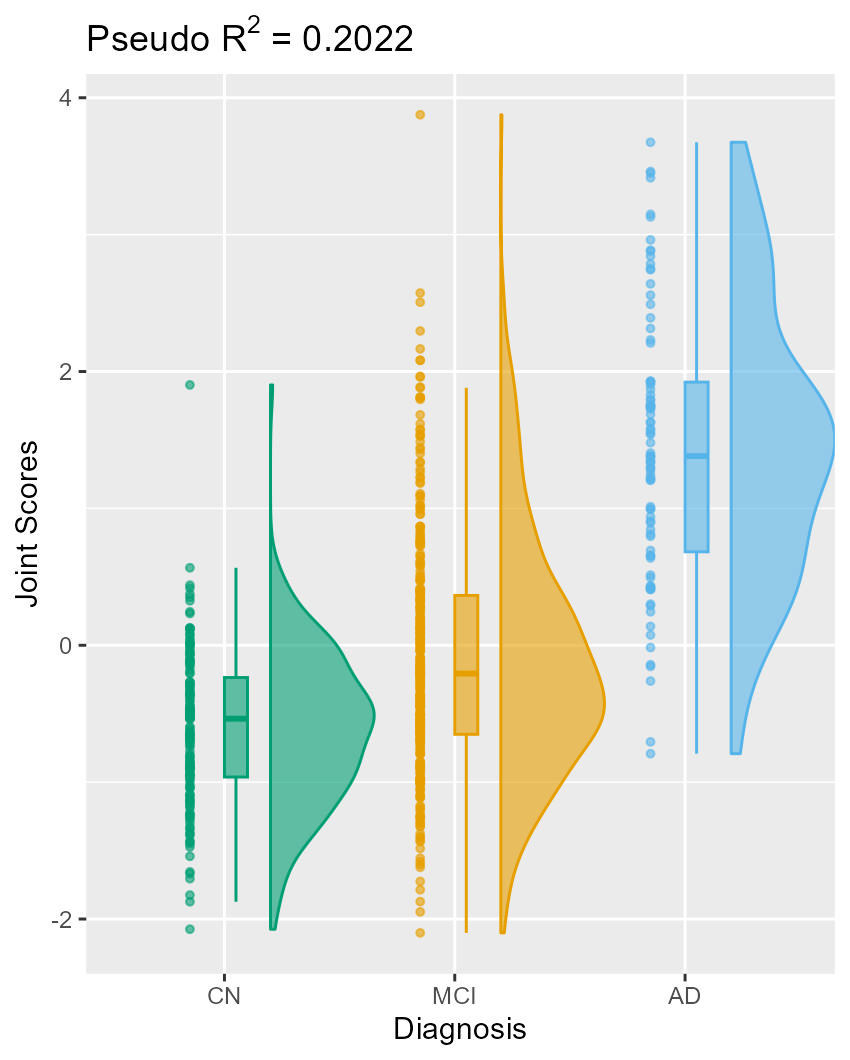} \\
    (c) \includegraphics[width=0.35\textwidth]{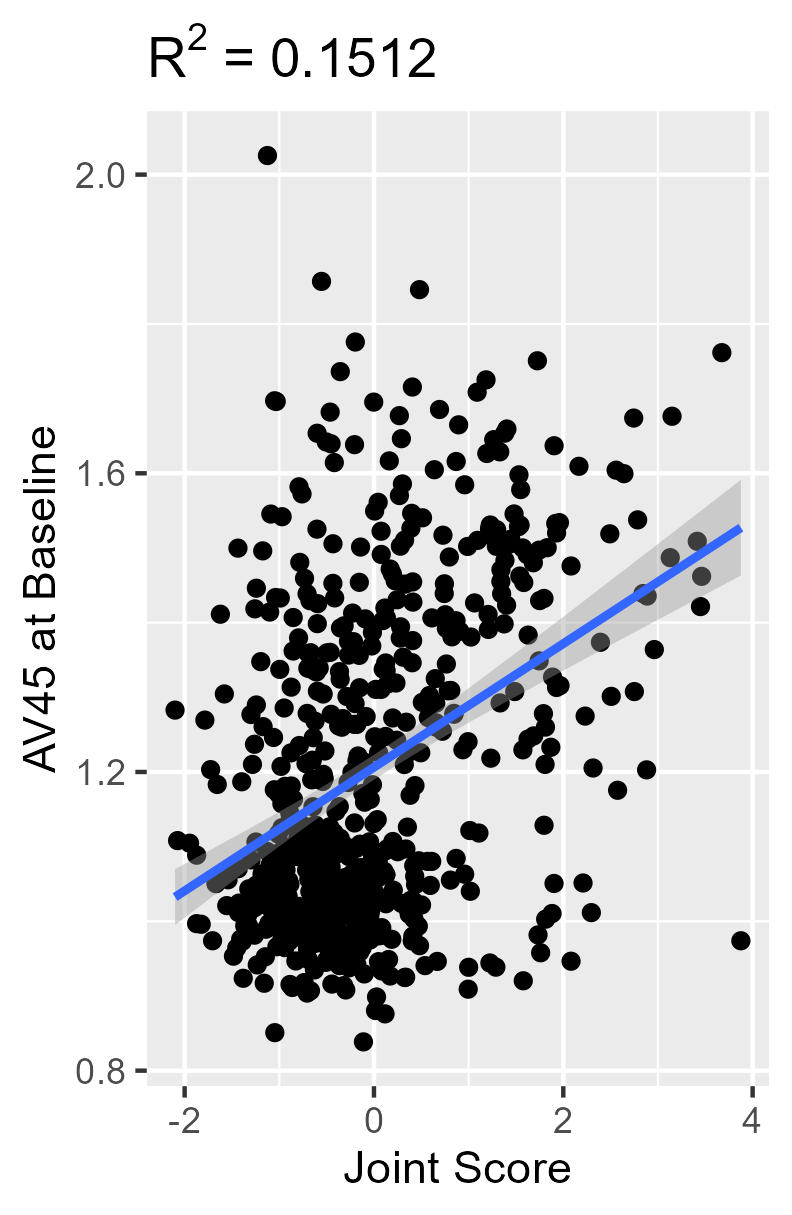}
    (d) \includegraphics[width=0.35\textwidth]{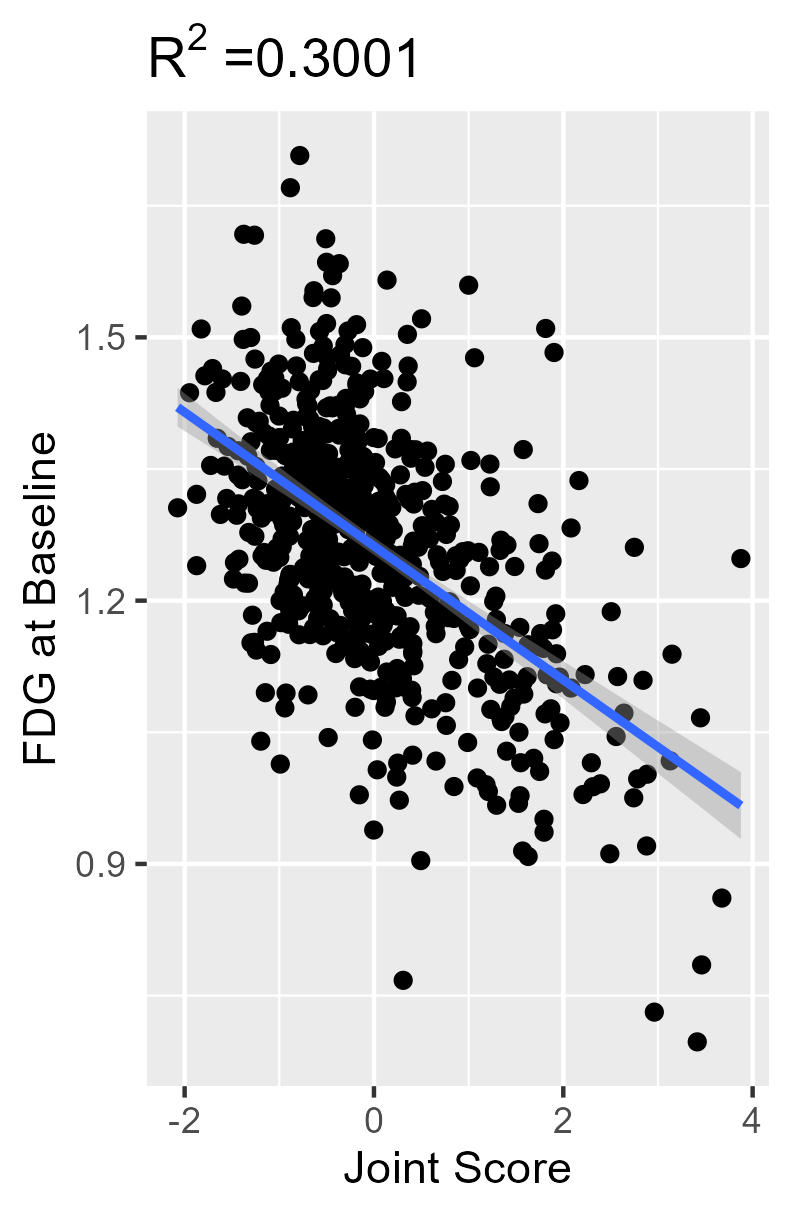} 
    \caption{}
    \label{fig:joint.scores}
\end{figure}    
\pagebreak
\clearpage

\section{Figures Captions}
\begin{itemize}
    \item Figure \ref{fig:sim_gg_norms_rJ1}: Interquartile and 95\% intervals for chordal distances between estimated and true subspaces with data generated from the ProJIVE model with $r_J=1$ joint component, $n=1000$ observational units, and $p_2 = 20$ and $p_2=200$ in sub-figures A and B, respectively.
    \item Figure \ref{fig:sim_gg_norms_rJ3}: Interquartile and 95\% intervals for chordal distances between estimated and true subspaces with data generated from the ProJIVE model with $r_J=3$ joint components, $n=1000$ observational units, and $p_2 = 20$ and $p_2=200$ in sub-figures A and B, respectively.
    \item Figure \ref{fig:sim_mixrad_norms_rJ1}: Interquartile and 95\% intervals for chordal distances between estimated and true subspaces where data are generated using joint subject scores from a mixture of Gaussian distributions, individual subject scores from standard Gaussian distributions, and variable loadings (joint and individual) from Rademacher distributions with  A) $p_2 = 20$, and B) $p_2=200$. $n=1000$ in all settings.
    \item Figure \ref{fig:sim_mixrad_norms_rJ3}: Interquartile and 95\% intervals for chordal distances between estimated and true subspaces where data are generated using joint subject scores from a mixture of Gaussian distributions, individual subject scores from standard Gaussian distributions, and variable loadings (joint and individual) from Rademacher distributions with  A) $p_2 = 20$, and B) $p_2=200$. $n=1000$ in all settings.
    \item Figure \ref{fig:sim_feng}: Interquartile and 95\% intervals for chordal distances between estimated and true subspaces using scores and loadings based on the simulation design in \citet{Feng2018Angle-basedExplained} with $n=100$, $p_1 = 100$, and $p_2 = 1,000$.
    \item Figure \ref{fig:joint.loads.all}: Variable loadings from the ten most extreme joint cognition loadings and upper $90^{th}$ percentile of the absolute value of joint brain loadings estimated via ProJIVE.
    \item Figure \ref{fig:joint.loads.brain}: Brain loadings occurring in the upper 90th percentile of the absolute value of the loadings for (a) cortical thickness and (b) cortical volume.
    \item Figure \ref{fig:joint.scores}: Joint subject scores estimated via ProJIVE show separation by (a) the count of ApoE4 allele counts and (b) diagnosis via raincloud plots. (c) Scatter-plot of joint subject scores and AV45 at baseline with OLS regression line and confidence interval. (d) Scatter-plot of joint subject scores and FDG at baseline with OLS regression line and confidence interval.
\end{itemize}

\end{document}


\title{Supporting Information for Probabilistic Joint and Individual Variation Explained for Data Integration}
\author[1]{Raphiel J. Murden}
\author[1]{Ganzhong Tian}
\author[2]{Deqiang Qiu}
\author[1]{Benjamin B. Risk\footnote{brisk@emory.edu, 1518 Clifton Rd NE, Atlanta, GA 30322}}

\affil[1]{Department of Biostatistics and Bioinformatics, Rollins School of Public Health, Emory University, 1518 Clifton Rd NE, Atlanta, GA 30322, USA 
}
\affil[2]{Department of Radiology and Imaging Sciences, Emory University School of Medicine, 1364 Clifton Rd NE, Atlanta, GA  30322, USA}

\maketitle

\appendix

\linenumbers
\doublespacing
\section{Proof of Theorem 1}\label{wa:thm:1}

\begin{thm}
Suppose $K=2$ and let $f_\Phi$ define the multivariate normal density according to the parameters $\{\bW_{J1},\bW_{J2},\bW_{I1},\bW_{I2},\;\bD_1 = \sigma_1^2 \bI,\;\bD_2  = \sigma_2^2 \bI\}$ and the assumptions in (1) of the main manuscript. Let $f_{\Phi^*}$ denote the multivariate normal density with parameters $\{\bW_{J1}^*,\bW_{J2}^*,\bW_{I1}^*,\bW_{I2}^*,\bD_1^{*} = \sigma_1^{*2} \bI,\bD_2^{*} = \sigma_2^{*2} \bI\}$ and the assumptions in (1) of the main manuscript. Then $f_\Phi = f_{\Phi^*}$ if and only if 
\begin{enumerate}
    \item $\sigma_1^{*2}=\sigma_1^2,\;\sigma_2^{*2}=\sigma_2^{2}$, 
    \item There exists an $r_J \times r_J$ non-singular matrix $\bT_1$ such that 
    \begin{enumerate}
    \item $\bW^*_{J1} = \bW_{J1} \bT_1$ and $\bW^*_{J2} = \bW_{J2} (\bT_1^{-1})^\top$,
    \item $\bT_1$ satisfies
    \begin{align}
\bW_{I1} \bW_{I1}^\top + \bW_{J1}(\bI - \bT_1 \bT_1^\top )\bW_{J1}^\top \succeq 0,\label{eq:wspart1} \\ 
\bW_{I2} \bW_{I2}^\top + \bW_{J2}(\bI - (\bT_1^{-1})^\top \bT_1^{-1})\bW_{J2}^\top \succeq 0 \label{eq:wspart2},
\end{align}
\end{enumerate}
\item $\bW_{I1}^*$ is defined such that
\begin{linenomath}
\begin{align}
\bW_{I1}^* (\bW_{I1}^*)^\top = \bW_{I1} \bW_{I1}^\top + \bW_{J1}(\bI - \bT_1 \bT_1^\top )\bW_{J1}^\top,\label{eq:wspart3.1}
\end{align}
\end{linenomath}

$\bW_{I2}^*$ is defined such that
\begin{linenomath}
\begin{align}
\bW_{I2}^* (\bW_{I2}^*)^\top = \bW_{I2}\bW_{I2}^\top + \bW_{J2}(\bI - (\bT_1^{-1})^\top \bT_1^{-1})\bW_{J2}^\top.\label{eq:wspart3.2}
\end{align}
\end{linenomath}
\end{enumerate}\label{thm:wskequal2}
\end{thm}
\begin{proof}
\textcolor{blue}{The proof builds on results from \cite{Browne1979TheAnalysis}.} It is straightforward to verify that conditions (1)--(3) result in $f_\Phi = f_{\Phi^*}.$

To show $f_\Phi = f_{\Phi^*}$ implies (1)--(3), let $\bC^*$ denote the covariance for a ProJIVE model with parameters $\{\bW_{Jk}^*, \bW_{Ik}^*, \; \sigma_k^{*2},\;k=1,2\}$. We need to show $\bC^* = \bC$ implies (1)--(3). 

First, we show $f_\Phi = f_{\Phi^*}$ implies (1). Define $\bC_k = \Cov(\bx_{ik})$. Since $\bC_k$ is positive definite, we can write its eigenvalue decomposition: $\bC_k = \bU \bLambda \bU^\top$, which has $p_k$ non-zero eigenvalues. Since by assumption the noise is isotropic and rank$([\bW_{Jk},\bW_{Ik}])<p_k$, the smallest eigenvalue is equal to $\sigma_k^2$. Let $\bC_k^*= \Cov(\bW_{Jk}^* \bz_i^* + \bW_{Ik}^* \bb_{ik}^* + \bepsilon_{ik}^*)$, where we again assume rank$(\bW_{Jk}^*+\bW_{Ik}^*)<p_k$. Then $\sigma_k^{*2}$ is the smallest eigenvalue. Since $\bC_k=\bC_k^*$, the same eigenvalue decomposition applies, and we have $\sigma_k^{*2} = \sigma_k^2$.

Next, we will show $f_\Phi = f_{\Phi^*}$ implies (2a). Define $\bC_{12}=\Cov(\bx_{i1},\bx_{i2})$. $\bC_{12} = \bW_{J1} \bW_{J2}^\top$, and $\bC_{12}^* = \bW_{J1}^* (\bW_{J2}^*)^\top$. By assumption, rank$(\bW_{Jk})=r_J$ for $k=1,2$, and also rank$(\bW_{Jk}^*)=r_J$. Since rank($\bW_{J2}^\top)=r_J$, rank$(\bW_{J1} \bW_{J2}^\top) = \textrm{rank}(\bW_{J1})=r_J$. Both $\bW_{J1}$ and $\bW_{J1}^*$ are minimal spanning sets of $\Col (\bC_{12}) = \Col (\bC_{12}^*)$. Hence there exists a change of coordinates matrix $\bT_1$ such that $\bW_{J1}^* = \bW_{J1} \bT_1$. The same argument applies to $\bC_{21}$ such that $\bW_{J2}^* = \bW_{J2} \bA$ for some $\bA$. Then we have $\bW_{J1} \bW_{J2}^\top = \bW_{J1} \bT_1 \bA^\top \bW_{J2}^\top$. It follows that $\bW_{J2}^* = \bW_{J2} (\bT_1^{-1})^\top$. 

We next show $f_\Phi = f_{\Phi^*}$ implies (2b). Define $\bT_2 = (\bT_1^{-1})^\top$. For $k=1,2$, substituting $\bW_{Jk}^* = \bW_{Jk} \bT_k$, we write $\bC_k^*= \bW_{Jk} \bT_k \bT_k^\top \bW_{Jk}^\top + \Cov(\bW_{Ik}^* \bb_{ik}^*) + \sigma_k^2 \bI$. Then since $\bC_k = \bC_k^*$, we have $\Cov(\bW_{Ik}^* \bb_{ik}^*) = \bW_{Ik} \bW_{Ik}^\top + \bW_{Jk}(\bI - \bT_k\bT_k^\top)\bW_{Jk}^\top$. Since this must be PSD, this defines the conditions for $\bT_1$ in \eqref{eq:wspart1} and \eqref{eq:wspart2}. 

Finally, $f_\Phi = f_{\Phi*}$ implies (3) follows from the previous result that $\bC_k^* = \bC_k \implies \Cov(\bW_{Ik}^* \bb_{ik}^*) = \bW_{Ik} \bW_{Ik}^\top + \bW_{Jk}(\bI - \bT_k\bT_k^\top)\bW_{Jk}^\top$.

 \end{proof}

\section{Supplemental Information for Simulation Study}
\label{appendix:simulation}

\subsection{Proportions of Variance Explained}\label{appendix:R.squared}
A simulation study was conducted using a $2^4$ full factorial design to examine the effectiveness of ProJIVE, AJIVE, and R.JIVE for estimating the JIVE model. 

To achieve the desired proportions of variance explained via joint and individual signals, we derive numerical solutions to:
\begin{linenomath}
\begin{align*}
  R_{Ik}^2 = \frac{c_k^2 \tr(\bA_k \bA_k^\top)}{c_k^2 \tr(\bA_k \bA_k^\top) + 2c_k \tr(\bA_k \bE^\top)  + d_k^2\tr(\bJ_k \bJ_k^\top) + \tr(\bE_k \bE_k^\top) + 2d_k\tr(\bJ_k \bE_k^\top)},
\end{align*}
\end{linenomath}

\begin{linenomath}
\begin{align*}
  R_{Jk}^2 = \frac{d_k^2 \tr(\bJ_k \bJ_k^\top)}{d_k^2 \tr(\bJ_k \bJ_k^\top) + 2d_k \tr(\bJ_k \bE^\top)  + c_k^2\tr(\bA_k \bA_k^\top) + \tr(\bE_k \bE_k^\top) + 2c_k\tr(\bA_k \bE_k^\top)}.
\end{align*}
\end{linenomath}

\subsection{Chordal Subspace Norm}\label{appendix:chord.norm}

The chordal subspace norm is a distance metric for linear subspaces that has been generalized to matrices, say $\bF_1, \bF_2$, of possibly different ranks \citep{YeLim2014} and can be calculated as 
\begin{linenomath}
\begin{equation}\label{eqn:chord.norm}
  \delta(\bF_1,\bF_2) = \sqrt{\sum_{m = 1}^q \sin^2\theta_m}, 
\end{equation}
\end{linenomath}
where $q = \min_k(\textrm{rank}(\bF_k))$ and $\theta_m$ are the principal angles between the column space of $\bF_1$ and $\bF_2$. In practice we report $\delta^*(\bF_1,\bF_2) = \delta(\bF_1,\bF_2)/q$ so that the measure is bounded by $[0,1]$. We use this metric in our simulation studies to describe the accuracy of JIVE estimates. 
\newpage
\subsection{Tables of Gaussian Simulation Results}\label{appendix:gg.sim.table.jnt.score}
\subsubsection{Joint Scores}
Summary statistics of the chordal distance between true and estimated joint scores when data are simulated from the generative ProJIVE model. The lowest average chordal norm is displayed in bold print to indicate which method performed the best. Figures 1 and 2 in the main manuscript display box plots from the same simulations, where Figure 1 corresponds to rows with $r_J=1$, and Figure 2, those with $r_J=3$.
\begin{table}[h]
\label{tbl:summary.stats.jnt.scores.gg}
\begin{tabular}{|p{.2in}|c|c|c|c|ccccc|}
\hline
\multirow{2}{*}{}                      & $r_J$                 & $P_2$                  & $R_{J1}^2$                  & $R_{J2}^2$ & ProJIVE     & AJIVE       & R.JIVE      & GIPCA       & D-CCA       \\ \hline
                                       & \multicolumn{3}{l}{} & & \multicolumn{5}{c|}{Mean (SD)}\\ \hline
\parbox[t]{2mm}{\multirow{16}{*}{\rotatebox[origin=c]{90}{Joint Subject Scores}}} & \multirow{8}{*}{1} & \multirow{4}{*}{20}  & \multirow{2}{*}{0.1} & 0.1 & \textbf{0.39 (0.01)} & \textbf{0.39 (0.01)} & 0.40 (0.02) & 0.68 (0.23) & 0.49 (0.17) \\ \cline{5-10}
                                       &                    &                      &                      & 0.5 & \textbf{0.16 (0.01)} & 0.28 (0.01) & 0.30 (0.14) & 0.26 (0.08) & 0.38 (0.20) \\ \cline{4-10}
                                       &                    &                      & \multirow{2}{*}{0.5} & 0.1 & \textbf{0.16 (0.01)} & 0.28 (0.01) & 0.20 (0.06) & 0.26 (0.08) & 0.39 (0.17) \\ \cline{5-10}
                                       &                    &                      &                      & 0.5 & \textbf{0.12 (0.00)} & 0.12 (0.01) & 0.12 (0.01) & 0.16 (0.05) & 0.26 (0.23) \\ \cline{3-10}
                                       &                    & \multirow{4}{*}{200} & \multirow{2}{*}{0.1} & 0.1 & \textbf{0.17 (0.00)} & 0.28 (0.01) & 0.31 (0.01) & 0.95 (0.10) & 0.35 (0.13) \\ \cline{5-10}
                                       &                    &                      &                      & 0.5 & \textbf{0.06 (0.01)} & 0.27 (0.01) & 0.25 (0.17) & 0.13 (0.07) & 0.39 (0.19) \\ \cline{4-10}
                                       &                    &                      & \multirow{2}{*}{0.5} & 0.1 & \textbf{0.12 (0.00)} & \textbf{0.12 (0.00)} & 0.14 (0.01) & 0.71 (0.26) & 0.27 (0.22) \\ \cline{5-10}
                                       &                    &                      &                      & 0.5 & \textbf{0.05 (0.00)} & 0.09 (0.00) & 0.09 (0.01) & 0.11 (0.08) & 0.24 (0.23) \\ \cline{2-10}
                                       & \multirow{8}{*}{3} & \multirow{4}{*}{20}  & \multirow{2}{*}{0.1} & 0.1 & 0.69 (0.02) & \textbf{0.59 (0.05}) & 0.69 (0.02) & 0.85 (0.02) & 0.71 (0.07) \\ \cline{5-10}
                                       &                    &                      &                      & 0.5 & \textbf{0.38 (0.04)} & 0.50 (0.06) & 0.49 (0.08) & 0.64 (0.05) & 0.53 (0.07) \\ \cline{4-10}
                                       &                    &                      & \multirow{2}{*}{0.5} & 0.1 & \textbf{0.39 (0.04)} & 0.50 (0.05) & 0.49 (0.07) & 0.66 (0.07) & 0.52 (0.07) \\ \cline{5-10}
                                       &                    &                      &                      & 0.5 & \textbf{0.28 (0.03)} & 0.30 (0.03) & 0.29 (0.03) & 0.61 (0.07) & 0.29 (0.06) \\ \cline{3-10}
                                       &                    & \multirow{4}{*}{200} & \multirow{2}{*}{0.1} & 0.1 & \textbf{0.40 (0.02)} & 0.50 (0.06) & 0.67 (0.03) & 0.80 (0.04) & 0.54 (0.08) \\ \cline{5-10}
                                       &                    &                      &                      & 0.5 & \textbf{0.15 (0.02)} & 0.45 (0.05) & 0.40 (0.05) & 0.58 (0.01) & 0.47 (0.07) \\ \cline{4-10}
                                       &                    &                      & \multirow{2}{*}{0.5} & 0.1 & \textbf{0.30 (0.02)} & 0.31 (0.02) & 0.34 (0.03) & 0.70 (0.08) & 0.30 (0.06) \\ \cline{5-10}
                                       &                    &                      &                      & 0.5 & \textbf{0.12 (0.01)} & 0.22 (0.02) & 0.22 (0.01) & 0.58 (0.01) & 0.21 (0.04) \\ \hline
\end{tabular}
\end{table}

\newpage
\FloatBarrier  
\subsubsection{Joint Loadings}\label{appendix:gg.sim.table.jnt.ld1}
Summary statistics of the chordal distance between true and estimated joint loadings when data are simulated from the generative ProJIVE model. The lowest average chordal norm is displayed in bold print to indicate which method performed the best. Figures 1 and 2 in the main manuscript display box plots from the same simulations, where Figure 1 corresponds to rows with $r_J=1$, and Figure 2, those with $r_J=3$.
\begin{table}[h]
\begin{tabular}{|p{.2in}|c|c|c|c|ccccc|}
\hline
\multirow{2}{*}{}                      & $r_J$                 & $P_2$                  & $R_{J1}^2$                  & $R_{J2}^2$ & ProJIVE     & AJIVE       & R.JIVE      & GIPCA       & D-CCA       \\ \hline
                                       & \multicolumn{3}{l}{} & & \multicolumn{5}{c|}{Mean (SD)}\\ \hline
\parbox[t]{1mm}{\multirow{16}{*}{\rotatebox[origin=c]{90}{Joint Loadings X1}}}    & \multirow{8}{*}{1} & \multirow{4}{*}{20}  & \multirow{2}{*}{0.1} & 0.1 & \textbf{0.10 (0.02)} & \textbf{0.10 (0.02)} & \textbf{0.10 (0.02)} & 0.43 (0.30) & \textbf{0.10 (0.02)} \\ \cline{5-10}
                                       &                    &                      &                      & 0.5 & \textbf{0.09 (0.02)} & \textbf{0.09 (0.02)} & 0.14 (0.09) & 0.10 (0.03) & 0.10 (0.02) \\ \cline{4-10}
                                       &                    &                      & \multirow{2}{*}{0.5} & 0.1 & \textbf{0.03 (0.01)} & \textbf{0.03 (0.01)} & 0.04 (0.02) & 0.10 (0.05) & \textbf{0.03 (0.01)} \\ \cline{5-10}
                                       &                    &                      &                      & 0.5 & \textbf{0.03 (0.01)} & \textbf{0.03 (0.01)} & \textbf{0.03 (0.01)} & 0.05 (0.02) & \textbf{0.03 (0.01)} \\  \cline{3-10}
                                       &                    & \multirow{4}{*}{200} & \multirow{2}{*}{0.1} & 0.1 & \textbf{0.09 (0.02)} & \textbf{0.09 (0.02)} &\textbf{ 0.09 (0.02)} & 0.50 (0.28) & 0.10 (0.02) \\ \cline{5-10}
                                       &                    &                      &                      & 0.5 & \textbf{0.09 (0.02)} & \textbf{0.09 (0.02)} & 0.14 (0.10) & \textbf{0.09 (0.02)} & 0.10 (0.02) \\ \cline{4-10}
                                       &                    &                      & \multirow{2}{*}{0.5} & 0.1 & \textbf{0.03 (0.01)} & \textbf{0.03 (0.01)} & \textbf{0.03 (0.01)} & 0.08 (0.06) & \textbf{0.03 (0.01)} \\ \cline{5-10}
                                       &                    &                      &                      & 0.5 & \textbf{0.03 (0.01)} & \textbf{0.03 (0.01)} & \textbf{0.03 (0.01)} & 0.03 (0.02) & \textbf{0.03 (0.01)} \\ \cline{2-10}
                                       & \multirow{8}{*}{3} & \multirow{4}{*}{20}  & \multirow{2}{*}{0.1} & 0.1 & 0.41 (0.09) & \textbf{0.23 (0.09)} & 0.39 (0.08) & 0.69 (0.06) & 0.52 (0.09) \\  \cline{5-10}
                                       &                    &                      &                      & 0.5 & 0.25 (0.05) & 0.31 (0.12) & \textbf{0.30 (0.11)} & 0.53 (0.09) & 0.45 (0.11) \\  \cline{4-10}
                                       &                    &                      & \multirow{2}{*}{0.5} & 0.1 & \textbf{0.15 (0.07)} & 0.15 (0.09) & 0.20 (0.10) & 0.56 (0.04) & 0.20 (0.12) \\  \cline{5-10}
                                       &                    &                      &                      & 0.5 & \textbf{0.09 (0.03)} & \textbf{0.09 (0.03)} & 0.10 (0.04) & 0.48 (0.13) & \textbf{0.09 (0.03)} \\  \cline{3-10}
                                       &                    & \multirow{4}{*}{200} & \multirow{2}{*}{0.1} & 0.1 & \textbf{0.26 (0.05)} & 0.32 (0.12) & 0.40 (0.10) & 0.67 (0.07) & 0.45 (0.10) \\  \cline{5-10}
                                       &                    &                      &                      & 0.5 & \textbf{0.23 (0.04)} & 0.31 (0.11) & 0.26 (0.07) & 0.54 (0.06) & 0.44 (0.11) \\  \cline{4-10}
                                       &                    &                      & \multirow{2}{*}{0.5} & 0.1 & \textbf{0.09 (0.03)} & \textbf{0.09 (0.03)} & 0.10 (0.04) & 0.52 (0.13) & \textbf{0.09 (0.03)} \\ \cline{5-10}
                                       &                    &                      &                      & 0.5 & \textbf{0.08 (0.02)} & \textbf{0.08 (0.02)} & \textbf{0.08 (0.02)} & 0.43 (0.13) & 0.09 (0.02) \\ \hline
\parbox[t]{2mm}{\multirow{16}{*}{\rotatebox[origin=c]{90}{Joint Loadings X2}}}    & \multirow{8}{*}{1} & \multirow{4}{*}{20}  & \multirow{2}{*}{0.1} & 0.1 & \textbf{0.10 (0.02)} & \textbf{0.10 (0.02)} & 0.12 (0.04) & 0.43 (0.30) & 0.11 (0.02) \\ \cline{5-10}
                                       &                    &                      &                      & 0.5 & \textbf{0.03 (0.01)} & \textbf{0.03 (0.01)} & 0.06 (0.02) & 0.10 (0.05) & \textbf{0.03 (0.01)} \\\cline{4-10}
                                       &                    &                      & \multirow{2}{*}{0.5} & 0.1 & \textbf{0.09 (0.02)} & \textbf{0.09 (0.02)} & \textbf{0.09 (0.02)} & 0.10 (0.02) & 0.10 (0.02) \\ \cline{5-10}
                                       &                    &                      &                      & 0.5 & \textbf{0.03 (0.01)} & \textbf{0.03 (0.01)} & \textbf{0.03 (0.01)} & 0.05 (0.03) & \textbf{0.03 (0.01)} \\ \cline{3-10}
                                       &                    & \multirow{4}{*}{200} & \multirow{2}{*}{0.1} & 0.1 & \textbf{0.10 (0.02)} & \textbf{0.10 (0.02)} & \textbf{0.10 (0.02)} & 0.95 (0.10) & \textbf{0.10 (0.02)} \\ \cline{5-10}
                                       &                    &                      &                      & 0.5 & \textbf{0.03 (0.01}) & \textbf{0.03 (0.01)} & 0.05 (0.02) & 0.07 (0.04) & \textbf{0.03 (0.01)} \\ \cline{4-10}
                                       &                    &                      & \multirow{2}{*}{0.5} & 0.1 & \textbf{0.10 (0.01)} & \textbf{0.10 (0.01)} & \textbf{0.10 (0.01)} & 0.75 (0.28) & \textbf{0.10 (0.01)} \\ \cline{5-10}
                                       &                    &                      &                      & 0.5 & \textbf{0.03 (0.01)} & \textbf{0.03 (0.01)} & \textbf{0.03 (0.01)} & 0.07 (0.05) & \textbf{0.03 (0.01)} \\ \cline{2-10}
                                       & \multirow{8}{*}{3} & \multirow{4}{*}{20}  & \multirow{2}{*}{0.1} & 0.1 & 0.40 (0.09) & \textbf{0.23 (0.10)} & 0.38 (0.08) & 0.69 (0.06) & 0.50 (0.08) \\ \cline{5-10}
                                       &                    &                      &                      & 0.5 & \textbf{0.14 (0.06)} & 0.14 (0.08) & 0.21 (0.12) & 0.56 (0.04) & 0.20 (0.11) \\ \cline{4-10}
                                       &                    &                      & \multirow{2}{*}{0.5} & 0.1 & \textbf{0.25 (0.05)} & 0.32 (0.09) & 0.30 (0.09) & 0.54 (0.09) & 0.43 (0.09) \\ \cline{5-10}
                                       &                    &                      &                      & 0.5 & \textbf{0.09 (0.02)} & \textbf{0.09 (0.02)} & 0.10 (0.03) & 0.51 (0.10) & 0.09 (0.03) \\ \cline{3-10}
                                       &                    & \multirow{4}{*}{200} & \multirow{2}{*}{0.1} & 0.1 & \textbf{0.32 (0.07)} & 0.32 (0.11) & 0.43 (0.08) & 0.79 (0.04) & 0.39 (0.10) \\ \cline{5-10}
                                       &                    &                      &                      & 0.5 & \textbf{0.14 (0.06)} & 0.15 (0.09) & 0.14 (0.07) & 0.58 (0.00) & 0.20 (0.11) \\ \cline{4-10}
                                       &                    &                      & \multirow{2}{*}{0.5} & 0.1 & \textbf{0.25 (0.03)} & \textbf{0.25 (0.03)} & \textbf{0.25 (0.03)} & 0.64 (0.07) & 0.26 (0.03) \\ \cline{5-10}
                                       &                    &                      &                      & 0.5 & \textbf{0.08 (0.02)} & 0.09 (0.02) & \textbf{0.08 (0.02)} & 0.58 (0.00) & \textbf{0.08 (0.02)} \\ \hline
\end{tabular}
\end{table}
\FloatBarrier  

\newpage
\FloatBarrier  
\subsubsection{Individual $X_1$ Scores and Loadings}\label{appendix:gg.sim.table.indiv.1}
Summary statistics of the chordal distance between true and estimated individual scores and individual loadings for data set $\bX_1$ when data are simulated from the generative ProJIVE model. The lowest average chordal norm is displayed in bold print to indicate which method performed the best. Figures 1 and 2 in the main manuscript display box plots from the same simulations, where Figure 1 corresponds to rows with $r_J=1$, and Figure 2, those with $r_J=3$.
\begin{table}[h]
\begin{tabular}{|p{.2in}|c|c|c|c|ccccc|}
\hline
\multirow{2}{*}{}                      & $r_J$                 & $P_2$                  & $R_{J1}^2$                  & $R_{J2}^2$ & ProJIVE     & AJIVE       & R.JIVE      & GIPCA       & D-CCA       \\ \hline
                                       & \multicolumn{3}{l}{} & & \multicolumn{5}{c|}{Mean (SD)}\\ \hline
\parbox[t]{1mm}{\multirow{16}{*}{\rotatebox[origin=c]{90}{Individual $X_1$ Scores}}}    & \multirow{8}{*}{1} & \multirow{4}{*}{20}  & \multirow{2}{*}{0.1} & 0.1 & \textbf{0.54 (0.04)} & \textbf{0.54 (0.04)} & \textbf{0.54 (0.04)} & 0.64 (0.13) & \textbf{0.54 (0.04)} \\ \cline{5-10}
                                               &                    &                      &                      & 0.5 & \textbf{0.53 (0.04)} & 0.54 (0.04) & 0.58 (0.05) & 0.54 (0.04) & 0.54 (0.04) \\  \cline{4-10}
                                               &                    &                      & \multirow{2}{*}{0.5} & 0.1 & \textbf{0.38 (0.04)} & 0.44 (0.10) & 0.40 (0.06) & 0.41 (0.05) & 0.65 (0.10) \\  \cline{5-10}
                                               &                    &                      &                      & 0.5 & \textbf{0.38 (0.04)} & \textbf{0.38 (0.04)} & 0.39 (0.04) & 0.39 (0.06) & \textbf{0.38 (0.04)} \\  \cline{3-10}
                                               &                    & \multirow{4}{*}{200} & \multirow{2}{*}{0.1} & 0.1 & \textbf{0.53 (0.04)} & 0.54 (0.04) & 0.54 (0.04) & 0.71 (0.07) & 0.54 (0.04) \\  \cline{5-10}
                                               &                    &                      &                      & 0.5 & \textbf{0.53 (0.04)} & 0.54 (0.04) & 0.58 (0.06) & 0.54 (0.04) & 0.54 (0.04) \\  \cline{4-10}
                                               &                    &                      & \multirow{2}{*}{0.5} & 0.1 & \textbf{0.38 (0.04)} & \textbf{0.38 (0.04)} & 0.39 (0.04) & 0.67 (0.13) & \textbf{0.38 (0.04)} \\  \cline{5-10}
                                               &                    &                      &                      & 0.5 & \textbf{0.38 (0.04)} & \textbf{0.38 (0.04)} & \textbf{0.38 (0.04)} & 0.41 (0.11) & \textbf{0.38 (0.04)} \\  \cline{2-10}
                                               & \multirow{8}{*}{3} & \multirow{4}{*}{20}  & \multirow{2}{*}{0.1} & 0.1 & \textbf{0.54 (0.04)} & 0.55 (0.04) & 0.54 (0.05) & 0.83 (0.05) & \textbf{0.54 (0.04)} \\  \cline{5-10}
                                               &                    &                      &                      & 0.5 & \textbf{0.53 (0.04) }& 0.56 (0.05) & 0.56 (0.07) & 0.57 (0.10) & \textbf{0.53 (0.04)} \\  \cline{4-10}
                                               &                    &                      & \multirow{2}{*}{0.5} & 0.1 & \textbf{0.40 (0.04)} & 0.49 (0.10) & 0.45 (0.09) & 0.78 (0.06) & 0.67 (0.08) \\  \cline{5-10}
                                               &                    &                      &                      & 0.5 & \textbf{0.38 (0.04)} & 0.39 (0.04) & 0.40 (0.04) & 0.64 (0.16) & 0.40 (0.05) \\  \cline{3-10}
                                               &                    & \multirow{4}{*}{200} & \multirow{2}{*}{0.1} & 0.1 & \textbf{0.53 (0.04)} & 0.56 (0.05) & 0.56 (0.05) & 0.58 (0.08) & \textbf{0.53 (0.04)} \\  \cline{5-10}
                                               &                    &                      &                      & 0.5 & \textbf{0.53 (0.04)} & 0.56 (0.05) & 0.54 (0.06) & \textbf{0.53 (0.04)} & \textbf{0.53 (0.04)} \\  \cline{4-10}
                                               &                    &                      & \multirow{2}{*}{0.5} & 0.1 & \textbf{0.39 (0.04)} & \textbf{0.39 (0.04)} & 0.40 (0.04) & 0.65 (0.15) & \textbf{0.39 (0.04)} \\  \cline{5-10}
                                               &                    &                      &                      & 0.5 & \textbf{0.37 (0.04)} & 0.38 (0.04) & 0.38 (0.04) & 0.48 (0.13) & 0.38 (0.04) \\ \hline
\parbox[t]{2mm}{\multirow{16}{*}{\rotatebox[origin=c]{90}{Individual Loadings $X_1$}}}    & \multirow{8}{*}{1} & \multirow{4}{*}{20}  & \multirow{2}{*}{0.1} & 0.1 & \textbf{0.12 (0.03)} & 0.13 (0.04) & 0.13 (0.04) & 0.37 (0.25) & \textbf{0.12 (0.03)} \\  \cline{5-10}
                                               &                    &                      &                      & 0.5 & \textbf{0.12 (0.03)} & 0.13 (0.04) & 0.19 (0.11) & 0.13 (0.05) & 0.13 (0.04) \\  \cline{4-10}
                                               &                    &                      & \multirow{2}{*}{0.5} & 0.1 & \textbf{0.08 (0.03)} & 0.19 (0.15) & 0.14 (0.11) & 0.31 (0.14) & 0.56 (0.16) \\  \cline{5-10}
                                               &                    &                      &                      & 0.5 & \textbf{0.06 (0.02)} & 0.07 (0.02) & 0.07 (0.02) & 0.11 (0.08) & \textbf{0.06 (0.02)} \\  \cline{3-10}
                                               &                    & \multirow{4}{*}{200} & \multirow{2}{*}{0.1} & 0.1 & \textbf{0.12 (0.03)} & 0.13 (0.04) & 0.13 (0.04) & 0.58 (0.17) & 0.12 (0.04) \\  \cline{5-10}
                                               &                    &                      &                      & 0.5 & \textbf{0.12 (0.03)} & 0.13 (0.04) & 0.19 (0.12) & 0.13 (0.07) & 0.13 (0.04) \\  \cline{4-10}
                                               &                    &                      & \multirow{2}{*}{0.5} & 0.1 & \textbf{0.06 (0.02)} & 0.07 (0.02) & 0.07 (0.02) & 0.59 (0.22) & 0.07 (0.02) \\  \cline{5-10}
                                               &                    &                      &                      & 0.5 & \textbf{0.06 (0.01)} & 0.07 (0.02) & 0.07 (0.02) & 0.12 (0.18) & 0.07 (0.02) \\  \cline{2-10}
                                               & \multirow{8}{*}{3} & \multirow{4}{*}{20}  & \multirow{2}{*}{0.1} & 0.1 & 0.14 (0.08) & 0.15 (0.06) & \textbf{0.14 (0.07)} & 0.66 (0.08) & \textbf{0.14 (0.07)} \\  \cline{5-10}
                                               &                    &                      &                      & 0.5 & \textbf{0.11 (0.03)} & 0.18 (0.08) & 0.17 (0.15) & 0.19 (0.16) & 0.13 (0.05) \\  \cline{4-10}
                                               &                    &                      & \multirow{2}{*}{0.5} & 0.1 & \textbf{0.12 (0.05)} & 0.25 (0.17) & 0.24 (0.15) & 0.71 (0.09) & 0.57 (0.14) \\  \cline{5-10}
                                               &                    &                      &                      & 0.5 & \textbf{0.07 (0.02)} & 0.08 (0.02) & 0.10 (0.04) & 0.46 (0.23) & 0.08 (0.06) \\  \cline{3-10}
                                               &                    & \multirow{4}{*}{200} & \multirow{2}{*}{0.1} & 0.1 & \textbf{0.11 (0.03)} & 0.18 (0.08) & 0.18 (0.08) & 0.23 (0.16) & 0.13 (0.05) \\  \cline{5-10}
                                               &                    &                      &                      & 0.5 & \textbf{0.11 (0.03)} & 0.18 (0.08) & 0.14 (0.10) & 0.13 (0.05) & 0.14 (0.05) \\  \cline{4-10}
                                               &                    &                      & \multirow{2}{*}{0.5} & 0.1 & \textbf{0.07 (0.02)} & 0.08 (0.02) & 0.09 (0.03) & 0.53 (0.21) & 0.08 (0.03) \\  \cline{5-10}
                                               &                    &                      &                      & 0.5 & \textbf{0.06 (0.01)} & 0.08 (0.03) & 0.07 (0.02) & 0.28 (0.21) & 0.07 (0.02) \\ \hline
\end{tabular}
\end{table}
\FloatBarrier  

\newpage
\FloatBarrier  
\subsubsection{Individual $X_2$ Scores and Loadings}\label{appendix:gg.sim.table.indiv.2}
Summary statistics of the chordal distance between true and estimated individual scores and individual loadings for data set $\bX_2$ when data are simulated from the generative ProJIVE model. The lowest average chordal norm is displayed in bold print to indicate which method performed the best. Figures 1 and 2 in the main manuscript display box plots from the same simulations, where Figure 1 corresponds to rows with $r_J=1$, and Figure 2, those with $r_J=3$.
\begin{table}[h]
\begin{tabular}{|p{.2in}|c|c|c|c|ccccc|}
\hline
\multirow{2}{*}{}                      & $r_J$                 & $P_2$                  & $R_{J1}^2$                  & $R_{J2}^2$ & ProJIVE     & AJIVE       & R.JIVE      & GIPCA       & D-CCA       \\ \hline
                                       & \multicolumn{3}{l}{} & & \multicolumn{5}{c|}{Mean (SD)}\\ \hline
\parbox[t]{1mm}{\multirow{16}{*}{\rotatebox[origin=c]{90}{Individual $X_2$ Scores}}}    & \multirow{8}{*}{1} & \multirow{4}{*}{20}  & \multirow{2}{*}{0.1} & 0.1 &  \textbf{0.53 (0.04)} & 0.54 (0.04) & 0.54 (0.04) & 0.64 (0.12) & 0.54 (0.04) \\ \cline{5-10}
                                               &                    &                      &                      & 0.5 & \textbf{0.38 (0.04)} & 0.44 (0.09) & 0.47 (0.10) & 0.41 (0.05) & 0.63 (0.10) \\ \cline{4-10}
                                               &                    &                      & \multirow{2}{*}{0.5} & 0.1 & \textbf{0.53 (0.04)} & 0.54 (0.04) & 0.53 (0.04) & 0.54 (0.05) & \textbf{0.53 (0.04)} \\ \cline{5-10}
                                               &                    &                      &                      & 0.5 & \textbf{0.38 (0.04)} & \textbf{0.38 (0.04)} & \textbf{0.38 (0.04)} & 0.39 (0.05) & \textbf{0.38 (0.04)} \\ \cline{3-10}
                                               &                    & \multirow{4}{*}{200} & \multirow{2}{*}{0.1} & 0.1 & \textbf{0.20 (0.01)} & 0.21 (0.01) & 0.21 (0.01) & 0.69 (0.08) & 0.21 (0.01) \\ \cline{5-10}
                                               &                    &                      &                      & 0.5 & \textbf{0.13 (0.01)} & 0.20 (0.09) & 0.33 (0.17) & 0.16 (0.04) & 0.69 (0.03) \\ \cline{4-10}
                                               &                    &                      & \multirow{2}{*}{0.5} & 0.1 & \textbf{0.20 (0.01)} & \textbf{0.20 (0.01)} & 0.21 (0.01) & 0.53 (0.17) & \textbf{0.20 (0.01)} \\ \cline{5-10}
                                               &                    &                      &                      & 0.5 & \textbf{0.13 (0.01)} & \textbf{0.13 (0.01)} & 0.14 (0.01) & 0.16 (0.08) & \textbf{0.13 (0.01)} \\ \cline{2-10}
                                               & \multirow{8}{*}{3} & \multirow{4}{*}{20}  & \multirow{2}{*}{0.1} & 0.1 & 0.55 (0.05) & 0.56 (0.05) & 0.56 (0.06) & 0.84 (0.04) & \textbf{0.55 (0.04)} \\ \cline{5-10}
                                               &                    &                      &                      & 0.5 & \textbf{0.41 (0.05)} & 0.50 (0.09) & 0.47 (0.10) & 0.79 (0.05) & 0.68 (0.07) \\ \cline{4-10}
                                               &                    &                      & \multirow{2}{*}{0.5} & 0.1 & \textbf{0.54 (0.04)} & 0.56 (0.05) & 0.57 (0.09) & 0.62 (0.14) & \textbf{0.54 (0.04)} \\ \cline{5-10}
                                               &                    &                      &                      & 0.5 & \textbf{0.39 (0.04)} & 0.40 (0.04) & 0.41 (0.05) & 0.70 (0.15) & 0.40 (0.05) \\ \cline{3-10}
                                               &                    & \multirow{4}{*}{200} & \multirow{2}{*}{0.1} & 0.1 & 0.22 (0.02) & 0.24 (0.04) & 0.22 (0.01) & 0.95 (0.07) & \textbf{0.21 (0.01) }\\ \cline{5-10}
                                               &                    &                      &                      & 0.5 & \textbf{0.17 (0.03)} & 0.28 (0.12) & 0.21 (0.07) & 0.72 (0.01) & 0.70 (0.02) \\ \cline{4-10}
                                               &                    &                      & \multirow{2}{*}{0.5} & 0.1 & \textbf{0.21 (0.01)} & \textbf{0.21 (0.01)} & \textbf{0.22 (0.01)} & 0.76 (0.08) & \textbf{0.21 (0.01)} \\ \cline{5-10}
                                               &                    &                      &                      & 0.5 & \textbf{0.14 (0.01)} & 0.15 (0.01) & 0.16 (0.02) & 0.71 (0.01) & \textbf{0.14 (0.01)} \\ \hline
\parbox[t]{2mm}{\multirow{16}{*}{\rotatebox[origin=c]{90}{Individual Loadings $X_2$}}}    & \multirow{8}{*}{1} & \multirow{4}{*}{20}  & \multirow{2}{*}{0.1} & 0.1 & \textbf{0.12 (0.03)} & 0.13 (0.04) & 0.13 (0.04) & 0.37 (0.25) & \textbf{0.12 (0.03)} \\ \cline{5-10}
                                               &                    &                      &                      & 0.5 & \textbf{0.12 (0.03)} & 0.13 (0.04) & 0.19 (0.11) & 0.13 (0.05) & 0.13 (0.04) \\ \cline{4-10}
                                               &                    &                      & \multirow{2}{*}{0.5} & 0.1 & \textbf{0.08 (0.03)} & 0.19 (0.15) & 0.14 (0.11) & 0.31 (0.14) & 0.56 (0.16) \\ \cline{5-10}
                                               &                    &                      &                      & 0.5 & \textbf{0.06 (0.02)} & 0.07 (0.02) & 0.07 (0.02) & 0.11 (0.08) & \textbf{0.06 (0.02)} \\ \cline{3-10}
                                               &                    & \multirow{4}{*}{200} & \multirow{2}{*}{0.1} & 0.1 & \textbf{0.12 (0.03)} & 0.13 (0.04) & 0.13 (0.04) & 0.58 (0.17) & 0.12 (0.04) \\ \cline{5-10}
                                               &                    &                      &                      & 0.5 & \textbf{0.12 (0.03)} & 0.13 (0.04) & 0.19 (0.12) & 0.13 (0.07) & 0.13 (0.04) \\ \cline{4-10}
                                               &                    &                      & \multirow{2}{*}{0.5} & 0.1 & \textbf{0.06 (0.02)} & 0.07 (0.02) & 0.07 (0.02) & 0.59 (0.22) & 0.07 (0.02) \\ \cline{5-10}
                                               &                    &                      &                      & 0.5 & \textbf{0.06 (0.01)} & 0.07 (0.02) & 0.07 (0.02) & 0.12 (0.18) & 0.07 (0.02) \\ \cline{2-10}
                                               & \multirow{8}{*}{3} & \multirow{4}{*}{20}  & \multirow{2}{*}{0.1} & 0.1 & 0.14 (0.08) & 0.15 (0.06) & \textbf{0.14 (0.07)} & 0.66 (0.08) & \textbf{0.14 (0.07)} \\ \cline{5-10}
                                               &                    &                      &                      & 0.5 & \textbf{0.11 (0.03)} & 0.18 (0.08) & 0.17 (0.15) & 0.19 (0.16) & 0.13 (0.05) \\ \cline{4-10}
                                               &                    &                      & \multirow{2}{*}{0.5} & 0.1 & \textbf{0.12 (0.05)} & 0.25 (0.17) & 0.24 (0.15) & 0.71 (0.09) & 0.57 (0.14) \\ \cline{5-10}
                                               &                    &                      &                      & 0.5 & \textbf{0.07 (0.02)} & 0.08 (0.02) & 0.10 (0.04) & 0.46 (0.23) & 0.08 (0.06) \\ \cline{3-10}
                                               &                    & \multirow{4}{*}{200} & \multirow{2}{*}{0.1} & 0.1 & \textbf{0.11 (0.03)} & 0.18 (0.08) & 0.18 (0.08) & 0.23 (0.16) & 0.13 (0.05) \\ \cline{5-10}
                                               &                    &                      &                      & 0.5 & \textbf{0.11 (0.03)} & 0.18 (0.08) & 0.14 (0.10) & 0.13 (0.05) & 0.14 (0.05) \\ \cline{4-10}
                                               &                    &                      & \multirow{2}{*}{0.5} & 0.1 & \textbf{0.07 (0.02)} & 0.08 (0.02) & 0.09 (0.03) & 0.53 (0.21) & 0.08 (0.03) \\ \cline{5-10}
                                               &                    &                      &                      & 0.5 & \textbf{0.06 (0.01)} & 0.08 (0.03) & 0.07 (0.02) & 0.28 (0.21) & 0.07 (0.02) \\ \hline
\end{tabular}
\end{table}
\FloatBarrier  

\newpage
\FloatBarrier  
\subsection{Tables of Mixture-Rademacher Simulation Results}
\subsubsection{Joint Subject Scores}\label{appendix:mr.sim.table.jnt.scores}
Summary statistics of the chordal distance between true and estimated joint scores when data \textbf{\underline{are not}} simulated from the generative ProJIVE model. The lowest average chordal norm is displayed in bold print to indicate which method performed the best. Figures 3 and 4 in the main manuscript display box plots from the same simulations, where Figure 3 corresponds to rows with $r_J=1$, and Figure 4, those with $r_J=3$.
\begin{table}[h]
\begin{tabular}{|p{.2in}|c|c|c|c|ccccc|}
\hline
\multirow{2}{*}{} & $r_J$ & $P_2$ & $R_{J1}^2$ & $R_{J2}^2$ & ProJIVE & AJIVE & R.JIVE & GIPCA & D-CCA \\ \hline
& \multicolumn{3}{l}{} & & \multicolumn{5}{c|}{Mean (SD)}\\ \hline
\parbox[t]{1mm}{\multirow{16}{*}{\rotatebox[origin=c]{90}{Joint Subject Scores}}} 
& \multirow{8}{*}{1} & \multirow{4}{*}{20} & \multirow{2}{*}{0.1} & 0.1 
& \textbf{0.39 (0.02)} & \textbf{0.39 (0.02)} & 0.42 (0.02) & 0.75 (0.22) & 0.51 (0.18) \\ \cline{5-10}

& & & & 0.5 
& \textbf{0.16 (0.01)} & 0.28 (0.01) & 0.33 (0.12) & 0.31 (0.08) & 0.43 (0.22) \\ \cline{4-10}

& & & \multirow{2}{*}{0.5} & 0.1 
& \textbf{0.16 (0.01)} & 0.28 (0.01) & 0.23 (0.02) & 0.31 (0.09) & 0.42 (0.20) \\ \cline{5-10}

& & & & 0.5 
& \textbf{0.12 (0.01)} & \textbf{0.12 (0.01)} & 0.18 (0.01) & 0.22 (0.06) & 0.28 (0.25) \\ \cline{3-10}

& & \multirow{4}{*}{200} & \multirow{2}{*}{0.1} & 0.1 
& \textbf{0.17 (0.01)} & 0.28 (0.01) & 0.34 (0.01) & 0.97 (0.06) & 0.40 (0.19) \\ \cline{5-10}

& & & & 0.5 
& \textbf{0.06 (0.01)} & 0.27 (0.01) & 0.29 (0.14) & 0.18 (0.03) & 0.40 (0.20) \\ \cline{4-10}

& & & \multirow{2}{*}{0.5} & 0.1 
& \textbf{0.12 (0.00)} & \textbf{0.12 (0.00)} & 0.19 (0.01) & 0.72 (0.26) & 0.27 (0.23) \\ \cline{5-10}

& & & & 0.5 
& \textbf{0.05 (0.00)} & 0.09 (0.01) & 0.16 (0.01) & 0.16 (0.03) & 0.25 (0.23) \\ \cline{2-10}

& \multirow{8}{*}{3} & \multirow{4}{*}{20} & \multirow{2}{*}{0.1} & 0.1 
& \textbf{0.83 (0.01)} & \textbf{0.83 (0.01)} & \textbf{0.83 (0.01)} & 0.86 (0.01) & \textbf{0.83 (0.07)} \\ \cline{5-10}

& & & & 0.5 
& \textbf{0.66 (0.04)} & 0.74 (0.03) & 0.75 (0.03) & 0.82 (0.02) & 0.74 (0.07) \\ \cline{4-10}

& & & \multirow{2}{*}{0.5} & 0.1 
& \textbf{0.66 (0.04)} & 0.74 (0.03) & 0.75 (0.04) & 0.82 (0.01) & 0.72 (0.08) \\ \cline{5-10}

& & & & 0.5 
& \textbf{0.54 (0.04)} & 0.56 (0.04) & 0.55 (0.03) & 0.81 (0.01) & \textbf{0.54 (0.09)} \\ \cline{3-10}

& & \multirow{4}{*}{200} & \multirow{2}{*}{0.1} & 0.1 
& \textbf{0.71 (0.02)} & 0.76 (0.02) & 0.82 (0.01) & 0.83 (0.00) & 0.76 (0.08) \\ \cline{5-10}

& & & & 0.5 
& \textbf{0.46 (0.10)} & 0.65 (0.04) & 0.76 (0.02) & 0.82 (0.01) & 0.66 (0.10) \\ \cline{4-10}

& & & \multirow{2}{*}{0.5} & 0.1 
& \textbf{0.61 (0.04)} & 0.64 (0.03) & 0.64 (0.04) & 0.82 (0.00) & 0.63 (0.11) \\ \cline{5-10}

& & & & 0.5 
& \textbf{0.29 (0.01)} & 0.40 (0.03) & 0.52 (0.03) & 0.80 (0.02) & 0.40 (0.08) \\ \hline
\end{tabular}
\end{table}
\FloatBarrier  

\newpage
\FloatBarrier  
\subsubsection{Joint Loadings}\label{appendix:mr.sim.table.jnt.loads}
Summary statistics of the chordal distance between true and estimated joint loadings when \textbf{\underline{data are}} simulated from the generative ProJIVE model. The lowest average chordal norm is displayed in bold print to indicate which method performed the best. Figures 3 and 4 in the main manuscript display box plots from the same simulations, where Figure 3 corresponds to rows with $r_J=1$, and Figure 4, those with $r_J=3$.
\begin{table}[h]
\begin{tabular}{|p{.2in}|c|c|c|c|ccccc|}
\hline
\multirow{2}{*}{} & $r_J$ & $P_2$ & $R_{J1}^2$ & $R_{J2}^2$ & ProJIVE & AJIVE & R.JIVE & GIPCA & D-CCA \\ \hline
& \multicolumn{3}{l}{} & & \multicolumn{5}{c|}{Mean (SD)}\\ \hline

\parbox[t]{1mm}{\multirow{16}{*}{\rotatebox[origin=c]{90}{Joint Loadings $X_1$}}}
& \multirow{8}{*}{1} & \multirow{4}{*}{20} & \multirow{2}{*}{0.1} & 0.1
& \textbf{0.10 (0.02)} & \textbf{0.10 (0.02)} & 0.11 (0.03) & 0.48 (0.30) & \textbf{0.10 (0.02)} \\ \cline{5-10}

& & & & 0.5
& \textbf{0.09 (0.02)} & \textbf{0.09 (0.02)} & 0.13 (0.07) & 0.11 (0.10) & 0.10 (0.02) \\ \cline{4-10}

& & & \multirow{2}{*}{0.5} & 0.1
& \textbf{0.03 (0.01)} & \textbf{0.03 (0.01)} & 0.04 (0.01) & 0.12 (0.08) & \textbf{0.03 (0.01)} \\ \cline{5-10}

& & & & 0.5
& 0.04 (0.05) & \textbf{0.03 (0.01)} & \textbf{0.03 (0.01)} & 0.05 (0.04) & \textbf{0.03 (0.01)} \\ \cline{3-10}

& & \multirow{4}{*}{200} & \multirow{2}{*}{0.1} & 0.1
& \textbf{0.09 (0.02)} & \textbf{0.09 (0.02)} & \textbf{0.09 (0.02)} & 0.50 (0.26) & 0.10 (0.02) \\ \cline{5-10}

& & & & 0.5
& \textbf{0.09 (0.02)} & \textbf{0.09 (0.02)} & 0.13 (0.07) & \textbf{0.09 (0.02)} & 0.10 (0.02) \\ \cline{4-10}

& & & \multirow{2}{*}{0.5} & 0.1
& \textbf{0.03 (0.01)} & \textbf{0.03 (0.01)} & \textbf{0.03 (0.01)} & 0.10 (0.14) & \textbf{0.03 (0.01)} \\ \cline{5-10}

& & & & 0.5
& \textbf{0.03 (0.01)} & \textbf{0.03 (0.01)} & \textbf{0.03 (0.01)} & \textbf{0.03 (0.01)} & \textbf{0.03 (0.01)} \\ \cline{2-10}

& \multirow{8}{*}{3} & \multirow{4}{*}{20} & \multirow{2}{*}{0.1} & 0.1
& 0.70 (0.05) & 0.73 (0.05) & \textbf{0.69 (0.05)} & 0.77 (0.03) & 0.73 (0.05) \\ \cline{5-10}

& & & & 0.5
& \textbf{0.59 (0.06)} & 0.67 (0.05) & 0.63 (0.05) & 0.75 (0.04) & 0.69 (0.05) \\ \cline{4-10}

& & & \multirow{2}{*}{0.5} & 0.1
& \textbf{0.51 (0.10)} & 0.55 (0.08) & 0.57 (0.07) & 0.73 (0.05) & 0.55 (0.08) \\ \cline{5-10}

& & & & 0.5
& 0.29 (0.09) & 0.33 (0.10) & \textbf{0.26 (0.07)} & 0.66 (0.06) & 0.35 (0.10) \\ \cline{3-10}

& & \multirow{4}{*}{200} & \multirow{2}{*}{0.1} & 0.1
& \textbf{0.63 (0.04)} & 0.68 (0.05) & 0.70 (0.05) & 0.77 (0.03) & 0.70 (0.05) \\ \cline{5-10}

& & & & 0.5
& \textbf{0.52 (0.08)} & 0.65 (0.06) & 0.64 (0.04) & 0.76 (0.04) & 0.69 (0.05) \\ \cline{4-10}

& & & \multirow{2}{*}{0.5} & 0.1
& \textbf{0.40 (0.13)} & 0.51 (0.09) & 0.45 (0.12) & 0.75 (0.05) & 0.51 (0.09) \\ \cline{5-10}

& & & & 0.5
& \textbf{0.20 (0.05)} & 0.25 (0.07) & 0.27 (0.10) & 0.68 (0.07) & 0.29 (0.08) \\ \hline

\parbox[t]{1mm}{\multirow{16}{*}{\rotatebox[origin=c]{90}{Joint Loadings $X_2$}}}
& \multirow{8}{*}{1} & \multirow{4}{*}{20} & \multirow{2}{*}{0.1} & 0.1
& \textbf{0.10 (0.02)} & \textbf{0.10 (0.02)} & 0.12 (0.04) & 0.52 (0.31) & \textbf{0.10 (0.02)} \\ \cline{5-10}

& & & & 0.5
& \textbf{0.03 (0.01)} & \textbf{0.03 (0.01)} & 0.07 (0.03) & 0.13 (0.08) & \textbf{0.03 (0.01)} \\ \cline{4-10}

& & & \multirow{2}{*}{0.5} & 0.1
& \textbf{0.09 (0.02)} & \textbf{0.09 (0.02)} & \textbf{0.09 (0.02)} & 0.11 (0.07) & 0.10 (0.02) \\ \cline{5-10}

& & & & 0.5
& 0.05 (0.07) & \textbf{0.03 (0.01)} & \textbf{0.03 (0.01)} & 0.05 (0.03) & \textbf{0.03 (0.01)} \\ \cline{3-10}

& & \multirow{4}{*}{200} & \multirow{2}{*}{0.1} & 0.1
& 0.12 (0.04) & \textbf{0.10 (0.02)} & \textbf{0.10 (0.02)} & 0.97 (0.06) & \textbf{0.10 (0.01)} \\ \cline{5-10}

& & & & 0.5
& \textbf{0.03 (0.01)} & \textbf{0.03 (0.01)} & 0.06 (0.03) & 0.07 (0.03) & \textbf{0.03 (0.01)} \\ \cline{4-10}

& & & \multirow{2}{*}{0.5} & 0.1
& \textbf{0.09 (0.01)} & \textbf{0.09 (0.01)} & \textbf{0.09 (0.01)} & 0.74 (0.30) & \textbf{0.09 (0.01)} \\ \cline{5-10}

& & & & 0.5
& \textbf{0.03 (0.01)} & \textbf{0.03 (0.01)} & \textbf{0.03 (0.01)} & 0.06 (0.04) & \textbf{0.03 (0.01)} \\ \cline{2-10}

& \multirow{8}{*}{3} & \multirow{4}{*}{20} & \multirow{2}{*}{0.1} & 0.1
& \textbf{0.71 (0.05)} & 0.72 (0.05) & \textbf{0.70 (0.06)} & 0.77 (0.03) & 0.72 (0.05) \\ \cline{5-10}

& & & & 0.5
& \textbf{0.51 (0.10)} & 0.55 (0.08) & 0.58 (0.06) & 0.73 (0.05) & 0.55 (0.08) \\ \cline{4-10}

& & & \multirow{2}{*}{0.5} & 0.1
& \textbf{0.59 (0.07)} & 0.67 (0.07) & 0.63 (0.06) & 0.75 (0.04) & 0.69 (0.06) \\ \cline{5-10}

& & & & 0.5
& \textbf{0.28 (0.09)} & 0.33 (0.10) & \textbf{0.27 (0.08)} & 0.68 (0.07) & 0.35 (0.10) \\ \cline{3-10}

& & \multirow{4}{*}{200} & \multirow{2}{*}{0.1} & 0.1
& \textbf{0.67 (0.04)} & 0.69 (0.05) & 0.76 (0.04) & 0.81 (0.00) & 0.69 (0.05) \\ \cline{5-10}

& & & & 0.5
& \textbf{0.54 (0.07)} & 0.55 (0.09) & 0.59 (0.03) & 0.81 (0.01) & 0.55 (0.10) \\ \cline{4-10}

& & & \multirow{2}{*}{0.5} & 0.1
& \textbf{0.59 (0.04)} & 0.63 (0.02) & 0.60 (0.03) & 0.81 (0.01) & 0.64 (0.02) \\ \cline{5-10}

& & & & 0.5
& \textbf{0.25 (0.07)} & 0.27 (0.08) & 0.28 (0.09) & 0.80 (0.01) & 0.27 (0.08) \\ \hline

\end{tabular}
\end{table}
\FloatBarrier  

\newpage
\FloatBarrier  
\subsubsection{Individual $X_1$ Scores and Loadings}\label{appendix:mr.sim.table.indiv.1}
Summary statistics of the chordal distance between true and estimated individual scores and individual loadings for data set $\bX_1$ when \textbf{\underline{data are}} simulated from the generative ProJIVE model. The lowest average chordal norm is displayed in bold print to indicate which method performed the best. Figures 3 and 4 in the main manuscript display box plots from the same simulations, where Figure 3 corresponds to rows with $r_J=1$, and Figure 4, those with $r_J=3$.

\begin{table}[h]
\begin{tabular}{|p{.2in}|c|c|c|c|ccccc|}
\hline
\multirow{2}{*}{} & $r_J$ & $P_2$ & $R_{J1}^2$ & $R_{J2}^2$ & ProJIVE & AJIVE & R.JIVE & GIPCA & D-CCA \\ \hline
 & \multicolumn{3}{l}{} & & \multicolumn{5}{c|}{Mean (SD)}\\ \hline


\parbox[t]{1mm}{\multirow{16}{*}{\rotatebox[origin=c]{90}{Individual $X_1$ Scores}}} 
& \multirow{8}{*}{1} & \multirow{4}{*}{20} & \multirow{2}{*}{0.1} & 0.1 
& \textbf{0.53 (0.02)} & \textbf{0.53 (0.02)} & 0.54 (0.02) & 0.66 (0.11) & \textbf{0.53 (0.02)} \\ \cline{5-10}

& & & & 0.5 
& \textbf{0.53 (0.02)} & \textbf{0.53 (0.02)} & 0.58 (0.05) & 0.54 (0.03) & \textbf{0.53 (0.02)} \\ \cline{4-10}

& & & \multirow{2}{*}{0.5} & 0.1
& 0.38 (0.02) & 0.45 (0.06) & 0.39 (0.02) & 0.42 (0.06) & \textbf{0.38 (0.02)} \\ \cline{5-10}

& & & & 0.5
& \textbf{0.38 (0.03)} & \textbf{0.38 (0.02)} & \textbf{0.38 (0.02)} & 0.39 (0.06) & \textbf{0.38 (0.02)} \\ \cline{3-10}

& & \multirow{4}{*}{200} & \multirow{2}{*}{0.1} & 0.1
& \textbf{0.53 (0.02)} & \textbf{0.53 (0.02)} & \textbf{0.53 (0.02)} & 0.72 (0.05) & \textbf{0.53 (0.02)} \\ \cline{5-10}

& & & & 0.5
& \textbf{0.53 (0.02)} & \textbf{0.53 (0.02)} & 0.58 (0.06) & 0.53 (0.03) & \textbf{0.53 (0.02)} \\ \cline{4-10}

& & & \multirow{2}{*}{0.5} & 0.1
& \textbf{0.37 (0.02)} & 0.38 (0.02) & 0.38 (0.02) & 0.66 (0.13) & \textbf{0.38 (0.02)} \\ \cline{5-10}

& & & & 0.5
& \textbf{0.37 (0.01)} & 0.38 (0.02) & \textbf{0.37 (0.02)} & 0.39 (0.07) & \textbf{0.37 (0.02)} \\ \cline{2-10}

& \multirow{8}{*}{3} & \multirow{4}{*}{20} & \multirow{2}{*}{0.1} & 0.1
& 0.57 (0.06) & 0.70 (0.06) & 0.57 (0.08) & 0.83 (0.02) & \textbf{0.54 (0.02)} \\ \cline{5-10}

& & & & 0.5
& \textbf{0.54 (0.04)} & 0.67 (0.05) & 0.70 (0.10) & 0.80 (0.06) & \textbf{0.53 (0.02)} \\ \cline{4-10}

& & & \multirow{2}{*}{0.5} & 0.1
& 0.51 (0.13) & 0.65 (0.09) & \textbf{0.41 (0.03)} & 0.79 (0.04) & 0.67 (0.05) \\ \cline{5-10}

& & & & 0.5
& \textbf{0.39 (0.04)} & 0.42 (0.04) & \textbf{0.39 (0.02)} & 0.77 (0.02) & \textbf{0.39 (0.02)} \\ \cline{3-10}

& & \multirow{4}{*}{200} & \multirow{2}{*}{0.1} & 0.1
& \textbf{0.53 (0.02)} & 0.68 (0.05) & 0.82 (0.06) & \textbf{0.53 (0.02)} & \textbf{0.53 (0.02)} \\ \cline{5-10}

& & & & 0.5
& \textbf{0.53 (0.02)} & 0.64 (0.06) & 0.76 (0.08) & \textbf{0.53 (0.02)} & \textbf{0.53 (0.02)} \\ \cline{4-10}

& & & \multirow{2}{*}{0.5} & 0.1
& \textbf{0.39 (0.02)} & 0.52 (0.08) & 0.53 (0.15) & 0.59 (0.09) & \textbf{0.39 (0.02)} \\ \cline{5-10}

& & & & 0.5
& \textbf{0.37 (0.02)} & 0.39 (0.03) & 0.40 (0.07) & 0.45 (0.09) & \textbf{0.38 (0.02)} \\ \hline


\parbox[t]{1mm}{\multirow{16}{*}{\rotatebox[origin=c]{90}{Individual Loadings $X_1$}}}
& \multirow{8}{*}{1} & \multirow{4}{*}{20} & \multirow{2}{*}{0.1} & 0.1
& \textbf{0.11 (0.02)} & 0.12 (0.02) & 0.12 (0.02) & 0.42 (0.24) & \textbf{0.11 (0.02)} \\ \cline{5-10}

& & & & 0.5
& \textbf{0.11 (0.02)} & 0.12 (0.02) & 0.18 (0.11) & 0.13 (0.08) & \textbf{0.12 (0.02)} \\ \cline{4-10}

& & & \multirow{2}{*}{0.5} & 0.1
& \textbf{0.08 (0.02)} & 0.20 (0.10) & 0.12 (0.06) & 0.32 (0.15) & 0.58 (0.13) \\ \cline{5-10}

& & & & 0.5
& \textbf{0.07 (0.01)} & \textbf{0.06 (0.01)} & \textbf{0.07 (0.01)} & 0.12 (0.12) & \textbf{0.06 (0.01)} \\ \cline{3-10}

& & \multirow{4}{*}{200} & \multirow{2}{*}{0.1} & 0.1
& \textbf{0.11 (0.02)} & 0.12 (0.02) & 0.12 (0.02) & 0.60 (0.14) & \textbf{0.12 (0.02)} \\ \cline{5-10}

& & & & 0.5
& \textbf{0.11 (0.02)} & 0.12 (0.02) & 0.18 (0.12) & 0.12 (0.06) & \textbf{0.12 (0.02)} \\ \cline{4-10}

& & & \multirow{2}{*}{0.5} & 0.1
& \textbf{0.06 (0.01)} & \textbf{0.06 (0.01)} & 0.07 (0.01) & 0.57 (0.23) & \textbf{0.06 (0.01)} \\ \cline{5-10}

& & & & 0.5
& \textbf{0.06 (0.01)} & \textbf{0.06 (0.01)} & \textbf{0.06 (0.01)} & 0.09 (0.13) & \textbf{0.06 (0.01)} \\ \cline{2-10}

& \multirow{8}{*}{3} & \multirow{4}{*}{20} & \multirow{2}{*}{0.1} & 0.1
& \textbf{0.19 (0.12)} & 0.33 (0.12) & \textbf{0.19 (0.16)} & 0.65 (0.08) & \textbf{0.12 (0.02)} \\ \cline{5-10}

& & & & 0.5
& \textbf{0.13 (0.09)} & 0.30 (0.12) & 0.43 (0.21) & 0.53 (0.16) & \textbf{0.12 (0.02)} \\ \cline{4-10}

& & & \multirow{2}{*}{0.5} & 0.1
& 0.23 (0.17) & 0.40 (0.19) & \textbf{0.14 (0.05)} & 0.70 (0.07) & \textbf{0.58 (0.11)} \\ \cline{5-10}

& & & & 0.5
& \textbf{0.07 (0.03)} & 0.09 (0.03) & \textbf{0.07 (0.02)} & 0.65 (0.07) & \textbf{0.07 (0.02)} \\ \cline{3-10}

& & \multirow{4}{*}{200} & \multirow{2}{*}{0.1} & 0.1
& \textbf{0.11 (0.02)} & 0.31 (0.13) & 0.62 (0.14) & \textbf{0.12 (0.02)} & \textbf{0.12 (0.02)} \\ \cline{5-10}

& & & & 0.5
& \textbf{0.11 (0.02)} & 0.28 (0.11) & 0.54 (0.17) & \textbf{0.12 (0.02)} & \textbf{0.13 (0.02)} \\ \cline{4-10}

& & & \multirow{2}{*}{0.5} & 0.1
& \textbf{0.07 (0.01)} & 0.14 (0.06) & 0.26 (0.21) & 0.27 (0.16) & \textbf{0.07 (0.02)} \\ \cline{5-10}

& & & & 0.5
& \textbf{0.06 (0.01)} & \textbf{0.08 (0.02)} & 0.10 (0.10) & 0.15 (0.11) & \textbf{0.07 (0.01)}\\ \hline

\end{tabular}
\end{table}
\FloatBarrier  

\newpage
\FloatBarrier  
\subsubsection{Individual $X_2$ Scores and Loadings}\label{appendix:mr.sim.table.indiv.2}
Summary statistics of the chordal distance between true and estimated individual scores and individual loadings for data set $\bX_2$ when \textbf{\underline{data are}} simulated from the generative ProJIVE model. The lowest average chordal norm is displayed in bold print to indicate which method performed the best. Figures 3 and 4 in the main manuscript display box plots from the same simulations, as in section D.4.3 above.
\begin{table}[h]
\begin{tabular}{|p{.2in}|c|c|c|c|ccccc|}
\hline
\multirow{2}{*}{} & $r_J$ & $P_2$ & $R_{J1}^2$ & $R_{J2}^2$ & ProJIVE & AJIVE & R.JIVE & GIPCA & D-CCA \\ \hline
& \multicolumn{3}{l}{} & & \multicolumn{5}{c|}{Mean (SD)}\\ \hline

\parbox[t]{1mm}{\multirow{16}{*}{\rotatebox[origin=c]{90}{Individual $X_2$ Scores}}}

& \multirow{8}{*}{1} & \multirow{4}{*}{20} & \multirow{2}{*}{0.1} & 0.1 
& \textbf{0.53 (0.02)} & 0.54 (0.02) & 0.54 (0.02) & 0.67 (0.11) & 0.54 (0.02) \\\cline{5-10}

& & & & 0.5 
& \textbf{0.38 (0.02)} & 0.45 (0.07) & 0.48 (0.09) & 0.42 (0.06) & 0.66 (0.06) \\\cline{4-10}

& & & \multirow{2}{*}{0.5} & 0.1 
& \textbf{0.53 (0.02)} & \textbf{0.53 (0.02)} & \textbf{0.53 (0.02)} & 0.54 (0.04) & \textbf{0.53 (0.02)} \\\cline{5-10}

& & & & 0.5 
& \textbf{0.38 (0.03)} & \textbf{0.38 (0.03)} & \textbf{0.38 (0.02)} & 0.39 (0.05) & \textbf{0.38 (0.02)} \\\cline{3-10}

& & \multirow{4}{*}{200} & \multirow{2}{*}{0.1} & 0.1 
& \textbf{0.20 (0.01)} & \textbf{0.20 (0.01)} & 0.21 (0.01) & 0.70 (0.05) & \textbf{0.20 (0.01)} \\\cline{5-10}

& & & & 0.5 
& \textbf{0.13 (0.01)} & 0.27 (0.12) & 0.33 (0.15) & 0.16 (0.03) & 0.69 (0.02) \\\cline{4-10}

& & & \multirow{2}{*}{0.5} & 0.1 
& \textbf{0.20 (0.01)} & \textbf{0.20 (0.01)} & 0.21 (0.01) & 0.53 (0.18) & \textbf{0.20 (0.01)} \\\cline{5-10}

& & & & 0.5 
& \textbf{0.13 (0.00)} & 0.13 (0.01) & 0.14 (0.01) & 0.15 (0.03) & 0.13 (0.01) \\\cline{2-10}

& \multirow{8}{*}{3} & \multirow{4}{*}{20} & \multirow{2}{*}{0.1} & 0.1 
& 0.56 (0.06) & 0.70 (0.06) & 0.59 (0.10) & 0.84 (0.03) & \textbf{0.54 (0.02)} \\\cline{5-10}

& & & & 0.5 
& 0.47 (0.11) & 0.66 (0.09) & \textbf{0.41 (0.03)} & 0.79 (0.03) & 0.68 (0.05) \\\cline{4-10}

& & & \multirow{2}{*}{0.5} & 0.1 
& \textbf{0.53 (0.02)} & 0.67 (0.06) & 0.77 (0.09) & 0.80 (0.07) & \textbf{0.53 (0.02)} \\\cline{5-10}

& & & & 0.5 
& \textbf{0.38 (0.02)} & 0.42 (0.04) & 0.39 (0.04) & 0.77 (0.02) & 0.39 (0.02) \\\cline{3-10}

& & \multirow{4}{*}{200} & \multirow{2}{*}{0.1} & 0.1 
& 0.45 (0.18) & 0.47 (0.10) & \textbf{0.21 (0.01)} & 0.99 (0.01) & \textbf{0.21 (0.01)} \\\cline{5-10}

& & & & 0.5 
& 0.45 (0.18) & 0.55 (0.17) & \textbf{0.16 (0.02)} & 0.99 (0.01) & 0.69 (0.02) \\\cline{4-10}

& & & \multirow{2}{*}{0.5} & 0.1 
& 0.30 (0.17) & 0.39 (0.11) & 0.22 (0.07) & 0.90 (0.07) & \textbf{0.20 (0.01)} \\\cline{5-10}

& & & & 0.5 
& \textbf{0.15 (0.01)} & 0.16 (0.02) & 0.16 (0.05) & 0.95 (0.06) & \textbf{0.14 (0.01)} \\ \hline

\parbox[t]{1mm}{\multirow{16}{*}{\rotatebox[origin=c]{90}{Individual Loadings $X_2$}}}

& \multirow{8}{*}{1} & \multirow{4}{*}{20} & \multirow{2}{*}{0.1} & 0.1 
& \textbf{0.11 (0.02)} & 0.12 (0.03) & 0.12 (0.03) & 0.42 (0.24) & \textbf{0.11 (0.02)} \\\cline{5-10}

& & & & 0.5 
& \textbf{0.08 (0.02)} & 0.21 (0.11) & 0.32 (0.18) & 0.34 (0.14) & 0.58 (0.12) \\\cline{4-10}

& & & \multirow{2}{*}{0.5} & 0.1 
& \textbf{0.11 (0.02)} & 0.13 (0.03) & \textbf{0.11 (0.02)} & 0.13 (0.08) & 0.12 (0.03) \\\cline{5-10}

& & & & 0.5 
& \textbf{0.06 (0.01)} & \textbf{0.06 (0.01)} & \textbf{0.06 (0.01)} & 0.12 (0.11) & \textbf{0.06 (0.01)} \\\cline{3-10}

& & \multirow{4}{*}{200} & \multirow{2}{*}{0.1} & 0.1 
& \textbf{0.09 (0.01)} & 0.10 (0.01) & 0.10 (0.01) & 0.66 (0.07) & 0.10 (0.01) \\\cline{5-10}

& & & & 0.5 
& \textbf{0.07 (0.02)} & 0.22 (0.14) & 0.35 (0.19) & 0.17 (0.08) & 0.69 (0.02) \\\cline{4-10}

& & & \multirow{2}{*}{0.5} & 0.1 
& \textbf{0.09 (0.00)} & \textbf{0.09 (0.00)} & \textbf{0.09 (0.00)} & 0.45 (0.20) & \textbf{0.09 (0.00)} \\\cline{5-10}

& & & & 0.5 
& \textbf{0.06 (0.00)} & 0.06 (0.01) & 0.06 (0.01) & 0.13 (0.07) & 0.06 (0.01) \\\cline{2-10}

& \multirow{8}{*}{3} & \multirow{4}{*}{20} & \multirow{2}{*}{0.1} & 0.1 
& 0.17 (0.11) & 0.34 (0.11) & 0.21 (0.19) & 0.66 (0.08) & \textbf{0.12 (0.02)} \\\cline{5-10}

& & & & 0.5 
& 0.18 (0.14) & 0.43 (0.18) & \textbf{0.14 (0.06)} & 0.70 (0.06) & 0.59 (0.10) \\\cline{4-10}

& & & \multirow{2}{*}{0.5} & 0.1 
& \textbf{0.11 (0.02)} & 0.31 (0.12) & 0.52 (0.19) & 0.53 (0.17) & \textbf{0.12 (0.02)} \\\cline{5-10}

& & & & 0.5 
& \textbf{0.07 (0.01)} & 0.09 (0.03) & \textbf{0.07 (0.06)} & 0.64 (0.08) & \textbf{0.07 (0.01)} \\\cline{3-10}

& & \multirow{4}{*}{200} & \multirow{2}{*}{0.1} & 0.1 
& 0.22 (0.17) & 0.13 (0.02) & \textbf{0.10 (0.01)} & 0.94 (0.05) & \textbf{0.10 (0.01)} \\\cline{5-10}

& & & & 0.5 
& 0.22 (0.15) & 0.46 (0.21) & \textbf{0.11 (0.04)} & 0.96 (0.03) & 0.69 (0.02) \\\cline{4-10}

& & & \multirow{2}{*}{0.5} & 0.1 
& \textbf{0.15 (0.13)} & 0.12 (0.02) & \textbf{0.10 (0.07)} & 0.75 (0.07) & \textbf{0.09 (0.00)} \\\cline{5-10}

& & & & 0.5 
& \textbf{0.06 (0.00)} & 0.06 (0.01) & 0.07 (0.04) & 0.88 (0.11) & \textbf{0.06 (0.00)} \\ \hline

\end{tabular}
\end{table}
\FloatBarrier

\section{Supplemental information for ADNI analysis}

\subsection{Details of the participants}\label{sec:participantdetails}

There are approximately $n=1600$ participants in the TADPOLE datasets. We examined data from $n=587$ participants for whom the full battery of cognitive and behavioral assessments were available. We limit our analysis to observations taken at the participants' 6-month follow-up visit to minimize missingness and maximize sample size. Both ADNI-2 and ADNI-GO limit newly recruited participants to those between 55-90 years of age and require that each has an English- or Spanish-speaking study partner to provide independent evaluation of patient functioning. ADNI-GO inclusion criteria required that participants who rolled over from ADNI-1 were either diagnosed as cognitively normal (CN) or having mild cognitive impairment (MCI) at baseline. Newly recruited participants in ADNI-GO were all diagnosed with early mild cognitive impairment (EMCI). ADNI-2 participants were either rolled-over from ADNI-1/ADNI-GO or newly recruited participants diagnosed as CN, EMCI, LMCI (late mild cognitive impairment), or AD. We combine EMCI and LMCI participants with MCI and only use diagnoses from the 6-month follow-up visit. 

\begin{table}
\caption{Summary statistics for selected covariates of participants in ADNI-GO and ADNI2.} \label{tbl:summary.stats}
\begin{tabular}{llllll}
          & AD (N=88)       & MCI (N=340)     & CN (N=159)      & Total (N=587)   & p value         \\
\hline
& \multicolumn{4}{c}{Mean (S.D.) or N (\%)} \\
\hline
Age       &                 &                 &                 &                 & 0.006           \\
& 74.0 (7.92)  & 71.5 (7.57)  & 72.8 (5.85)  & 72.2 (7.25)  &  \\
Gender  &                 &                 &                 &                 & 0.006           \\
\;\; Female    & 28 (31.8\%)     & 150 (44.1\%)    & 84 (52.8\%)     & 262 (44.6\%)    &  \\
ApoE4 &                 &                 &                 &                 & \textless 0.001 \\
\;\; 0   & 21 (23.9\%)     & 178 (52.4\%)    & 111 (69.8\%)    & 310 (52.8\%)    &  \\
\;\; 1   & 45 (51.1\%)     & 126 (37.1\%)    & 46 (28.9\%)     & 217 (37.0\%)    &  \\
\;\; 2   & 22 (25.0\%)     & 36 (10.6\%)     & 2 (1.3\%)       & 60 (10.2\%)    &                
\end{tabular}
\end{table}
The cognitive measures include the following: 1) Clinical Dementia Rating - Sum of Boxes (CDR-SB) \citep{morris1997clinical}; 2)  Alzheimer's Disease Assessment Scale - Cognition (ADAS) \citep{rosen1984new}, in which the 11-item and 13-item scores were used as separate variables; 3) the Mini-Mental State Exam (MMSE) \citep{folstein1975mini}; 4) Rey's Auditory Verbal Learning Test (RAVLT) \citep{rey1941examen,rey1958examen}, in which Forgetting, Immediate, and Learning sub-scales were used as separate variables; 5) Montreal Cognitive Assessment (MOCA) \citep{nasreddine2005montreal}; and 6) Everyday Cognition (ECOG) \citep{farias2008measurement}, in which  seven pairs of sub-scales were used and each pair includes a response from the participant and his or her study partner. Table \ref{tbl:app2.cog.summary} provides summary statistics for the cognitive measures that were included in our ProJIVE analyses of TADPOLE/ADNI data. 

\begin{table}[htbp!]
\caption{Summary statistics for Cognition Measures.}
\label{tbl:app2.cog.summary}
\centering
\begin{tabular}{lp{0.75in}p{0.75in}p{0.75in}p{0.75in}l}
\hline
& AD (N=88) & MCI (N=340)   & CN (N=159)    & Total (N=587) &  \\
\hline
Cognition Score & \multicolumn{4}{c}{Mean (SD)} &  p value \\ 
\hline
CDRSB &4.8 (1.84)     & 1.4 (0.97)   & 0.1 (0.33)   & 1.6 (1.79)   & \textless 0.001 \\
ADAS11 & 20.4 (6.52) & 9.1 (4.69)   & 5.5 (2.76)   & 9.8 (6.60)   &  \\
ADAS13 & 30.8 (8.06)    & 14.3 (7.03)  & 8.5 (4.35)   & 15.2 (9.63)  & \textless 0.001    \\
MMSE   & & & & &  \textless 0.001 \\
MMSE & 23.1 (3.05)    & 27.8 (2.00)  & 29.0 (1.14)  & 27.4 (2.74)  & \textless 0.001 \\
RAVLT (immediate) & 20.5 (6.18)    & 34.7 (10.84) & 42.9 (9.35)  & 34.8 (12.08) & \textless 0.001 \\
RAVLT (learning) & 1.3 (1.75)     & 4.4 (2.51)   & 5.5 (2.19)   & 4.3 (2.67)   & \textless 0.001  \\
RAVLT (forgetting) & 4.0 (1.61)     & 4.7 (2.45)   & 4.2 (2.80)   & 4.46 (2.46)   &  0.019 \\
MOCA & 17.8 (4.36)    & 23.5 (3.26)  & 25.9 (2.45)  & 23.3 (4.12)  &    \textless 0.001  \\
EcogPTMem & 2.4 (0.75)     & 2.2 (0.74)   & 1.6 (0.46)   & 2.0 (0.74)   & \textless 0.001  \\
EcogPTLang & 1.9 (0.79)     & 1.8 (0.64)   & 1.4 (0.39)   & 1.7 (0.65)   & \textless 0.001 \\
EcogPTVisspat & 1.7 (0.71)     & 1.4 (0.54)   & 1.1 (0.22)   & 1.4 (0.53)   & \textless 0.001 \\
EcogPTPlan & 1.6 (0.72)     & 1.5 (0.53)   & 1.1 (0.26)   & 1.4 (0.54)   & \textless 0.001 \\
EcogPTOrgan & 1.8 (0.75)     & 1.6 (0.61)   & 1.3 (0.34)   & 1.5    (0.60)   & \textless 0.001 \\
EcogPTDivatt &1.98 (0.81)     & 1.90 (0.76)   & 1.45 (0.58)   & 1.79 (0.75)   &  \textless 0.001 \\
EcogPTTotal &1.91 (0.66)     & 1.74 (0.53)   & 1.33 (0.30)   & 1.65 (0.54)   & \textless 0.001 \\
EcogSPMem & 3.32 (0.63)     & 2.09 (0.78)   & 1.29 (0.36)   & 2.06 (0.92)   & \textless 0.001 \\
EcogSPLang & 2.63 (0.75)     & 1.65 (0.68)   & 1.11 (0.21)   & 1.65 (0.77)   &  \textless 0.001 \\
EcogSPVisspat &2.47 (0.85)     & 1.41 (0.54)   & 1.07 (0.19)   & 1.48 (0.69)   &  \textless 0.001 \\
EcogSPPlan & 2.66 (0.90)     & 1.55 (0.68)   & 1.10 (0.32)   & 1.60 (0.81)   & \textless 0.001  \\
EcogSPOrgan & 2.90 (0.85)     & 1.63 (0.78)   & 1.12 (0.26)   & 1.68 (0.88)   & \textless 0.001 \\
EcogSPDivatt &3.09 (0.84)     & 1.88 (0.79)   & 1.24 (0.46)   & 1.89 (0.93)   &  \textless 0.001 \\
EcogSPTotal &2.84 (0.62)     & 1.71 (0.61)   & 1.15 (0.22)   & 1.73 (0.75)   &  \textless 0.001
\end{tabular}
\end{table}

The morphometry data set contains measures of cortical thickness (CT) and cortical surface area (SA) for each of the thirty-four Desikan cortical surface ROIs in each hemisphere processed using FreeSurfer \citep{desikan2006automated}. Intra-cranial volume is also included as a measure of cortical volume (CVol). The volumes of subcortical ROIs (both gray matter and white matter regions/structures) are labelled as subcortical volumes (SVol). These include 17 subcortical gray matter structures, 16 of which form eight hemispheric left-right pairs. The unpaired subcortical gray matter ROI is the brainstem. There are 13 remaining SVol measures which include the ventricles, cerebrospinal fluid (CSF), optic chiasm, corpus callosum, and others. In total, 245 measures of brain morphometry are included.  

\subsection{Scree plots for brain morphometry and cognition/behavior}
\begin{figure}
\centering
    \includegraphics[height = 0.48\textheight, width = 0.58\textwidth] {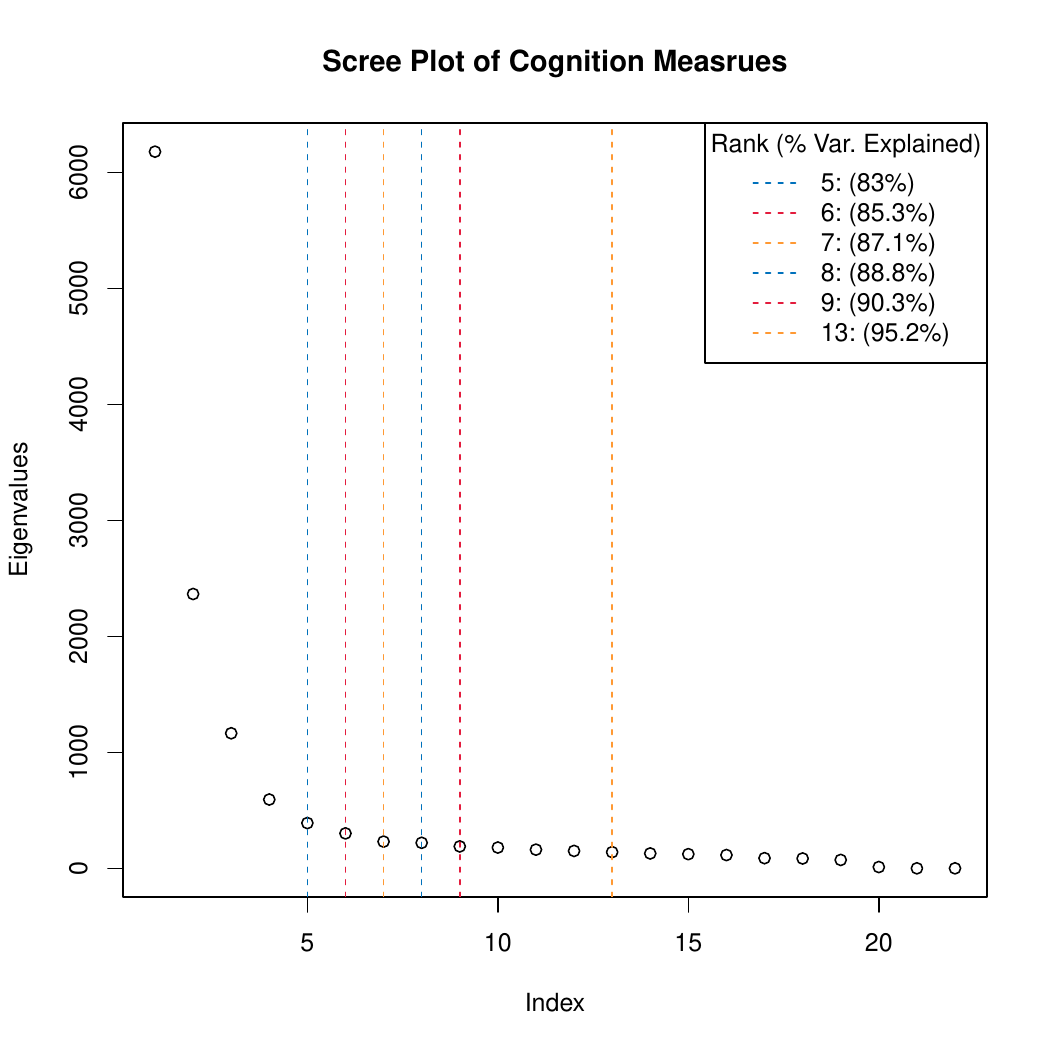}
    \includegraphics[height = 0.48\textheight, width = 0.58\textwidth] {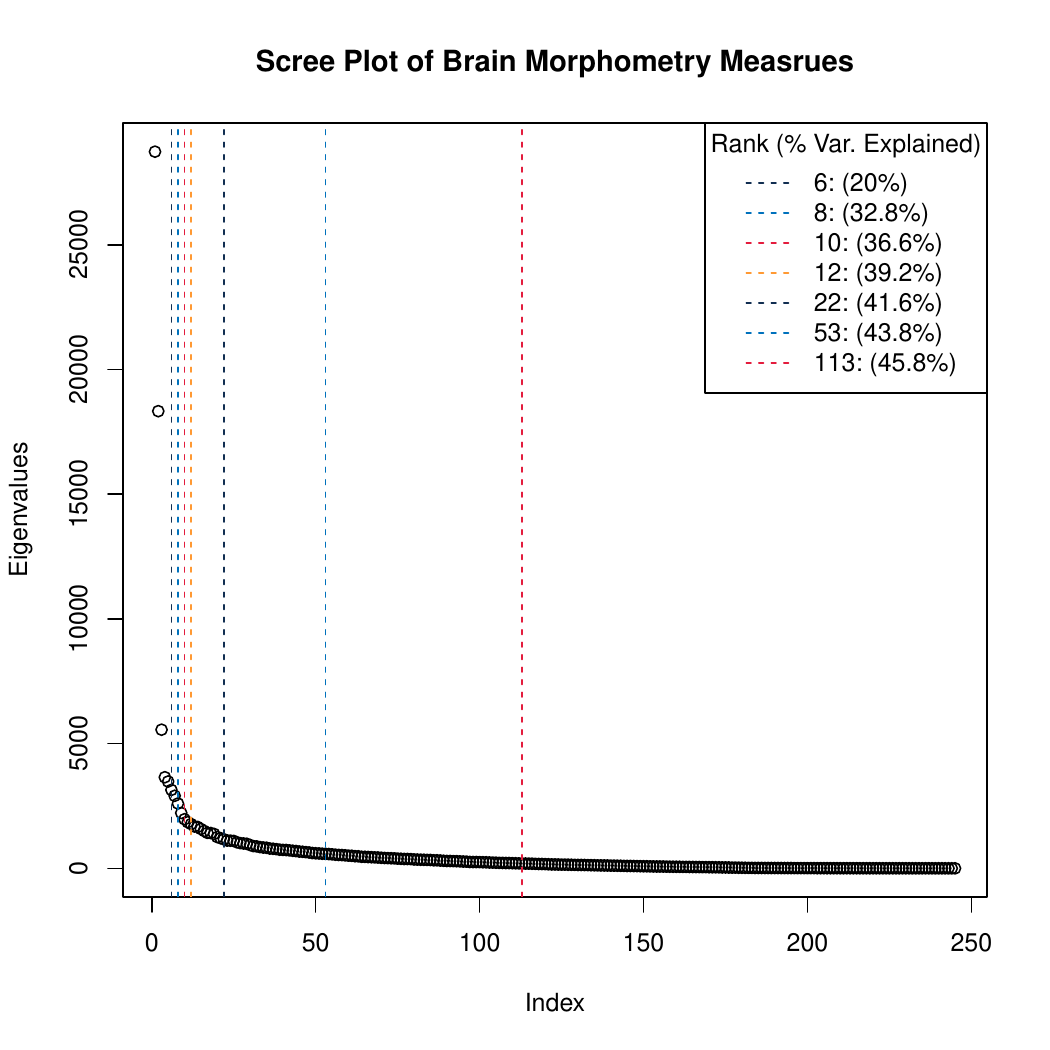}
    \caption{Scree plots shows which values were chosen as total signal ranks. Choices depended the `elbow' of the scree plot or by accounting for at least 80\%, 90\% ,or 95\% of total variance.}
    \label{fig:screeplots}
\end{figure}

The scree plots in Figure \ref{fig:screeplots} show the total rank choices for cognition (left) and brain morphometry (right) datasets. Dashed vertical lines correspond to the `elbow' of the scree plot (rank=5), the number of eigenvalues that account for at least 90\% of total variability (rank=9), and the number of eigenvalues accounting for at least 95\% of total variation (rank = 13).

With brain morphometry, two values were proposed as the `elbow' of the scree plot (rank = 10 or 22). The other two ranks were chosen to account for at least 80\% (rank = 53) and at least 95\% (rank = 113) of total variability. Figure \ref{fig:screeplots} shows the scree plot, each chosen rank and the corresponding percentage of total variation.

In our analysis, we elected to use rank 5 for cognition and rank 10 for brain morphometry based on our previous experience that choosing a lower rank can improve the selection of the joint rank and also lead to a simpler, more interpretable decomposition. 

\subsection{Individual subspaces}\label{supp.sec:indiv.loads}
\begin{figure}[htbp]
    \centering
     \includegraphics[]{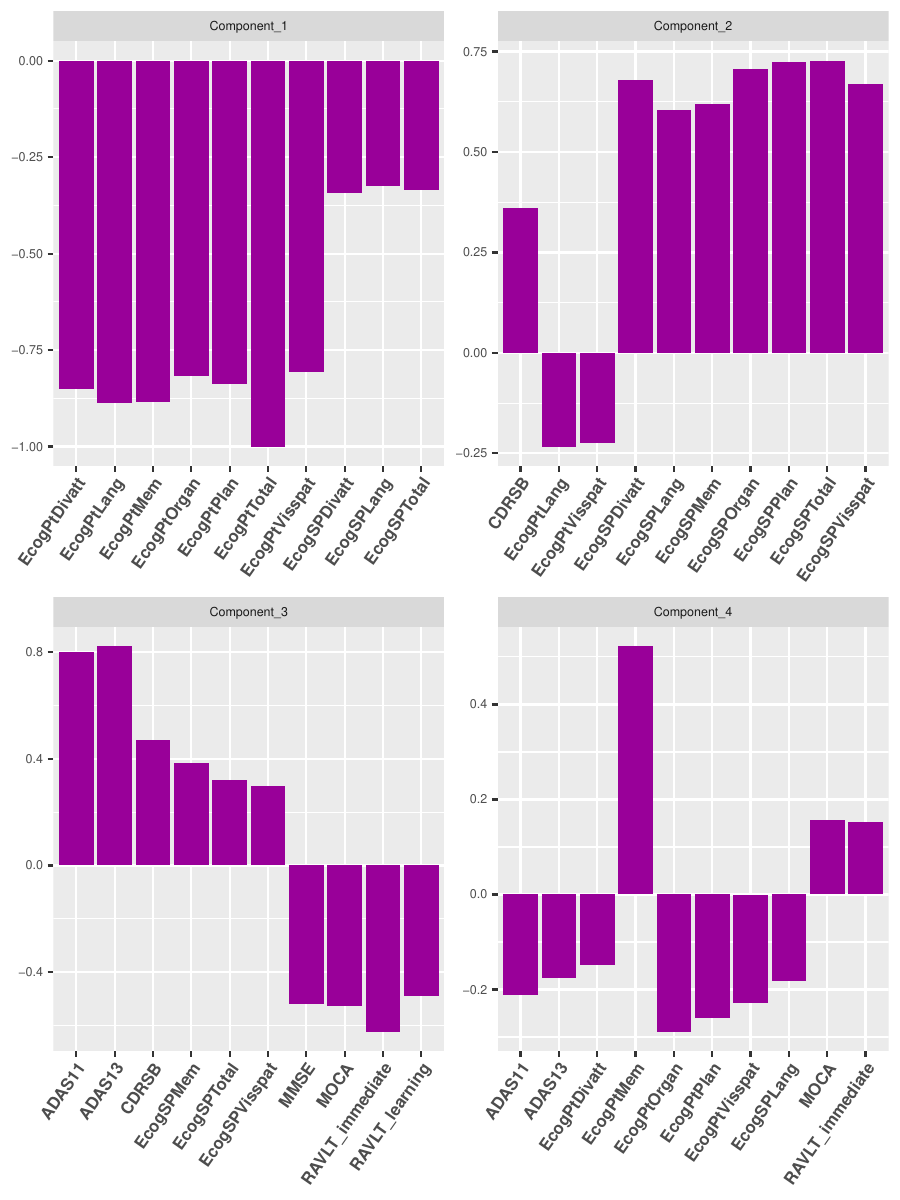} 
    \caption{$90^{th}$ percentile of the absolute value of loadings onto each of the four components that compose the individual cognition subspace estimated via ProJIVE in the setting $K=2$ (combining all brain morphometry measures as one dataset).}
    \label{fig:indiv.cognition.loads}
\end{figure}

\begin{figure}[htbp]
    \centering
     \includegraphics[]{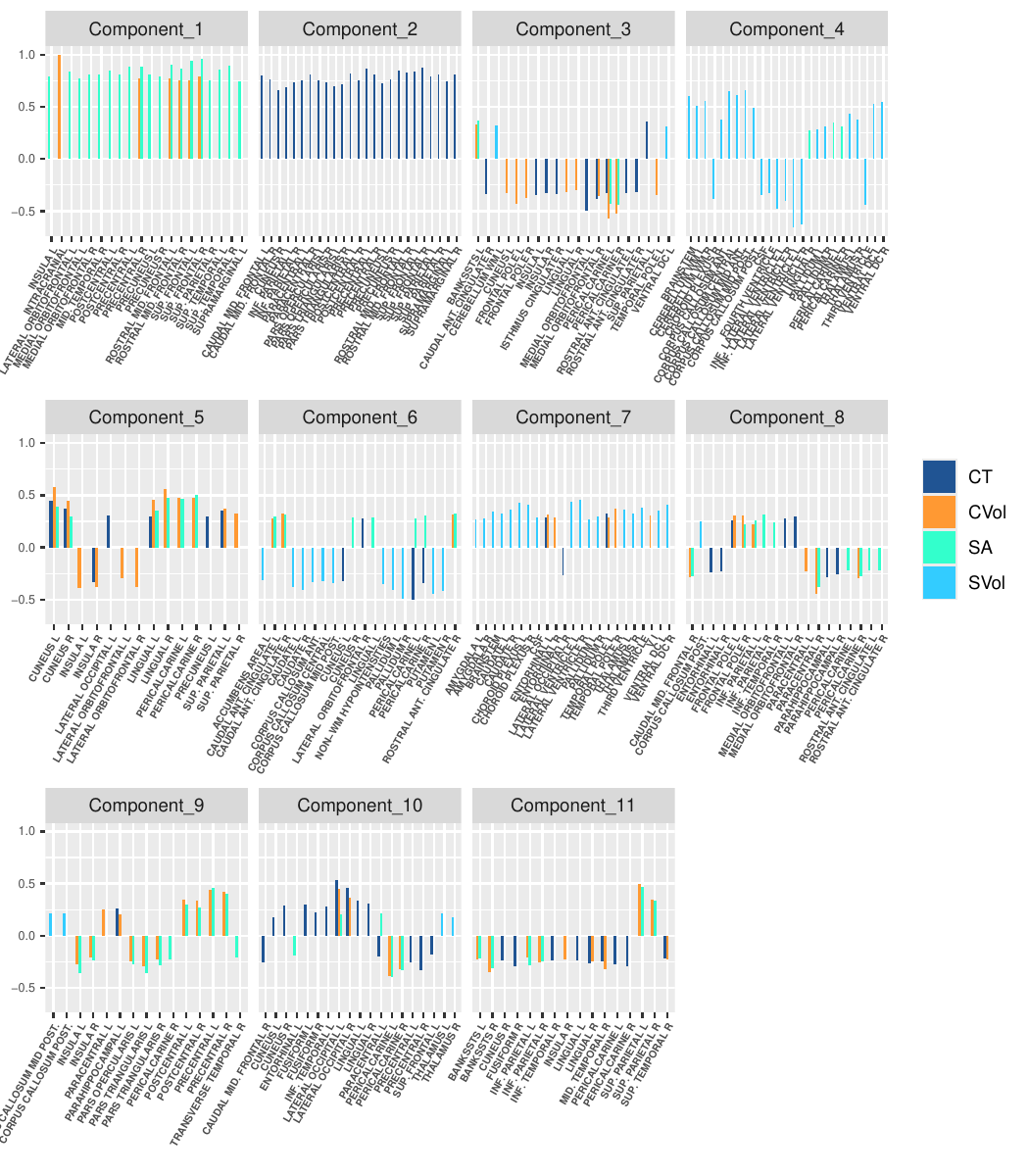} 
    \caption{$90^{th}$ percentile of the absolute value of loadings onto each of the nine components that compose the individual morphometry subspace estimated via ProJIVE in the setting $K=2$.}\label{fig:indiv.brain.loads}
\end{figure}

\begin{figure}[htbp]
    \centering
     \includegraphics[height = 0.93\textheight, width = 0.97\textwidth]{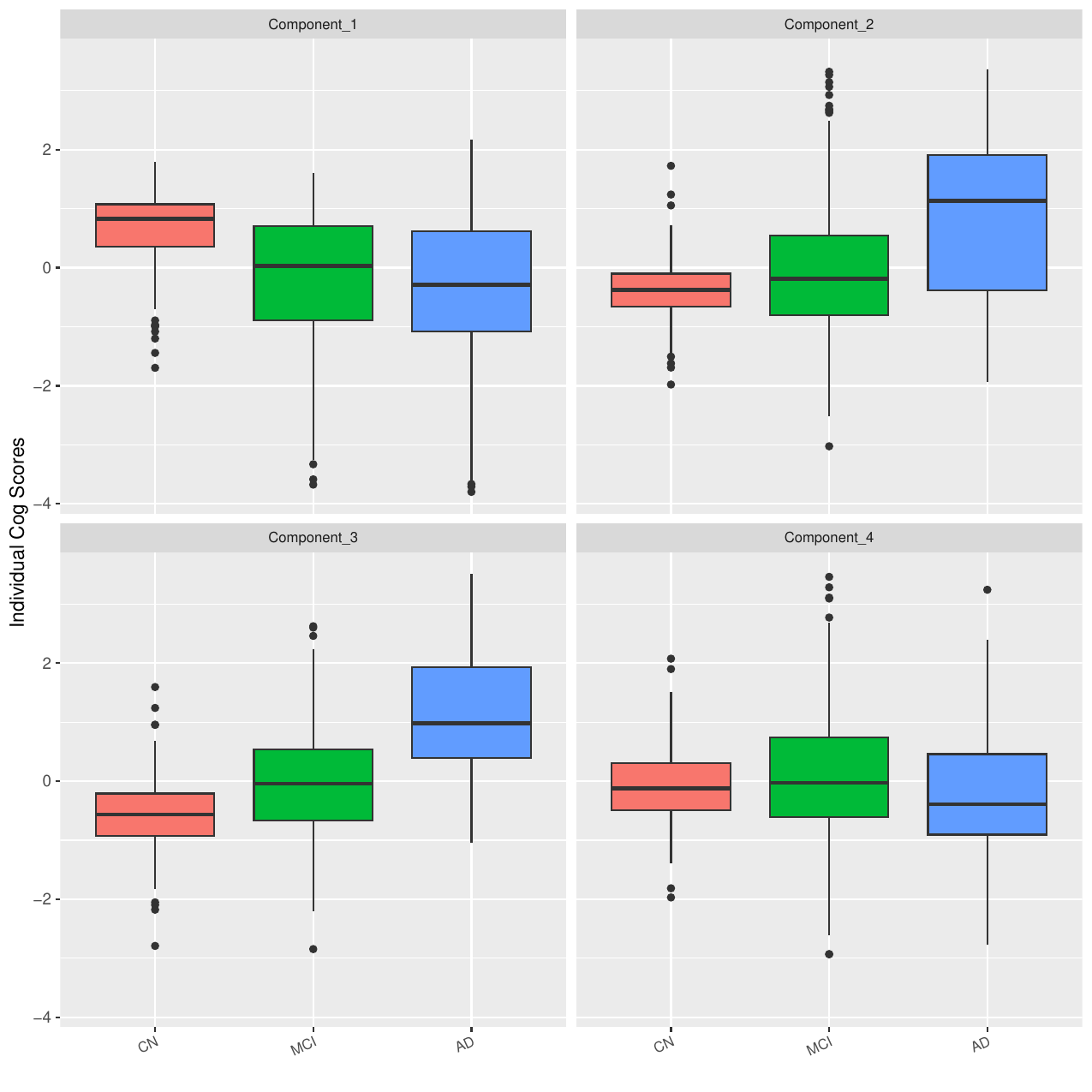} 
    \caption{Subject scores from each component of the individual cognition subspace estimated via ProJIVE in the setting $K=2$.}\label{fig:indiv.cognition.scores}
\end{figure}

\begin{figure}[htbp]
    \centering
     \includegraphics[height = 0.93\textheight, width = 0.97\textwidth]{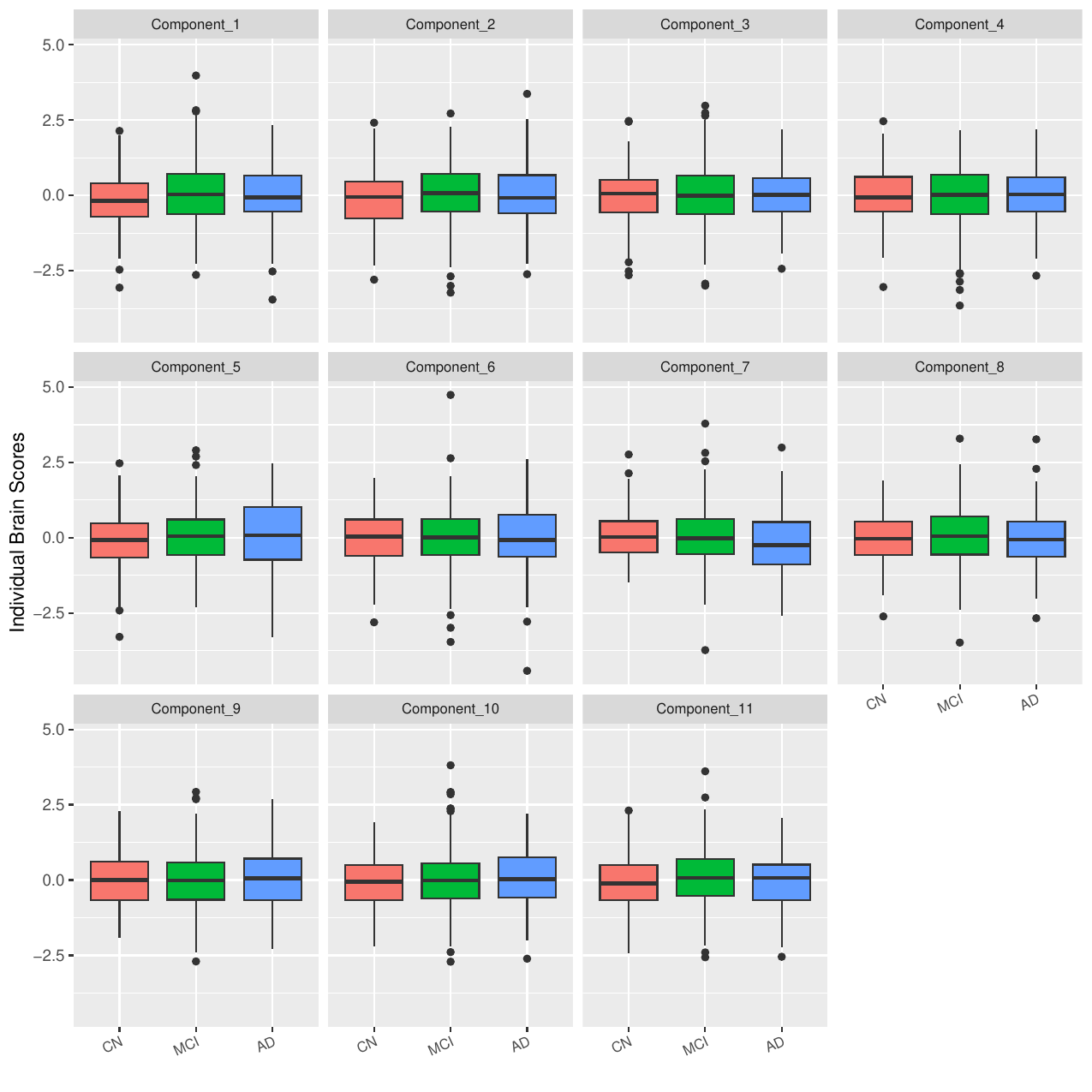} 
    \caption{Subject scores from each component of the individual brain morphometry subspace estimated via ProJIVE in the setting $K=2$.}
    \label{fig:indiv.brain.scores}
\end{figure}

Figures supporting the discussion of subject scores and variable loadings (Section 4.5 of the main manuscript) from each dataset's individual subspace are displayed in Figures \ref{fig:indiv.cognition.loads}, \ref{fig:indiv.brain.loads}, 
\ref{fig:indiv.cognition.scores}, and \ref{fig:indiv.brain.scores}. Cognition loadings are discussed in the main manuscript. Brain loadings tend to separate components by modality for some components; e.g., component 2 is dominated by CT and component 4, by SVol measures. Other individual brain loading components comprise fewer ROIs with multiple modalities at several. For example component 5 is dominated by SA and CVol in the same ROIs, including the insula and precentral gyrus.
    
\section{ProJIVE analysis of ADNI data with the specification of $K=5$ datasets}

The primary manuscript includes a brief review (Section 4.6) of ProJIVE as applied to the same ADNI/TADPOLE data when each set of brain morphometry measures (CT, CVol, SA, and SVol) were treated as separate datasets, resulting in a total of $K=5$ data sets. To complement this analysis, there is a brief discussion on choosing the features to comprise a ``dataset" (Section 5.)

\subsection{Joint Variable Loadings}
\begin{figure}[htbp]
    \centering
     \includegraphics[]{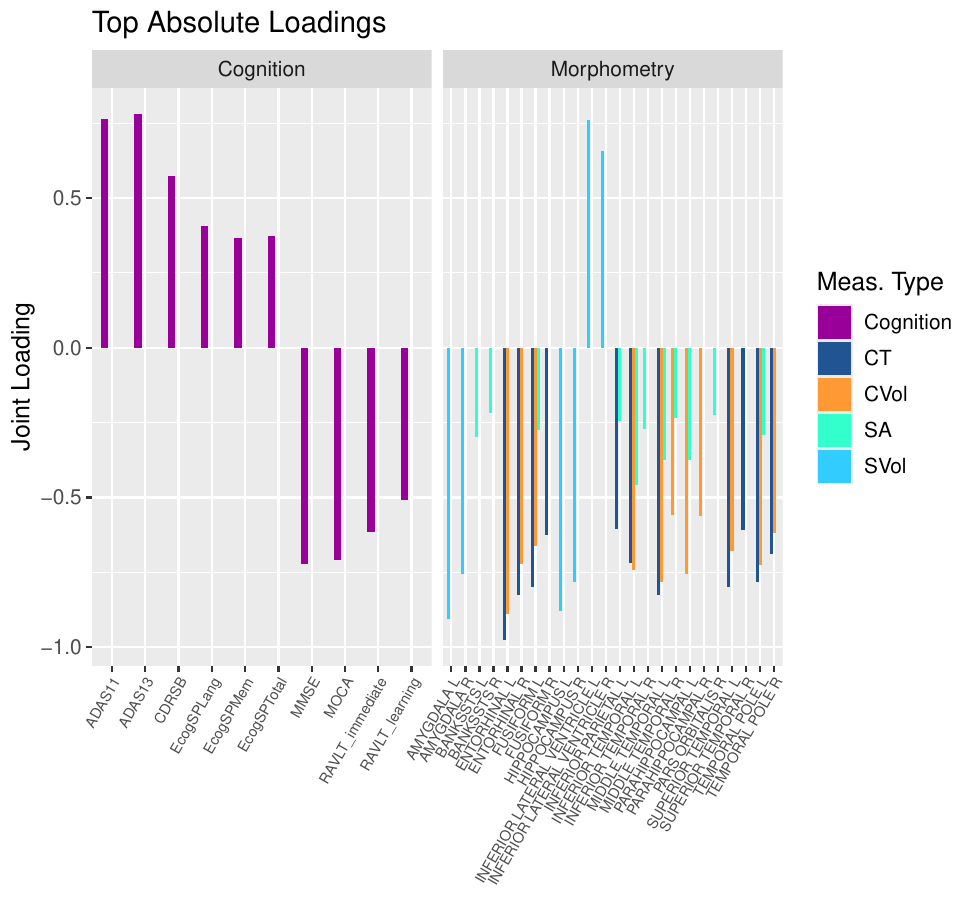} 
    \caption{Ten most extreme joint cognition loadings, six most extreme joint SVol loadings, and upper $85^{th}$ percentile, respectively, of the absolute joint CT, CVol, and SA loadings estimated via ProJIVE in the setting $K=5$ (treating each brain morphometry measure as a separate dataset).}\label{fig:jnt.loads.allK5}
\end{figure}

The top 10 variable loadings are displayed for cognition, and the top 15\% of joint variable loadings are displayed for cortical thickness, cortical volume, surface area, and subcortical volume (Figure \ref{fig:jnt.loads.allK5}). Notably, the distributions of the strongest variable loadings in the $K=5$ setting largely reflect findings from the $K=2$ setting. In cognition, the only difference in the pattern of these loadings is that EcogSPVisspat in the $K=2$ setting is replaced by the immediate-RAVLT score in the $K=5$ setting. For morphometry measures, a major difference lies in the presence of surface area (SA) loadings. However, recall that loadings are scaled by their maximum absolute value for a dataset. Since the joint SA loadings have a maximum absolute value of around 0.5, the strongest SA loadings are in the individual SA subspace, as in the $K=2$ case. 

\subsection{Joint subject scores}
\begin{figure}[htbp]
    \centering
    (a) \includegraphics[height=0.4\textheight, width = 0.4\textwidth]{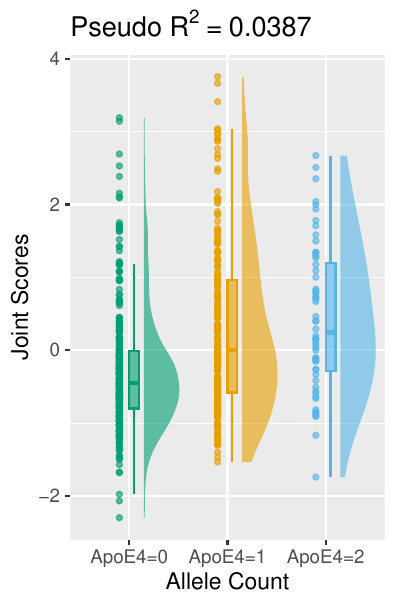} 
    (b) \includegraphics[height=0.4\textheight, width = 0.4\textwidth]{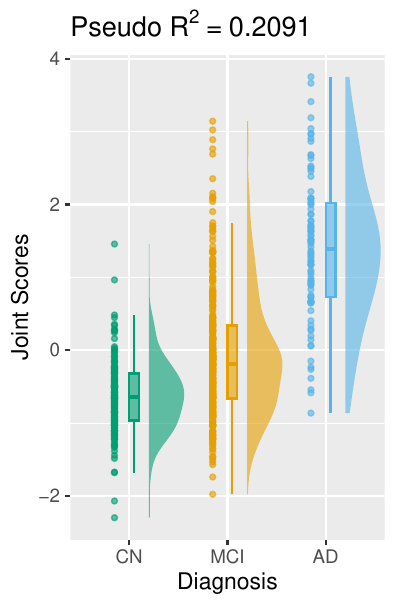}
    
    (c) \includegraphics[height=0.4\textheight, width = 0.4\textwidth]{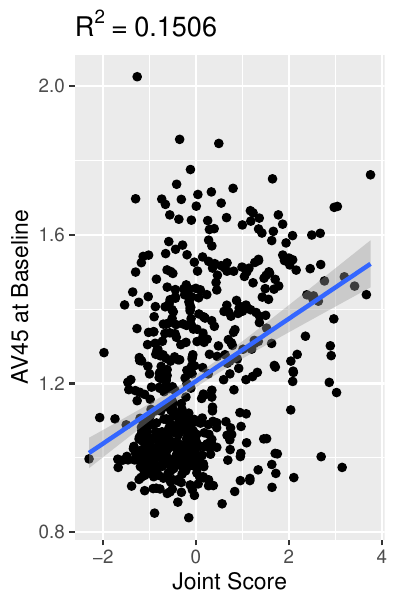}
    (d) \includegraphics[height=0.4\textheight, width = 0.4\textwidth]{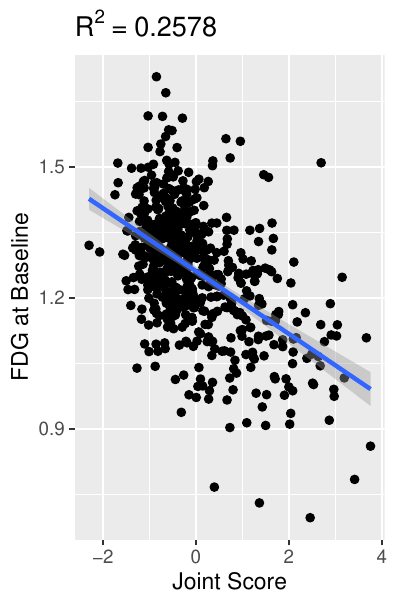} 
    \caption{Joint subject scores estimated via ProJIVE (for $K=5$ in which brain morphometry data are separated into four different datasets) show separation by (a) the count of ApoE4 allele counts and (b) diagnosis via raincloud plots. (c) Scatterplot of joint subject scores and AV45 at Baseline with OLS regression line and confidence interval. (d) Scatterplot of joint subject scores and FDG at Baseline with OLS regression line and confidence interval. Results resemble findings from $K=2$ in which all brain morphometry variables were combined into one dataset.}
    \label{fig:joint.scores.sep.morph}
\end{figure}

Joint subject scores from the $K=5$ case show similar associations with exogenous variables (i.e. PET biomarkers, diagnosis, and ApoE4) as in the $K=2$ case (Figure \ref{fig:joint.scores.sep.morph}).

\subsection{Individual variable loadings}\label{supp.sec:indiv.loads.K5}
\begin{figure}[htbp]
    \centering
     \includegraphics[]{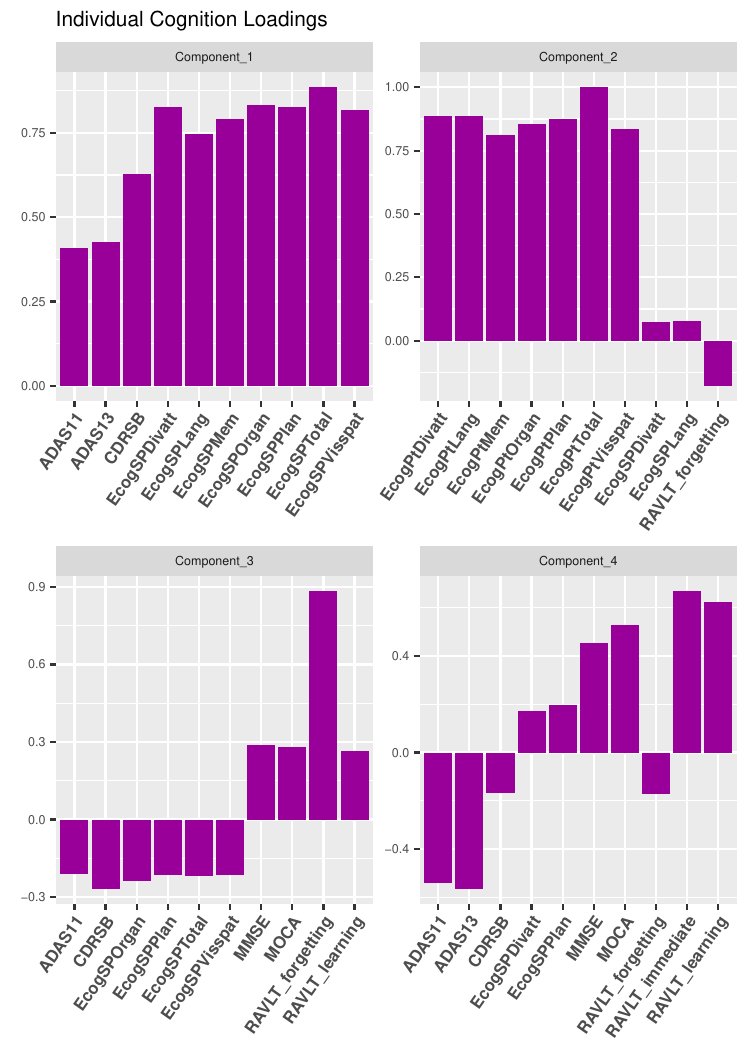} 
    \caption{subject scores individual cognition subspace estimated via ProJIVE in the setting $K=5$.}
    \label{fig:indiv.cog.scoresK5}
\end{figure}

Variable loadings onto the individual cognition subspace reflect somewhat divergent patterns for analyses parameterized  $K=5$ (Figure \ref{fig:indiv.cog.scoresK5}) and $K=2$ (Figure \ref{fig:indiv.cognition.scores}). Specifically, the strongest loadings for components 1 and 2 are for the Everyday Cognition (Ecog) measures in both settings. However, for $K=5$, the study partner measures dominate component one and patient measures dominate component 2, whereas the opposite holds for $K=2$. In component 3 of both analyses, the ADAS measures load in the opposite direction of MMSE, MOCA, and RAVLT measures. However, for $K=5$, RAVLT forgetting dominates component 3, similar to component 4 of the $K=2$ analyses. Lastly, component 4 loads strongly onto the RAVLT-intermediate and -learning, MMSE, and MOCA measures in opposing the direction to ADAS measures, similar to component 3 of the $K=2$ results. 

\begin{figure}[htbp]
    \centering
     (a) \includegraphics[height=0.46\textheight, width = 0.45\textwidth]{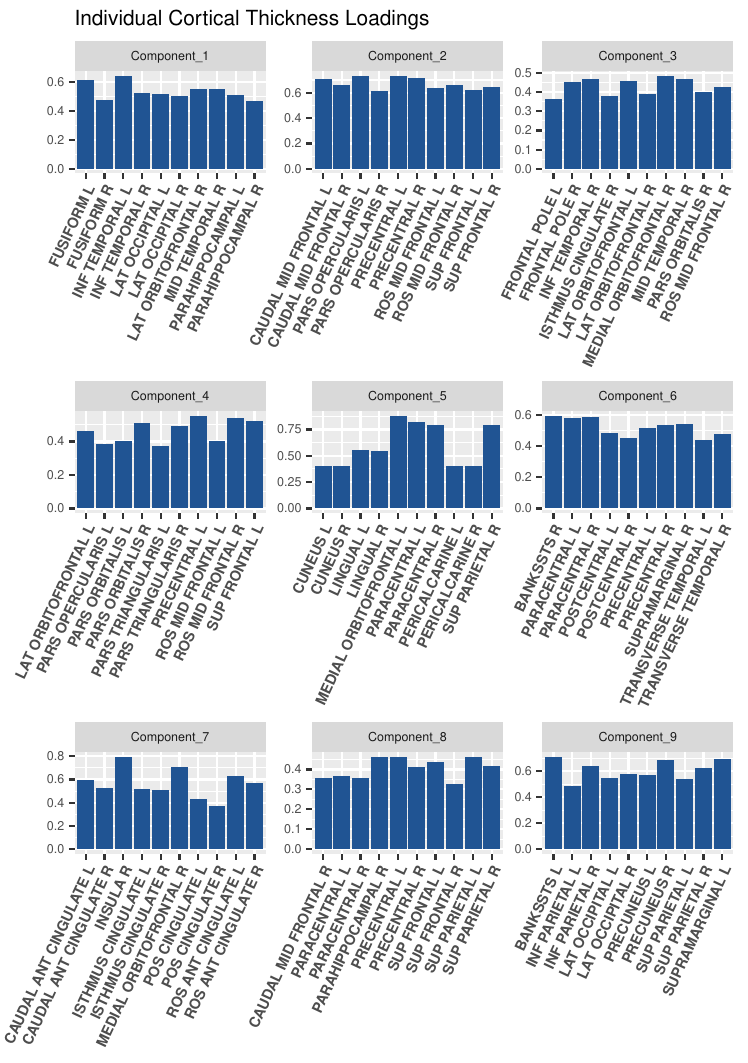}   
     (b) \includegraphics[height=0.46\textheight, width = 0.45\textwidth]{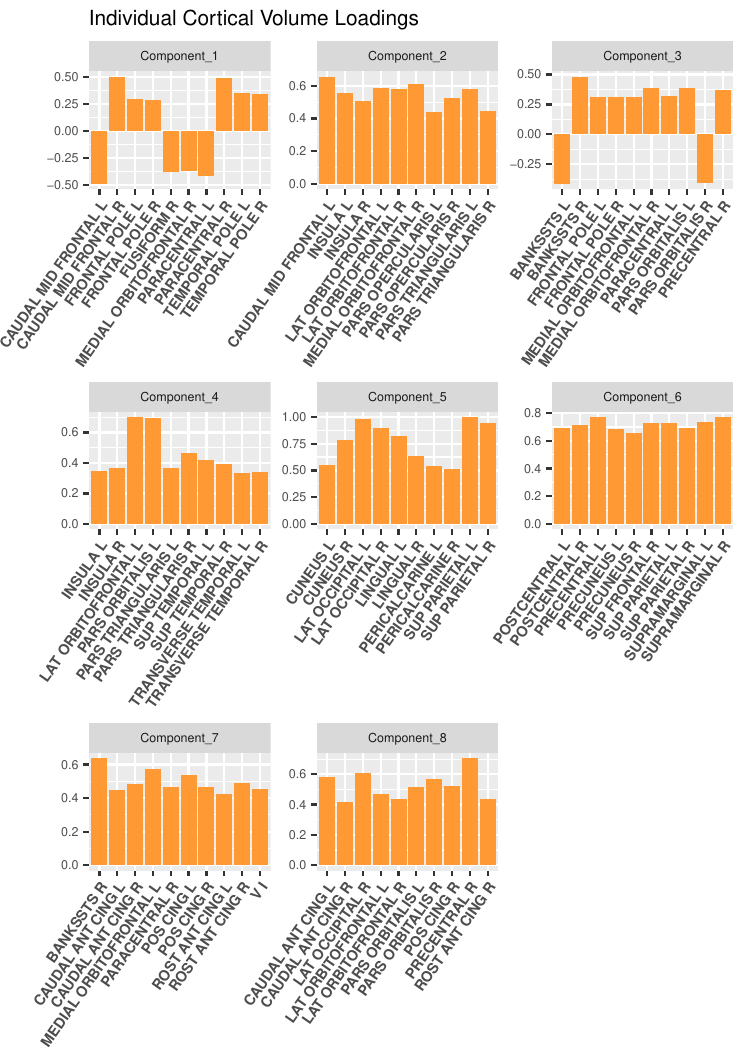} \\
     (c) \includegraphics[height=0.46\textheight, width = 0.45\textwidth]{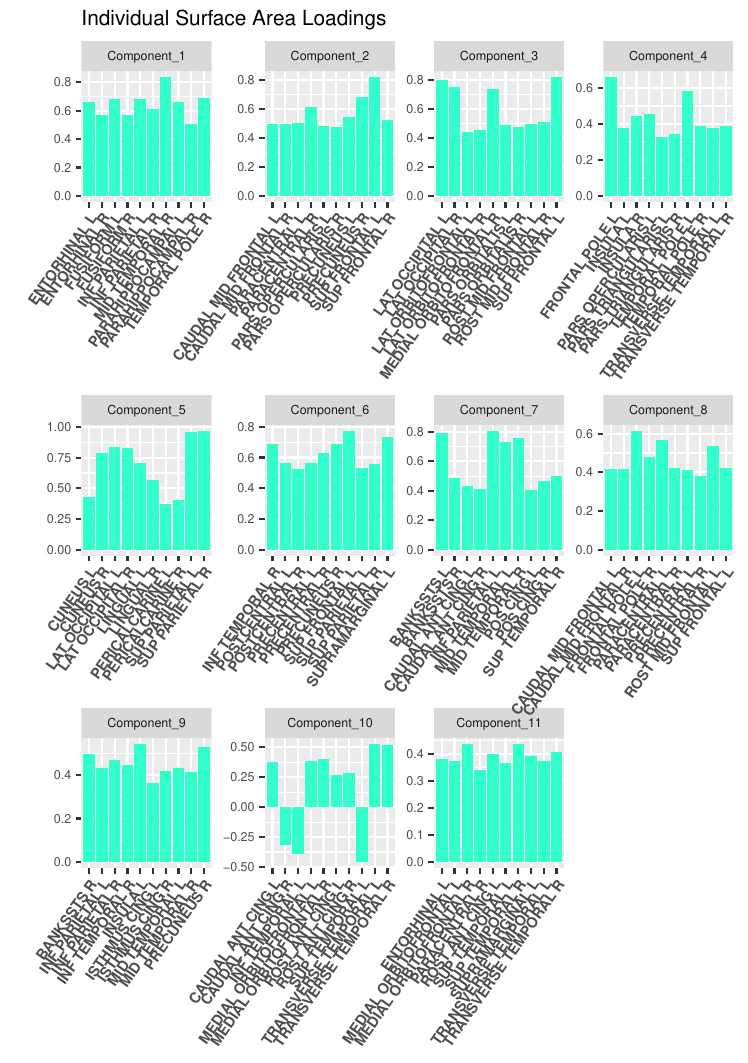}  
     (d) \includegraphics[height=0.46\textheight, width = 0.45\textwidth]{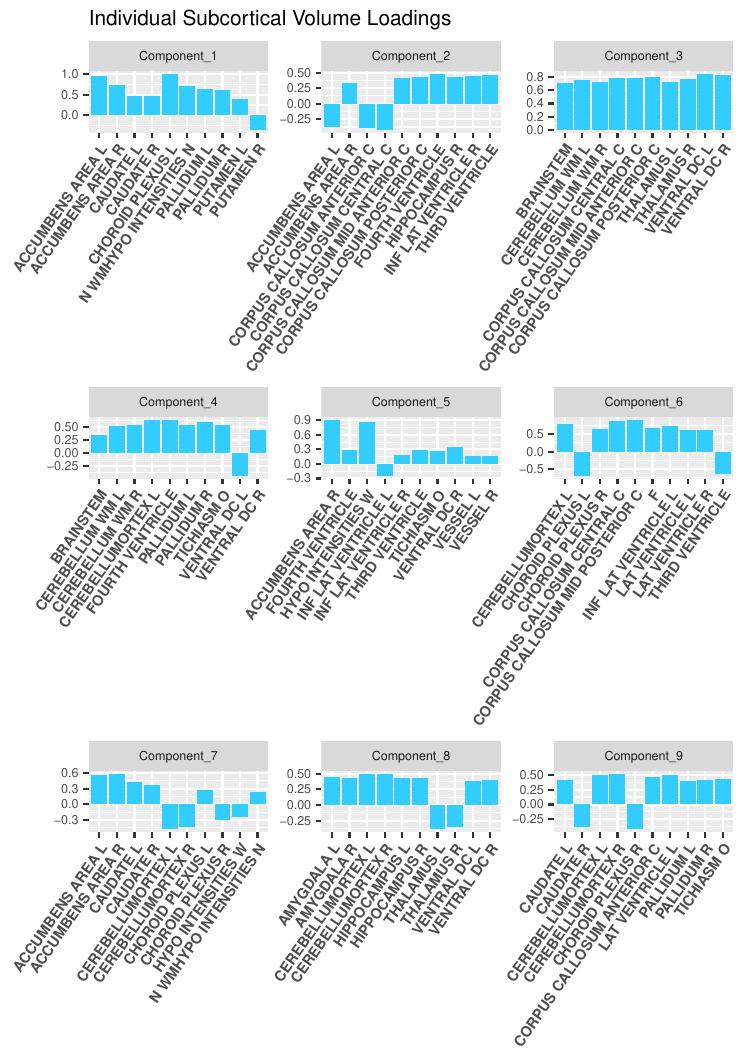} \\
    \caption{Variable loadings for the individual brain morphometry subspaces estimated via ProJIVE in the setting $K=5$. (a) CT, (b) CVol, (c) SA, (d) SVol.}
    \label{fig:indiv.brain.loadsK5}
\end{figure}

Comparing individual morphometry loadings across the two cases is not as straightforward since each set of morphometry measures has its subspace in the $K=5$ case, and all morphometry measures share an individual subspace in the $K=2$ case. In the $K=2$ case, several components of the individual morphometry subspace appear to reflect a particular type of morphometry measure. For instance, component 1 is dominated by SA measures, component 2 by CT measures, and component 4 by SVol measures. However, the ROIs that appear in component 1 do not dominate any component of the individual SA subspace found in the $K=5$ case, and similarly for CT and SVol. It is important to note that the individual morphometry subspaces estimated in the $K=2$ and $K=5$ cases represent quite different conceptualizations of the data. Therefore, divergence in their signals is to be expected. Each set of individual morphometry loadings from the $K=5$ case is displayed in Figure \ref{fig:indiv.brain.loadsK5}.

\subsection{Individual subject scores}
As in the $K=2$ case, all four individual cognition subject scores were associated with diagnoses (section 4.5.2), with component one exhibiting the strongest association (McFadden's pseudo-$R^2=0.11$). Components one and four were associated with location shifts between all three diagnoses (CN, MCI, and AD), while component two was associated with MCI and component three with AD. Furthermore, pseudo-$R^2$ was much higher for component three ($\approx$0.04) than the same values for components three ($\approx$0.02) and four ($\approx$0.01.) Associations between ApoE4 counts and individual cognition scores were strongest for components one and four. However, both were far weaker, each with pseudo-$R^2=0.01$, than the third component in the $K=2$ case. As for the PET biomarkers, component 1 most strongly associated with AV45 ($R^2 = 0.06$) and component 2 with FDG ($R^2 = 0.04$)

As in the main analysis of $K=2$ datasets, we found no significant associations between individual brain morphometry subject scores and diagnosis or ApoE4 allele counts. Individual subject scores from cortical thickness were associated with AV45 ($R^2 = 0.02$) and  FDG ($R^2 = 0.06$), especially component 9. Cortical volume scores were associated with FDG ($R^2=0.02$). Surface area scores were not associated with any exogenous variables. Subcortical volume scores were associated with with FDG ($R^2=0.04$) and AV45 ($R^2=0.02$).

\newpage
\bibliographystyle{apalike}
\bibliography{refs}

\AddToHook{enddocument/afteraux}{%
\immediate\write18{
cp output.aux SupplementalMaterial.aux
}%
}